\documentclass{article}

\usepackage[preprint]{output}

\usepackage[utf8]{inputenc} % allow utf-8 input
\usepackage[T1]{fontenc}    % use 8-bit T1 fonts
\usepackage{hyperref}       % hyperlinks
\usepackage{url}            % simple URL typesetting
\usepackage{booktabs}       % professional-quality tables
\usepackage{amsfonts}       % blackboard math symbols
\usepackage{nicefrac}       % compact symbols for 1/2, etc.
\usepackage{microtype}      % microtypography
\usepackage{xcolor}         % colors
\usepackage{amssymb}
\usepackage{xcolor}
\usepackage{amsmath}
\usepackage{multirow} % Add the multirow package
\usepackage{algorithm}
\usepackage{caption}
\usepackage{subcaption}
\usepackage{lipsum}%% a garbage package you don't need except to create examples.
\usepackage{fancyhdr}
\usepackage{algpseudocode}
\usepackage{makecell}
\usepackage{wrapfig}
\usepackage{booktabs}
\usepackage{hyperref}
\usepackage{url}
\usepackage{graphicx}
\usepackage{hyperref}
\usepackage{dirtytalk}
\usepackage{url}

\newcommand{\blank}{{\mspace{1mu}\cdot\mspace{1mu}}}

\title{Changes by Butterflies: Farsighted Forecasting with Group Reservoir Transformer}

\author{
Md Kowsher and Abdul Rafae Khan and Jia Xu} 
\begin{document}

\maketitle

\begin{abstract}
In Chaos, a minor divergence between two initial conditions exhibits exponential amplification over time, leading to far-away outcomes, known as the butterfly effect. Thus, the distant future is full of uncertainty and hard to forecast.  
We introduce Group Reservoir Transformer to predict long-term events more accurately and robustly by overcoming two challenges in Chaos: 
(1) the extensive historical sequences and (2) the sensitivity to initial conditions.
A reservoir is attached to a Transformer to efficiently handle arbitrarily long historical lengths, with an extension of a group of reservoirs to reduce the sensitivity to the initialization variations. 
Our architecture consistently outperforms state-of-the-art  models in multivariate time series, including TimeLLM, GPT2TS, PatchTST,  DLinear, TimeNet, and the baseline Transformer, with an error reduction of up to -59\% in various fields such as ETTh, ETTm, and air quality, demonstrating that an ensemble of butterfly learning can improve the adequacy and certainty of event prediction, despite of the traveling time to the unknown future. 
\end{abstract}

\section{Introduction}

Chaos theory offers a framework that underlies patterns and deterministic laws governing dynamical systems, applied in human society, such as biology, chemistry, physics, economics, and mathematics, among others ~\citep{liunonlinear}. Chaos are hard to predict due to the \say{butterfly effect}~\citep{jordan2007nonlinear}, i.e., when two initial conditions exhibit only a minor disparity, their divergence over time undergoes exponential amplification, contingent upon the Lyapunov time inherent to the system's dynamics \citep{wu2024generalized}. Consequently, forecasting the distant future in chaotic systems presents formidable challenges primarily rooted in two fundamental issues: (1) the need to account for extensive historical sequences and (2) the pronounced sensitivity to initial conditions.

Prior research has explored a multitude of techniques in chaotic event prediction, with a notable emphasis on the utilization of deep neural network models, exemplified by Transformer~\citep{zeng2022transformers}. Nevertheless, the Transformer architecture's quadratic time and memory complexity of the input length significantly limits the performance of long-term forecasting tasks~\citep{vaswani2017attention}. Several notable efforts, including but not limited to UnlimiFormer~\citep{bertsch2023unlimiformer}, Efficient Transformer~\citep{tay2022efficient}, and Reformer~\citep{kitaev2020reformer},  extend the input length of Transformers but primarily hinge on modifications to the attention model itself, often founded on certain assumptions to extend the  input length to a fixed value without considering the temporal dependency of the very far away events, thereby yielding limited adaptability for learning from and predicting arbitrary long sequences.

\textbf{Stream reservoir learning:} We integrate reservoir computing, as an \textbf{autoregressive} model, into the Transformer framework that effectively models sequences of inputs with arbitrary lengths. Reservoir computing, known for its simplicity and computational efficiency, excels in processing temporal data within the context of chaotic time series prediction~\citep{bollt2021explaining}. 
Within our architecture, the reservoir captures dependencies across all temporal events by discerning the sequence of events based on all historical data without window-sized context restriction. The reservoir learning processes the input in a stream fashion and requires \textbf{no training} by freezing reservoir weights. Its linear time complexity and the constant space complexity to the input length make handling the arbitrary context length possible. 
Subsequently, the Transformer digests short-term events in quadratic time, melding its insights with long-term events modeled by the reservoir to produce a more accurate final forecast. 
Our framework grapples with two technical obstacles related to Lyapunov time and the butterfly effect using the following two techniques: nonlinear readout and group reservoir. 

\textbf{Non-linearity:} The traditional linear readout is expensive since the number of reservoir nodes scales quadratically with the reservoir output size, which mirrors the input size of the Transformer. Our system employs a nonlinear readout rooted in single-head attention mechanisms to manage longer sequences efficiently, superseding the traditional linear reservoir readout. Given its heightened expressiveness, this nonlinear readout reduces reservoir outputs' dimensionality, thus providing the Transformer with more meaningful and potent feature inputs and dimension reduction.

\textbf{Group reservoir:} To stabilize the sensitivities on varied initializations, we incorporate ensemble learning using multiple reservoirs, we term \say{group reservoirs}. Each reservoir in the group takes a random initialization, and the output from their ensemble seamlessly integrates with the observation of the recent events modeled by the Transformer,  facilitating short-term context comprehension. 

Empirical results demonstrate that the nonlinear readout and group reservoir drastically improve the Transformer and its variants in time-series predictions with an error reduction of up to 59.70\%.
Our methodology surpasses Transformer and other leading-edge deep neural networks (DNNs). It aptly handles very long input lengths, signaling a notable improvement for the Transformer model.

Our contributions are mainly fourfold: (1) We integrate reservoir computing into the system, enabling the Transformer to tackle long-term predictions effectively. (2) We refine traditional reservoir computing techniques by improving the linear readout with a nonlinear counterpart, facilitating streamlined dimensionality reduction while keep the autoregressive learning on temporal patterns. (3) We implement the group reservoir to address the nuances of reservoir initialization sensitivity. (4) Our approach demonstrated superior performance in forecasting chaotic phenomena in long-term horizons. Furthermore, it achieved state-of-the-art results in forecasting multivariate time series data.

\section{Background}
\textbf{Preliminary:} We consider a time series dataset denoted by $\mathbf{u}_{1:T}$, where each element $\mathbf{u}_t \in \mathbb{R}^{N_u}$ represents a $N_u$-dimensional vector at time $t \in \{1, \cdots, T\}$. Given a rolling forecasting scenario with a fixed window size $w$, our objective is to predict $\tau$ future time steps, $\mathbf{u}_{t+1:t+\tau}$, based on the input $\mathbf{u}_{t-w:t}$. We propose the following model:
\begin{equation}\label{eq:model}
    \mathbf{\hat{u}}_{t+1:t+\tau} = \mathbb{M}(\mathbf{u}_{t-w:t}, \mathbf{\Phi})
\end{equation}
Here, $\mathbf{\Phi}$ represents learnable parameters, and $\mathbb{M}(\cdot)$ is a Transformer time-series forecasting model. % designed to forecast the next $\tau$ time steps. 

\section{Method}
Transformer has a quadratic time complexity in terms of the history length and prediction future time steps~\citep{jin2024timellm, nie2022time,beltagy2020longformer}. 
%Using transformers in time series analysis significantly improves prediction accuracy due to their attention mechanisms \cite{jin2024timellm, nie2022time}. However, increasing the window size \( w \) leads to higher computational costs in terms of memory and processing time. Additionally, if the time series data have long-term dependencies beyond the size of the window, these may not be captured effectively by the model \cite{beltagy2020longformer}. 
%To overcome these limitations, we integrate deep reservoir computing with the transformer approach. This integration is particularly aimed at enhancing the model's ability to handle multivariate time series data with chaotic elements, 
To reduce the computational complexity of window sizes, we introduce reservoir computing $\mathcal{R}$ to model the entire history of the time series efficiently: % within a fixed-size memory $m$, enabling us to process any input length efficiently:
\begin{equation}
    \mathbf{\hat{u}}_{t+1:t+\tau} = \mathbb{M}(\mathbf{u}_{t-k:t} \uplus \mathcal{R}(\mathbf{u}_{1:T}), \mathbf{\Phi})\label{eq-reservoir}
\end{equation}
Here, $\mathcal{R}$ is a Reservoir, which takes input as time steps sequentially and outputs $m$ size vectors.  The concatenation operator $\uplus$ combines the short-term window $\mathbf{u}_{t-k:t}$ with the long-term historical context $\mathcal{R}(\mathbf{u}_{1:T})$. The resulting effective input size is significantly smaller, $m + k \ll w$, where $k$ is the reduced window size, and  $\mathcal{R}$ handles all history without input length restriction. %, optimizing both memory usage and computational efficiency. 
Below, we will describe how to model $\mathcal{R}$ is a Reservoir. 

\subsection{Cross-attention for Embedding} \label{sec:corss-att}

We first embed the time series data for the neural networks. 
Since the embedding window has a fixed size, time series sequential input usually are chunked into segments, and the learning happens segment by segment. To make each segment comparable to the others, the full time series input are often normalized. 
%often needs normalization, because the mean and variance of the distribution change over time when the short context window shifts. 
Here we use the reversible instance normalization technique for normalization~\citep{kim2021reversible,zhou2021informer, wu2021autoformer, liu2021pyraformer}, where the full sequence $\mathbf{u}_{1:T}$ is normalized to have zero mean and unit variance. We denote each token embedding as  $\epsilon(\mathbf{u_t})  \in \mathbb{R}^{d_{\epsilon}}$, where $\quad\mathbf{u}_t  \in \mathbb{R}^{d_u}$, and the input $\mathbf{u}_{1:T}$ is embedded into 
${\epsilon({\mathbf{u}}_1), \epsilon({\mathbf{u}}_2), \ldots , \epsilon({\mathbf{u}}_T })$. 
To address the dependency of each time step in one segment, we apply the cross-attention~\citep{zhang2022crossformer,rombach2022high}, where each vector  $\epsilon(\mathbf{u}_{t-k:t})$ considers the local temporal relationship with its  $i$-window-sized neighbors $\epsilon(\mathbf{u}_{t-i: t+ i})$: 

%\textcolor{blue}{ Previous Transformer-based models mainly capture cross-time dependency, while the cross-dimension dependency is not explicitly captured during embedding, which limits their forecasting capability~\citep{zhang2022crossformer}\textcolor{red}{what do you mean?}. To address this, we apply a cross-attention mechanism \cite{rombach2022high}, where each vector $\epsilon(\mathbf{u}_t)$ considers the local temporal relationships with its neighboring vectors, thereby enhancing the model's ability to interpret time-based patterns\textcolor{red}{what do you mean?}. Specifically, if we consider an $i$-neighbor window, then $\epsilon(\mathbf{u}_{t-i: t+ i})$ are local neighbors for $\epsilon(\mathbf{u}_{t-k:t})$:}
\begin{equation}
    \mathbf{h}_{t} = \text{softmax}\left(\frac{\mathbf{W}_k\cdot\epsilon(\mathbf{u}_{t-k:t}) (\mathbf{W}_{q}\cdot\epsilon(\mathbf{u}_{t-i: t+ i}))^T}{\sqrt{d_{\epsilon}}}\right)\mathbf{W}_{v}\cdot\epsilon(\mathbf{u}_{t-i: t+ i}), 
\end{equation}

where $ \mathbf{h}_{t} \in \mathbb{R}^{k \times d_\epsilon}$ carries the local temporal relationship for time step $t$. 
Here $\mathbf{W}_q$, $\mathbf{W}_k$, $\mathbf{W}_v$ are trainable matrices to learn queries, keys, and values, respectively~\citep{wei2020multi}.

%\textcolor{blue}{This method improves the model's ability to capture time-related patterns by concentrating on nearby time points.  }\textcolor{red}{what do you mean?}

\subsection{Deep Reservoir Computing} \label{sec:drc}

The Echo State Network (ESN)~\citep{jaeger2004harnessing} implements discrete-time dynamical systems as one of the reservoir computing frameworks~\citep{lukovsevivcius2009reservoir,verstraeten2007experimental}. 
The recurrent reservoir layers are untrained and provide a suffix-based Markovian representation of the past input history~\citep{tivno2007markovian}, attaching a trained linear readout. 
%In this work we consider the Leaky Integrator ESN (LI-ESN) model \cite{jaeger2007optimization}, a variant of the basic ESN in which leaky integrator reservoir units are adopted.

\iffalse
by means of the computation carried out by an untrained recurrent reservoir layer, providing a suffix-based Markovian representation of the past input history \cite{tivno2007markovian}, and by a trained linear readout.

, is a state-of-the-art approach for efficiently modeling RNNs. ESNs

by means of the computation carried out by an untrained recurrent reservoir layer, providing a suffix-based Markovian representation of the past input history \cite{tivno2007markovian}, and by a trained linear readout. In this work we consider the Leaky Integrator ESN (LI-ESN) model \cite{jaeger2007optimization}, a variant of the basic ESN in which leaky integrator reservoir units are adopted.
\fi

We use the Leaky Integrator ESN (LI-ESN) model \citep{jaeger2007optimization}, a variant of the basic ESN, in which leaky integrator reservoir units are adopted. 
In a LI-ESN with \( N_u \) input units and \( N_r \) reservoir units the state is updated according to the following state transition function:
\begin{equation}
\mathbf{x}_{t} = (1-\alpha)\mathbf{x}_{t-1} + \alpha\tanh(\mathbf{W}_{in}\mathbf{h}_{t} + \boldsymbol{\theta} + \mathbf{W}\mathbf{x}_{t-1}), 
\end{equation}
where \( \mathbf{x}_t \in \mathbb{R}^{N_r} \) denote the input and the reservoir state at time \( t \), respectively. \( \mathbf{W}_{in} \in \mathbb{R}^{N_r \times d_\epsilon} \) is the input-to-reservoir weight matrix, \( \boldsymbol{\theta} \in \mathbb{R}^{N_r} \) is the bias-to-reservoir weight vector. % where we assume one input bias equal to 1 for the reservoir units. 
\( \mathbf{W} \in \mathbb{R}^{N_r \times N_r} \) is the recurrent reservoir weight matrix, \( \tanh \) is the element-wise hyperbolic tangent activation function, and \( \alpha \in [0,1] \) is the leaky parameter. The reservoir parameters are initialized according to the constraints specified by the Echo State Property (ESP) \citep{jaeger2004harnessing,tivno2007markovian} and then are left untrained, which makes its computing efficient. Accordingly, the weight values in \( \mathbf{W}_{in} \) and in \( \boldsymbol{\theta} \) are chosen from a uniform distribution over \( [-\sigma_{in}, \sigma_{in}] \), where \( \sigma_{in} \) represents an input-scaling parameter. The values in matrix \( \mathbf{W} \) are randomly selected from a uniform distribution and then re-scaled such that the spectral radius (i.e. the largest eigenvalue in absolute value) of matrix \( \mathbf{W} = (1-\alpha)I + \alpha \mathbf{W} \), denoted by \( \rho \), is smaller than 1, i.e. \( \rho < 1 \)~\citep{jaeger2004harnessing}.  Note that the leaky parameter \( \alpha \) and the spectral radius of the recurrent weight matrix \( \rho \) represent two relevant reservoir computing hyper-parameters.  %The value of \( \alpha \) is related to the speed of reservoir dynamics in response to the input, with larger values of \( \alpha \) resulting in reservoirs that react faster to the input \citep{lukovsevivcius2009reservoir}. The value of \( \rho \), besides of being linked to the ESP for valid ESN initialization, is related to the variable memory length and the degree of the connectivity of reservoir dynamics \citep{jaeger2004harnessing}, with larger values of \( \rho \) resulting in longer memory length \citep{gallicchio2017deep}.
$\alpha$ is set to a fixed value of $0.7$ for simplicity, and $\rho$ is uniform randomly sampled between [$0$,$1$].

\subsection{Linear Readout} \label{sec:lr}
As reservoir output, a readout component is used to linearly combine the outputs of all the reservoir units as in a standard ESN. The output of a reservoir at each time step $t$ can be computed as 
\begin{eqnarray} \label{linear_rd}
\mathbf{y}_t = \mathbf{W}_{out} {\mathbf{x}_{t}} +  \boldsymbol{\theta}_{out}, 
\end{eqnarray}

where  $\mathbf{y}_t \in \mathbb{R}^m$ is the linear readout with $m$ dimensionality.  $\mathbf{W}_{out} \in \mathbb{R}^{m\times N_r}$ represents the reservoir-to-readout weight matrix, connecting the reservoir units to the units in the readout. $\boldsymbol{\theta}_{out} \in \mathbb{R}^{m}$ is the bias-to-readout weight vector. 
%Training a deep- ESN can therefore be accomplished by direct methods, similar to a standard ESN. Although the main focus of our experimental analysis is on reservoir dynamics, readout training is also considered in this paper for the purposes of the MC task.
Training the readout involves solving a linear regression problem, typically approached by using Moore-Penrose pseudo-inversion, with further details in ~\cite{lukovsevivcius2009reservoir,jaeger2004harnessing}. 
Since only $\mathbf{W}_{out}$ are trained, the optimization problem boils down to linear regression.

\subsection{Nonlinear Readout } \label{sec:nr}
The output of the reservoir is the input of the Transformer. One downside of the reservoir is that if the reservoir output size is large, then the Transformer training will be slow due to its quadratic complexity to its input size.  Since a linear readout of a reservoir needs a much larger size to have the same expressive power as the nonlinear readout, we introduce nonlinear readout layers, which work quite well in reducing reservoir output dimension and improving prediction performance hereafter. Therefore, for a faster convergence and avoiding a local optimum~\citep{triefenbach2011can}, we append a nonlinear readout to the linear readout using the following non-linear activation function:  

\begin{equation}
    \mathbf{\Lambda}_t^l = \sigma(\mathbf{y}_t^l),
\end{equation}

    where $\sigma(\mathbf{y}_t^l)$ is the $l$-th reservoir output, and we use $L$ many reserviors in the following section.

\subsection{Group Reservoir} \label{sec:gr}

In Chaos theory, the butterfly effect refers to the sensitive dependence on initial conditions, where a small change in one state of a deterministic nonlinear system can result in large differences in a later state. As ensemble reduces the variants of the prediction, we integrate multiple reservoirs' nonlinear readout layers for the stability of our prediction. We consider \(L\) reservoirs, each with distinct decay rates (\(\alpha\) and \(\rho\)), initialized randomly \citep{gallicchio2017deep}. %Let \(\mathbf{r}_t^l\) represent the nonlinear readout of the \(l\)-th reservoir. We combine various leaky parameters (\(\alpha\)) and spectral radius (\(\rho\)) across the reservoirs. 
This results in a grouped Echo State Network (ESN) representation \(\mathbf{o}_t\), achieved by element-wise \(\oplus\) addition of all reservoirs' non-linear activation outputs:
\begin{equation}\label{eq:gr2}
\mathbf{o}_t = \mathbf{\Lambda}_t^1 \oplus \mathbf{\Lambda}_t^2 \oplus \ldots  \oplus \mathbf{\Lambda}_t^L
    %\mathbf{o}_t = \mathbf{r}_t^1 \oplus \mathbf{r}_t^2 \oplus \ldots  \oplus \mathbf{r}_t^G
\end{equation}

\subsection{Group Reservoir Readout with Self-Attention} \label{sec:gra}

In order to capture the input feature importance,  alleviating the problem of vanishing gradient of long-distance dependency, we have used a self-attention mechanism from the aggregate output of group reservoir in Equation \ref{eq:gr2} by following \cite{shaw2018self} :
\begin{eqnarray}
   e_{ij} = \frac{(\mathbf{o}_t^i  W^Q)(\mathbf{o}_t^j W^K)^T}{\sqrt{d_{\epsilon}}}\\
      \alpha_{ij} = \frac{\exp{e_{ij}}}{\sum_{k=1}^{n} \exp{e_{ik}}}\\
   \mathbf{z_t} = \sum_{j=1}^{n} \alpha_{ij} (\mathbf{o}_t^j W^V) 
   \label{eq:self_att_rd}
\end{eqnarray}

\subsection{Group Reservoir Transformer (RT)} \label{sec:rt}

Our group reservoir efficiently models observed time-series data in dynamical systems and require
minimal training data and computing resources~\citep{gauthier2021next} and is ideal for handling long sequential inputs. 
Here we mainly describe adopting a reservoir for Transformer as one of the most well-known DNN architectures. However, our methods are adaptable to other DNN model architectures, for whom we also compared results in our experiments. 

\iffalse
\begin{figure*}[!t]
%\begin{wrapfigure}{r}{0.7\textwidth}
%\captionsetup{font=footnotesize}
\caption{ }

\label{fig:memory}
 \begin{center}
      \includegraphics[width=1.05\linewidth]{figures/final_reservoir.pdf}
   \end{center}
%\end{wrapfigure}
\end{figure*}

\begin{figure*}[!t]
    \centering
    \begin{minipage}[t]{.50\textwidth}
        \centering
        \includegraphics[width=0.950\linewidth]{figures/reservoir.jpg}
        \caption*{\textbf{Figure a}: Non-linear readout for RT.}
        \label{figrt:sub1}
    \end{minipage}%
    \hfill
    \begin{minipage}[t]{.50\textwidth}
        \centering
        \includegraphics[width=0.950\linewidth]{figures/ensembling.jpg}
        \caption*{\textbf{Figure b}: Ensemble of multiple reservoirs }
        \label{figrt:sub2}
    \end{minipage}
    \caption{Description of deep Reservoir Transformer: In Figure (a), the readout obtained from self-attention, combined with the present input $\vec{u}(t+1)$, serves as the input for training the transformer architecture. In Figure (b), the input $\vec{u}(t)$ is processed individually by $Q$ reservoirs. The nonlinear output of each reservoir is later combined to form the ultimate output, labeled as $\vec{o}(t)$.}
    \label{fig:attentionreservoir}
\end{figure*}
\fi

%\begin{wrapfigure}{R}{0.75\textwidth}
\begin{figure}
    %\hspace{-2em}
    \centering
    \includegraphics[width=0.95\textwidth]{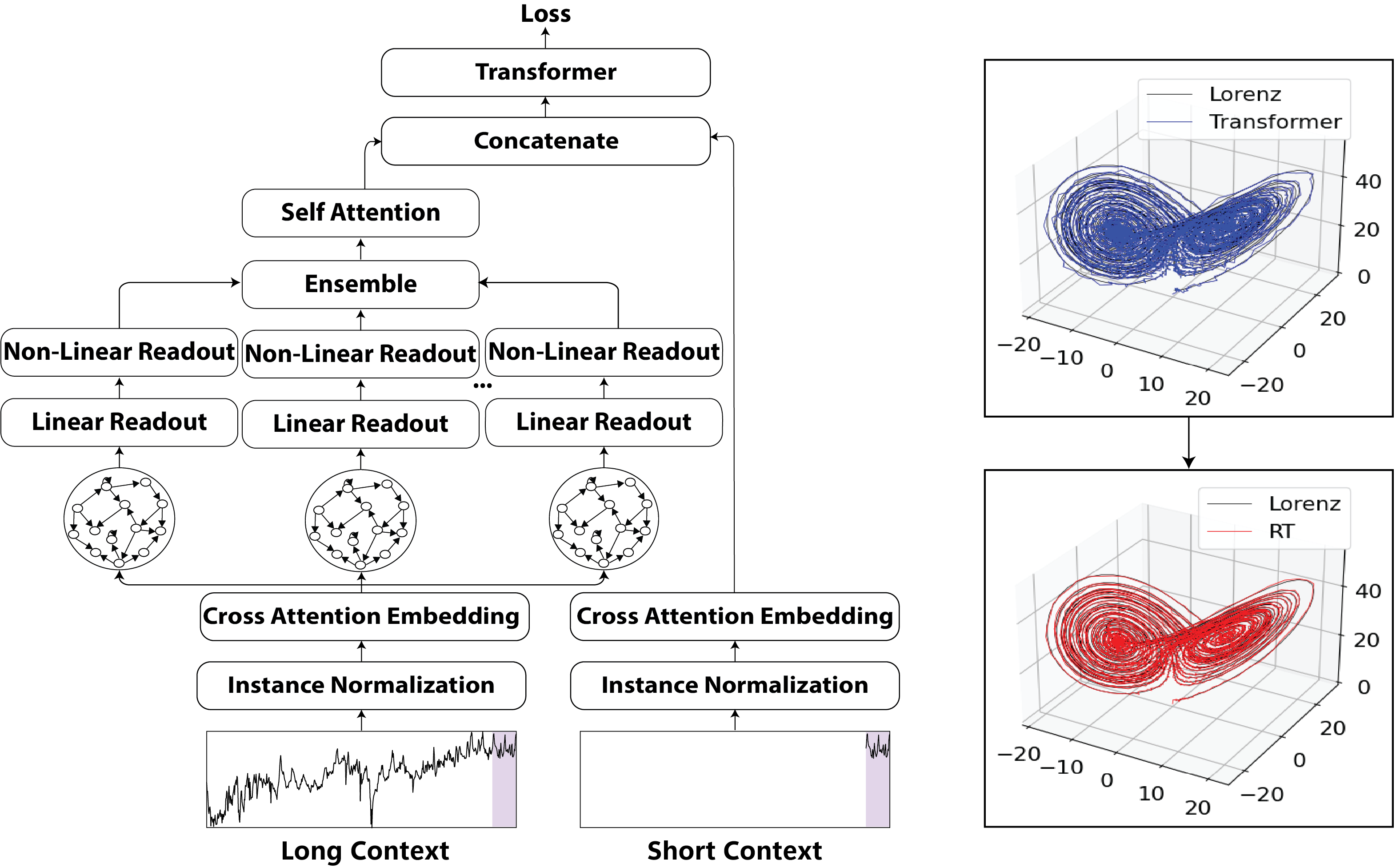}
    \caption{Left: Group Reservoir Transformer; Right: Our RT prediction is closer to the ground truth than the Transformer in Lorenz Extractor.}
    %\vspace{-2em}
    \label{fig:attentionreservoir}
\end{figure}
%\end{wrapfigure}

Figure~\ref{fig:attentionreservoir} (Left) shows the architecture of our reservoir-enhanced Transformer. We enhance the non-linear readout layers through self-attention mechanisms within ensemble reservoirs. This involves updating these layers and performing element-wise addition of the readouts, resulting in an output denoted as $\mathbf{z}_t$. This output is concatenated with the present input embedding $\mathbf{h}_t$, yielding $\mathbf{f}_t$, which then undergoes processing via the Transformer encoder $\mathbb{M}(.)$.

Our approach enables comprehensive modeling of entire historical data in decision-making. While the Transformer captures temporal dependencies, limited by a window of size $w$, our method incorporates reservoir pooling to overcome this limitation. This permits the consideration of the entire history without a rise in time complexity. By incorporating Equation \ref{eq:gr2} (ensemble of all $L$ reservoir readouts) and then following Equation~\ref{eq:self_att_rd}, we have:  
%\begin{equation}\label{drt2}
%    \mathbf{z}_t = (1- \kappa) \mathbf{o}_t \uplus  \kappa \mathbf{h}_{t-k:t}
%\end{equation}
\begin{equation}\label{eq:concat}
    \mathbf{f}_t = (1 - \kappa) \mathbf{z}_t \uplus \kappa \mathbf{h}_{t}
\end{equation}
\begin{equation}\label{eq:drt3}
    \mathbf{\hat{u}}_{t+1:t+\tau} = \mathbb{M}(\mathbf{f}_t)
\end{equation}

Here, $\uplus$ is the concatenation between group reservoirs readout and input time sequence, $\mathbf{f}_t \in \mathbb{R}^{m+k}$ denotes the input of Transformer $\mathbb{M}$ and $\kappa$ is a learnable parameter indicating the weights of input states and reservoir readout to predict $\mathbf{\hat{u}}_{t+1:t+\tau}$ by Equation \ref{eq:drt3} as a realization of Equation~\ref{eq-reservoir}. $\mathbf{z}_t$ is trained on the full historical sequence, and 
$\mathbf{h}_{t}$ is trained only on the most recent short history. 

\subsection{Training} \label{sec:train}
In this work, we use the Huber loss function \citep{meyer2021alternative} as our objective (Equation \ref{eq:regressio}): 
\begin{eqnarray}
\mathcal{L}(\mathbf{u}_{t+1:t+\tau}, \mathbf{\hat{u}}_{t+1:t+\tau})= \frac{1}{\tau}\sum_{i=t+1}^{t+\tau}
\begin{cases}
  \frac{1}{2}(\mathbf{u}_{i} - \mathbf{\hat{u}}_{i})^2 , & \text{if } | \mathbf{u}_{i} - \mathbf{\hat{u}}_{i}| \le \delta \\
  \delta (|\mathbf{u}_{i} - \mathbf{\hat{u}}_{i}| -\frac{1}{2} \delta ), & \text{otherwise}
\end{cases}
\label{eq:regressio}
\end{eqnarray}

Here $\delta$ is an adjustable parameter that controls where the function change occurs to keep the function differentiable.  For a comprehensive understanding of our training algorithm, refer to the pseudo-code provided in the Appendix~\ref{sec:code_rt}, along with a detailed explanation of the notation employed.
Initially, we sequentially train the model $ \mathbb{M}(.)$ across all $T$ time stamps using all the training dataset. Subsequently, we fine-tune our model using the validation set to achieve optimal model performance. The model is subjected to testing with varying hyper-parameters such as reservoir size, leaky rate, spectral radius, number of reservoirs, learning rate, attention size, Transformer block, and dropout rate. The choice of these hyper-parameters may differ depending on the specific dataset with details in Appendix~\ref{dataset_details}. 

\section{Experiments}
%RT consistently performs better than the latest long-term forecasting methods by a significant margin across various benchmarks and conditions. We compared our method with a wide range of current models, including a recent study that adapts a language model for time series analysis. To ensure a fair comparison, we followed the experimental setups described in ~\citep{zeng2023transformers} for all baseline models using a consistent evaluation process. Additionally, we have presented an analysis of chaotic behavior prediction in our experiments.

\textbf{Baselines:} We use well-known time-series forecasting models for our baseline comparisons, including Transformer based systems such as DLinear \citep{zeng2023transformers}, PatchTST \citep{nie2022time}, TimesNet \citep{wu2022timesnet}, FEDformer \citep{zhou2022fedformer}, Stationary \citep{liu2022non}, ESTformer \citep{woo2022etsformer} , LightTS \citep{campos2023lightts}, and Autoformer \citep{chen2021autoformer}, and pre-trained LLMs-based systems such as TimeLLM \citep{jin2024timellm} and GPT2TS \citep{zhou2024one}. The experimental setup follows ~\citep{zeng2023transformers} for all baseline models. %\textcolor{red}{for each citation, please write the name of the model. are they all and non-exclusively used in experiments?}

\textbf{Datasets:}
We evaluate RT on several datasets commonly used for benchmarking long-term multivariate forecasting models, including four ETT datasets (ETTh1, ETTh2, ETTm1, and ETTm2)  \citep{zhou2021informer}, Weather, Electricity (ECL), Traffic, and ILI  from~\cite{zeng2022transformers}.
%\textcolor{red}{citation of each}. Details on the implementation and datasets are available in Appendix B. 
To test the chaotic behavior on these datasets, we employ the correlation dimension, denoted as $D_2$ to serve as a reliable measure of chaotic behavior, with higher values reflecting more pronounced chaos~\citep{panis2020detection}. %, i.e., $D_2 = \underset{\epsilon \to 0}{\lim} \frac{\ln{ C(\epsilon)}}{\ln{\epsilon}}$. 
We compute correlation dimension \(D_2\) across multiple datasets using diverse state numbers ($100$, $500$, and $1000$). Table~\ref{tab:chaoticmeasure} shows the $D_2$ of each dataset. 
%Table ~\ref{tab:chaotic_data} and ~\ref{tab:chaotic} display the outcomes of a correlation dimension ($D_2$) analysis across various datasets. The correlation dimension gauges chaotic tendencies, with higher values suggesting heightened chaos. The findings reveal that, for all datasets, the correlation dimension increases with higher state numbers, indicating intensified chaotic behavior. Since MGS, ECG, and LA datasets are highly chaotic datasets, they showed a higher $D_2$. 

\label{tab:chaotic}
  \begin{table}[!ht]
\centering
\setlength{\tabcolsep}{3pt}
\renewcommand{\arraystretch}{1.1}
\fontsize{10pt}{100pt}\selectfont
% Changing the font size
\scriptsize 

\begin{tabular}{l|lllll|lllll|lll} 
\hline 
 Length & ETTh1 & ETTh2 & ETTm1 & ETTm2 & DDFO & Exchange & ILI & BTC & DGP & DWV & MGS & ECG & LA \\
\hline
100 & 0.790 & 1.190 & 0.560 & 0.299 & 0.540 & 0.995 & 0.995 & 0.961 & 1.070 & 1.343 & 3.242 & 2.913 & 2.515\\
500 & 1.340 & 0.973 & 0.931 & 0.535 & 0.540 & 1.099 & 1.249& 0.853 &1.626 & 1.823  &3.452 & 3.173 & 2.721\\
1000 & 1.266 & 0.866 & 1.061 & 0.667 & 0.540& 1.359 & 1.299 & 0.853 & 1.398 & 1.838  & 3.531 & 3.282 & 2.841\\
&\multicolumn{5}{c|}{low chao}&\multicolumn{5}{c|}{medium chao}&\multicolumn{3}{c}{high chao}\\\hline
\end{tabular}

\vspace{.4cm}
\caption{$D_2$ chaotic behavior measure: MGS, ECG, LA are high chaos; ETTh are low chaos. } 
\vspace{-2em}
\label{tab:chaoticmeasure}
\end{table}

\textbf{Results:} 
The time-series forecasting prediction error rate results are presented in Table~\ref{tab:mlt_cmpr}. The input time series length \(k\) is set to 300, and we test four different prediction horizons \(h \in \{96, 192, 336, 720\}\). The evaluation metrics are mean square error (MSE) and mean absolute error (MAE). Our RT consistently outperforms all baselines in most cases, achieving significant error reductions. Specifically, compared to DLinear, RT reduces MSE by an average of 4.2\%. Similarly, the reductions in MSE for PatchTST and TimeNet are 1.4\% and 2.4\%, respectively. 
Additionally, RT outperforms pre-trained LLM-based time series models in Table~\ref{tab:side-by-side-tables}.  %and univariate time series prediction approaches detailed in the Appendix.
%\textcolor{red}{change the table horizontal and vertical. keep numbers as percentages?}
Table \ref{table:result-mchao} shows that RT model outperforms baseline Transformer 
 on various datasets: Air Quality (AQ), Daily Website Visitors (DWV), Daily Gold Price (DGP), Daily Demand Forecasting Orders (DDF), Bitcoin Historical Dataset (BTC), Absenteeism at Work (AW), and Temperature Readings (TR). All results are reported in MSE and MAE values, with only AW and TR reporting Accuracy and F-Score metrics.
 Comparisons with more baselines in other evaluation criteria are detailed in Appendix~\ref{sec:more_results}. 
 
%of GRIN \citep{cini2021filling}, BRITS \citep{cao2018brits}, ST-MVL \citep{yi2016st}, M-RNN \citep{yoon2018estimating}, ImputeTS \citep{moritz2017imputets} on various regression and classification tasks with respect to MSE and MAE. Evaluation with other criteria such as accuracy, F1 Score, Jaccard Score, Roc Auc, Precision, and Recall are in Appendix.

\begin{table*}[h!]

\label{tab:my-table}
%\resizebox{1.01\textwidth}{!}{%
\setlength{\tabcolsep}{3pt}
\renewcommand{\arraystretch}{1.1}
\fontsize{7pt}{7pt}\selectfont
\begin{tabular}{|cc|cc|cc|cc|cc|cc|cc|cc}
\hline
\multicolumn{2}{c|}{Methods} & \multicolumn{2}{c|}{\begin{tabular}[c]{@{}c@{}}RT\\ (Ours)\end{tabular}} &  \multicolumn{2}{c|}{\begin{tabular}[c]{@{}c@{}}DLinear\\ (2023)\end{tabular}} & \multicolumn{2}{c|}{\begin{tabular}[c]{@{}c@{}}PatchTST\\ (2023)\end{tabular}} & \multicolumn{2}{c|}{\begin{tabular}[c]{@{}c@{}}TimesNet\\ (2023)\end{tabular}} & \multicolumn{2}{c|}{\begin{tabular}[c]{@{}c@{}}FEDformer\\ (2022)\end{tabular}} &   \multicolumn{2}{c|}{\begin{tabular}[c]{@{}c@{}}Stationary\\ (2022)\end{tabular}} & \multicolumn{2}{c}{\begin{tabular}[c]{@{}c@{}}ETSformer\\ (2022)\end{tabular}}  \\ \hline
\multicolumn{2}{c|}{Metric} & \multicolumn{1}{c}{MSE} & MAE & \multicolumn{1}{c}{MSE} & MAE & \multicolumn{1}{c}{MSE} & MAE & \multicolumn{1}{c}{MSE} & MAE & \multicolumn{1}{c}{MSE} & MAE & \multicolumn{1}{c}{MSE} & MAE & \multicolumn{1}{c}{MSE} & MAE \\ \hline
\multicolumn{1}{c|}{\multirow{5}{*}{ETTh1}} & 96 & \multicolumn{1}{c}{\textcolor{red} {\textbf{0.362}}} & {\textcolor{blue} {0.392}}  & \multicolumn{1}{c}{0.376} & 0.397 & \multicolumn{1}{c} {\textcolor{blue}{0.375}}  & 0.399  & \multicolumn{1}{c}{0.370} & 0.399  & \multicolumn{1}{c}{0.384} &  0.402 & \multicolumn{1}{c}{0.376} & 0.419 & \multicolumn{1}{c}{0.449} & 0.479 \\ 

\multicolumn{1}{c|}{} & 192 & \multicolumn{1}{c} {\textcolor{red} {\textbf{0.396}}}   & {\textcolor{red} {\textbf{0.412}}}  & \multicolumn{1}{c}{\textcolor{blue}{0.405}}  & {\textcolor{blue}{0.416}}  & \multicolumn{1}{c}{0.413} & 0.421 & \multicolumn{1}{c}{0.436} &  0.429 & \multicolumn{1}{c}{0.420} &  0.448  & \multicolumn{1}{c}{0.534} & 0.504 & \multicolumn{1}{c}{0.538} & 0.504 \\ 

\multicolumn{1}{c|}{} & 336 & \multicolumn{1}{c} {\textcolor{blue}{0.427}}  & {\textcolor{red} {\textbf{0.422}}}  & \multicolumn{1}{c}{0.439} & 0.443 & \multicolumn{1}{c} {\textcolor{red} {\textbf{0.422}}}  & {\textcolor{blue}{0.436}} & \multicolumn{1}{c}{0.491} & 0.469 & \multicolumn{1}{c}{0.459} & 0.465  & \multicolumn{1}{c}{0.588} & 0.535 & \multicolumn{1}{c}{0.574} &  0.521\\ 
\multicolumn{1}{c|}{} & 720 & \multicolumn{1}{c} {\textcolor{red} {\textbf{0.441}}}  & {\textcolor{red} { \textbf{0.455}}}  & \multicolumn{1}{c}{0.472} & 0.490 & \multicolumn{1}{c} {\textcolor{blue}{0.447}}  & {\textcolor{blue}{0.466}} & \multicolumn{1}{c}{0.521} & 0.500 & \multicolumn{1}{c}{0.506} & 0.507  & \multicolumn{1}{c}{0.643} & 0.616 & \multicolumn{1}{c}{0.562} & 0.535 \\ 
\multicolumn{1}{c|}{} & avg & \multicolumn{1}{c} {\textcolor{red} {\textbf{0.406}}} & {\textcolor{red} {\textbf{0.418}}}  & \multicolumn{1}{c}{0.422} &  0.437 & \multicolumn{1}{c} {\textcolor{blue}{0.413}}  & {\textcolor{blue}{0.430}}& \multicolumn{1}{c}{0.458} & 0.450 & \multicolumn{1}{c}{0.440} &  0.460  & \multicolumn{1}{c}{0.570} & 0.537 & \multicolumn{1}{c}{0.542} &  0.510 \\ \hline
\multicolumn{1}{c|}{\multirow{5}{*}{ETTh2}} & 96 & \multicolumn{1}{c}{\textcolor{red} {\textbf{0.262}}}  &  {\textcolor{red} {\textbf{0.320}}}  &  \multicolumn{1}{c}{0.289} & 0.353 & \multicolumn{1}{c}{\textcolor{blue}{0.274}}  & {\textcolor{blue}{0.336}} & \multicolumn{1}{c}{0.340} & 0.374 & \multicolumn{1}{c}{0.358} & 0.397 &  \multicolumn{1}{c}{0.476} & 0.458 & \multicolumn{1}{c}{0.340} & 0.391 \\ 
\multicolumn{1}{c|}{} & 192 & \multicolumn{1}{c} {\textcolor{red} {\textbf{0.331}}}  & {\textcolor{red} { \textbf{0.373}}} & \multicolumn{1}{c}{0.383} & 0.418 & \multicolumn{1}{c} {\textcolor{blue}{0.339}}  & {\textcolor{blue}{0.379}}  & \multicolumn{1}{c}{0.402} & 0.414 & \multicolumn{1}{c}{0.429} & 0.439  & \multicolumn{1}{c}{0.512} & 0.493 & \multicolumn{1}{c}{0.430} & 0.439 \\ 
\multicolumn{1}{c|}{} & 336 & \multicolumn{1}{c}{\textcolor{blue}{0.358}}  & {\textcolor{blue}{0.403}}  & \multicolumn{1}{c}{0.448} & 0.465 & \multicolumn{1}{c}{\textcolor{red} {\textbf{0.329}}}  & {\textcolor{red} {\textbf{0.380}}}  & \multicolumn{1}{c}{0.452} & 0.452 & \multicolumn{1}{c}{0.496} & 0.487  & \multicolumn{1}{c}{0.552} &  0.551 & \multicolumn{1}{c}{0.485} & 0.479 \\ 
\multicolumn{1}{c|}{} & 720 & \multicolumn{1}{c} {\textcolor{red} {\textbf{0.369}}}  & {\textcolor{red}{\textbf{0.411}}}  & \multicolumn{1}{c}{0.605} & 0.551 & \multicolumn{1}{c} {\textcolor{blue}{0.379}}  & {\textcolor{blue}{0.422}}   & \multicolumn{1}{c}{0.462} & 0.468 & \multicolumn{1}{c}{0.463} & 0.474  & \multicolumn{1}{c}{0.562} & 0.560 & \multicolumn{1}{c}{0.500} & 0.497 \\ 
\multicolumn{1}{c|}{} & avg & \multicolumn{1}{c} {\textcolor{red}{\textbf{0.330}}}  & {\textcolor{red}{\textbf{0.376}}}  & \multicolumn{1}{c}{0.431} & 0.446 & \multicolumn{1}{c}{\textcolor{red}{\textbf{0.330}}}  & {\textcolor{blue}{0.379}}  & \multicolumn{1}{c} {\textcolor{blue}{0.414}}  & 0.427  & \multicolumn{1}{c}{0.439} & 0.452 & \multicolumn{1}{c}{0.602} & 0.543 & \multicolumn{1}{c}{0.450} & 0.459\\ \hline

\multicolumn{1}{c|}{\multirow{5}{*}{ETTm1}} & 96 & \multicolumn{1}{c} {\textcolor{red}{\textbf{0.272}}} & {\textcolor{red}{\textbf{0.336}}}  &  \multicolumn{1}{c}{0.299} & 0.343 & \multicolumn{1}{c}  {\textcolor{blue}{0.290}}  & {\textcolor{blue}{0.342}}  & \multicolumn{1}{c}{0.338} & 0.375 & \multicolumn{1}{c}{0.379} & 0.419  & \multicolumn{1}{c}{0.386} & 0.398 & \multicolumn{1}{c}{0.375} & 0.398 \\ 

\multicolumn{1}{c|}{} & 192 & \multicolumn{1}{c} {\textcolor{red}{\textbf{0.310}}}  &  {\textcolor{blue}{0.357}}    & \multicolumn{1}{c}{0.335} &  {\textcolor{blue}{0.365}}  & \multicolumn{1}{c} {\textcolor{red}{\textbf{0.332}}}  & 0.369 & \multicolumn{1}{c}{0.374} & 0.387 & \multicolumn{1}{c}{0.426} & 0.441 & \multicolumn{1}{c}{0.459} & 0.444 & \multicolumn{1}{c}{0.408} & 0.410 \\ 

\multicolumn{1}{c|}{} & 336 & \multicolumn{1}{c} {\textcolor{red}{\textbf{0.351}}}  &  {\textcolor{blue}{0.382}}   & \multicolumn{1}{c}{0.369} & {\textcolor{blue}{0.386}} & \multicolumn{1}{c} {\textcolor{red}{\textbf{0.366}}}  & 0.392 & \multicolumn{1}{c}{0.410} & 0.411 & \multicolumn{1}{c}{0.445} & 0.459  & \multicolumn{1}{c}{0.495} & 0.464 & \multicolumn{1}{c}{0.435} & 0.428\\ 
\multicolumn{1}{c|}{} & 720 & \multicolumn{1}{c} {\textcolor{red}{\textbf{0.380}}}  & {\textcolor{red}{\textbf{0.409}}}   & \multicolumn{1}{c}{0.425} & 0.421 & \multicolumn{1}{c} {\textcolor{blue}{0.416}}  & {\textcolor{blue}{0.420}}  & \multicolumn{1}{c}{0.478} & 0.450 & \multicolumn{1}{c}{0.543} & 0.490  & \multicolumn{1}{c}{0.585} & 0.516 & \multicolumn{1}{c}{0.499} & 0.462 \\ 

\multicolumn{1}{c|}{} & avg & \multicolumn{1}{c} {\textcolor{red}{\textbf{0.328}}} & {\textcolor{red}{\textbf{0.371}}}   &  \multicolumn{1}{c}{0.357} & {\textcolor{blue}{0.378}} & \multicolumn{1}{c} {\textcolor{blue}{0.351}}  & 0.380 & \multicolumn{1}{c}{0.400} & 0.406 & \multicolumn{1}{c}{0.448} & 0.452 & \multicolumn{1}{c}{0.481} & 0.456 & \multicolumn{1}{c}{0.429} & 0.425 \\ \hline

\multicolumn{1}{c|}{\multirow{5}{*}{ETTm2}} & 96 & \multicolumn{1}{c}{\textcolor{red}{\textbf{0.163}}}  & {\textcolor{red}{\textbf{0.254}}} &  \multicolumn{1}{c}{0.167} & 0.269 & \multicolumn{1}{c} {\textcolor{blue}{0.165}}  & {\textcolor{blue}{0.255}}  & \multicolumn{1}{c}{0.187} & 0.267 & \multicolumn{1}{c}{0.203} & 0.287  & \multicolumn{1}{c}{0.192} & 0.274 & \multicolumn{1}{c}{0.189} & 0.280\\ 

\multicolumn{1}{c|}{} & 192 & \multicolumn{1}{c} {\textcolor{red}{\textbf{0.218}}}  & {\textcolor{red}{\textbf{0.290}}}  & \multicolumn{1}{c}{0.224} & 0.303 & \multicolumn{1}{c} {\textcolor{blue}{0.220}} & {\textcolor{blue}{0.292}}  & \multicolumn{1}{c}{0.249} & 0.309 & \multicolumn{1}{c}{0.269} & 0.328  & \multicolumn{1}{c}{0.280} & 0.339 & \multicolumn{1}{c}{0.253} & 0.319 \\ 

\multicolumn{1}{c|}{} & 336 & \multicolumn{1}{c} {\textcolor{red}{\textbf{0.271}}}  &{\textcolor{red}{\textbf{0.328}}}  & \multicolumn{1}{c}{0.281} & 0.342 & \multicolumn{1}{c} {\textcolor{blue}{0.274}}  & {\textcolor{blue}{0.329}}  & \multicolumn{1}{c}{0.321} & 0.351 & \multicolumn{1}{c}{0.325} & 0.366 & \multicolumn{1}{c}{0.334} & 0.361 & \multicolumn{1}{c}{0.314} & 0.357   \\

\multicolumn{1}{c|}{} & 720 & \multicolumn{1}{c} {\textcolor{red}{\textbf{0.356}}}  & {\textcolor{red}{\textbf{0.382} }}  & \multicolumn{1}{c}{0.397} & 0.421 & \multicolumn{1}{c} {\textcolor{blue}{0.362}}  & {\textcolor{blue}{0.385}}  & \multicolumn{1}{c}{0.408} & 0.403 & \multicolumn{1}{c}{0.421} &  0.415  & \multicolumn{1}{c}{0.417} &  0.413 & \multicolumn{1}{c}{0.414} & 0.413 \\ 

\multicolumn{1}{c|}{} & avg & \multicolumn{1}{c} {\textcolor{red}{\textbf{0.252}}}  & 0.627 & \multicolumn{1}{c}{ 0.267} & {\textcolor{blue}{0.333}}  & \multicolumn{1}{c} {\textcolor{blue}{0.255}}  & {\textcolor{red}{\textbf{0.315}}}  & \multicolumn{1}{c}{0.291} & {\textcolor{blue}{0.333}} & \multicolumn{1}{c}{0.305} & 0.349  & 0.371 & \multicolumn{1}{c|}{0.306} & 0.347 & \multicolumn{1}{c}{0.293} \\ \hline

\multicolumn{1}{c|}{\multirow{5}{*}{Weather}} & 96 & \multicolumn{1}{c}{\textcolor{red}{\textbf{0.146}}}   & {\textcolor{red}{\textbf{0.196}}}   & \multicolumn{1}{c}{0.176} & 0.237 & \multicolumn{1}{c} {\textcolor{blue}{0.149}} & {\textcolor{blue}{0.198}} & \multicolumn{1}{c}{0.172} & 0.220 & \multicolumn{1}{c}{0.217} & 0.296 & \multicolumn{1}{c}{0.173} & 0.223 & \multicolumn{1}{c}{0.197} & 0.281  \\ 

\multicolumn{1}{c|}{} & 192 & \multicolumn{1}{c} {\textcolor{red}{\textbf{0.193}}}   & {\textcolor{red}{\textbf{0.236}}}   &  \multicolumn{1}{c}{0.220} & 0.282 & \multicolumn{1}{c} {\textcolor{blue}{0.194}} 
  &  {\textcolor{blue}{0.241}}   & \multicolumn{1}{c}{0.219} & 0.261 & \multicolumn{1}{c}{0.276} & 0.336  & \multicolumn{1}{c}{0.245} & 0.285 & \multicolumn{1}{c}{0.237} & 0.312 \\ 
  
\multicolumn{1}{c|}{} & 336 & \multicolumn{1}{c} {\textcolor{red}{\textbf{0.239}}}  & {\textcolor{red}{\textbf{0.240}}}  & \multicolumn{1}{c}{0.265} &  0.319 & \multicolumn{1}{c}  {\textcolor{blue}{0.245}}  &  {\textcolor{blue}{0.282}}  & \multicolumn{1}{c}{0.280} & 0.306 & \multicolumn{1}{c}{0.339} & 0.380  & \multicolumn{1}{c}{0.321} & 0.338 & \multicolumn{1}{c}{0.298} & 0.353 \\ 

\multicolumn{1}{c|}{} & 720 & \multicolumn{1}{c} {\textcolor{red}{\textbf{0.301}}}   & {\textcolor{red}{\textbf{0.316}}}  &  \multicolumn{1}{c}{0.333} & 0.362 & \multicolumn{1}{c} {\textcolor{blue}{0.314}}  & {\textcolor{blue}{0.334}}  & \multicolumn{1}{c}{0.365} & 0.359 & \multicolumn{1}{c}{0.403} & 0.428  & \multicolumn{1}{c}{ 0.414} &  0.410 & \multicolumn{1}{c}{0.352} & 0.288 \\ 

\multicolumn{1}{c|}{} & avg & \multicolumn{1}{c} {\textcolor{red}{\textbf{0.219}}}   & {\textcolor{red}{\textbf{0.247}}}   & \multicolumn{1}{c}{0.248} & 0.300 & \multicolumn{1}{c}{\textcolor{blue}{0.225}}  & {\textcolor{blue}{0.264}}  & \multicolumn{1}{c}{0.259} & 0.287 & \multicolumn{1}{c}{ 0.309} & 0.360  & \multicolumn{1}{c}{0.288} & 0.314 & \multicolumn{1}{c}{0.271} & 0.334 \\ \hline

\multicolumn{1}{c|}{\multirow{5}{*}{Electricity}} & 96 & \multicolumn{1}{c}{\textcolor{blue}{0.130}}  & {\textcolor{red}{\textbf{0.219}}}  & \multicolumn{1}{c}{0.140} & 0.237  & \multicolumn{1}{c}{\textcolor{red}{\textbf{0.129}}}  & {\textcolor{blue}{0.222}}  & \multicolumn{1}{c}{0.168} &  0.272 & \multicolumn{1}{c}{0.193} & 0.308  & \multicolumn{1}{c}{0.169} & 0.273 & \multicolumn{1}{c}{0.187} & 0.304 \\ 
\multicolumn{1}{c|}{} & 192 & \multicolumn{1}{c} {\textcolor{blue}{0.154}}   & {\textcolor{blue}{0.243 }}  & \multicolumn{1}{c} {\textcolor{red}{\textbf{0.153}}}   & 0.249 & \multicolumn{1}{c}{0.157} & {\textcolor{red}{\textbf{0.240}}}   & \multicolumn{1}{c}{0.184} & 0.289 & \multicolumn{1}{c}{0.201} &  0.315  & \multicolumn{1}{c}{0.182} & 0.286 & \multicolumn{1}{c}{0.199} & 0.315 \\ 
\multicolumn{1}{c|}{} & 336 & \multicolumn{1}{c} {\textcolor{red}{\textbf{0.158}}}  & {\textcolor{red}{\textbf{0.245}}}  & \multicolumn{1}{c}{0.169} & 0.267 & \multicolumn{1}{c} {\textcolor{blue}{{0.163}}}  & {\textcolor{blue}{0.259}} & \multicolumn{1}{c}{0.198} &  0.300 & \multicolumn{1}{c}{0.214} & 0.329 & \multicolumn{1}{c}{0.200} & 0.304 & \multicolumn{1}{c}{0.212} & 0.329 \\

\multicolumn{1}{c|}{} & 720 & \multicolumn{1}{c} {\textcolor{red}{\textbf{0.190}}}   & {\textcolor{blue}{0.294}}  &  \multicolumn{1}{c}{0.203} & 0.301 & \multicolumn{1}{c} {\textcolor{blue}{0.197}}  & {\textcolor{red}{ \textbf{0.290}}}  & \multicolumn{1}{c}{0.220} & 0.320 & \multicolumn{1}{c}{0.246} & 0.355  & \multicolumn{1}{c}{0.222} & 0.321 & \multicolumn{1}{c}{0.233} & 0.345 \\ 
\multicolumn{1}{c|}{} & avg & \multicolumn{1}{c} {\textcolor{red}{\textbf{0.158}}}  & {\textcolor{red}{\textbf{0.251}}}    &\multicolumn{1}{c}{0.166} & 0.263 & \multicolumn{1}{c}{\textcolor{blue}{{0.161}}}  & {\textcolor{blue}{{ 0.252}}} & \multicolumn{1}{c}{0.192} &  0.295 & \multicolumn{1}{c}{0.214} & 0.327  & \multicolumn{1}{c}{ 0.193} & 0.296 & \multicolumn{1}{c}{ 0.208} & 0.323 \\ \hline

\multicolumn{1}{c|}{\multirow{5}{*}{Traffic}} & 96 & \multicolumn{1}{c}{\textcolor{blue}{{0.361}}}  & {\textcolor{red}{\textbf{0.248}}}  &  \multicolumn{1}{c}{0.410} & 0.282 & \multicolumn{1}{c}{\textcolor{red}{\textbf{0.360}}}   & {\textcolor{blue}{{0.249}}}  & \multicolumn{1}{c}{0.593} &  0.321 & \multicolumn{1}{c}{0.587} & 0.366 & \multicolumn{1}{c}{0.612} & 0.338 & \multicolumn{1}{c}{0.607} &  0.392  \\ 
\multicolumn{1}{c|}{} & 192 & \multicolumn{1}{c}{\textcolor{red}{\textbf{0.376}} }  & {\textcolor{red}{ \textbf{0.250}}} & \multicolumn{1}{c}{0.423} & 0.287 & \multicolumn{1}{c}{\textcolor{blue}{{0.379}}}  & {\textcolor{blue}{ 0.256}}  & \multicolumn{1}{c}{0.617} &  0.336 & \multicolumn{1}{c}{ 0.604} &  0.373 & \multicolumn{1}{c}{0.613} & 0.340 & \multicolumn{1}{c}{0.621} &  0.399  \\ 
\multicolumn{1}{c|}{} & 336 & \multicolumn{1}{c} {\textcolor{red}{\textbf{0.382}} }  & {\textcolor{blue}{0.268}}  &  \multicolumn{1}{c}{0.436} &  0.296 & \multicolumn{1}{c}{\textcolor{blue}{0.392}}  & {\textcolor{red}{ \textbf{0.264}}} & \multicolumn{1}{c}{0.629} & 0.336 & \multicolumn{1}{c}{0.621} & 0.383 & \multicolumn{1}{c}{0.618} &  0.328 & \multicolumn{1}{c}{0.622} & 0.396   \\ 
\multicolumn{1}{c|}{} & 720 & \multicolumn{1}{c} {\textcolor{red}{\textbf{0.426}}}  & {\textcolor{red}{\textbf{0.281}}}  & \multicolumn{1}{c}{0.466} & 0.315 & \multicolumn{1}{c}{\textcolor{blue}{{0.432}}}  &  {\textcolor{blue}{{0.286}}}  & \multicolumn{1}{c}{0.640} & 0.350  & \multicolumn{1}{c}{0.626} &  0.382  & \multicolumn{1}{c}{0.653} &  0.355 & \multicolumn{1}{c}{0.632} & 0.396\\ 
\multicolumn{1}{c|}{} & avg & \multicolumn{1}{c} {\textcolor{red}{\textbf{0.386}}}  & {\textcolor{red}{\textbf{0.261}}}  & \multicolumn{1}{c}{0.433} & 0.295 & \multicolumn{1}{c}{\textcolor{blue}{{0.390}}}   & {\textcolor{blue}{0.263}}   & \multicolumn{1}{c}{0.620} & 0.336 & \multicolumn{1}{c}{0.610} &  0.376 & \multicolumn{1}{c}{0.628} & 0.379 & \multicolumn{1}{c}{0.624} & 0.340 \\ 
\hline

\multicolumn{1}{c|}{\multirow{5}{*}{ILI}} & 24 & \multicolumn{1}{c} {\textcolor{red}{\textbf{1.301}}} & {\textcolor{red}{\textbf{0.739}}}  & \multicolumn{1}{c}{2.215} &  1.081 & \multicolumn{1}{c}{\textcolor{blue}{{1.319}}}   & {\textcolor{blue}{0.754}} 
 & \multicolumn{1}{c}{2.317} &  0.934 & \multicolumn{1}{c}{3.228} &  1.260  & \multicolumn{1}{c}{2.294} & 0.945 & \multicolumn{1}{c}{2.527} & 1.020 \\ 
\multicolumn{1}{c|}{} & 36 & \multicolumn{1}{c}{\textcolor{red}{\textbf{1.409}}}  & {\textcolor{red}{\textbf{0.824}}}  & \multicolumn{1}{c}{1.963} & 0.963 & \multicolumn{1}{c}{\textcolor{blue}{{1.430}}}  & {\textcolor{blue}{0.834}}  & \multicolumn{1}{c}{1.972} &0.920 & \multicolumn{1}{c}{2.679} &  1.080  & \multicolumn{1}{c}{1.825} &  0.848 & \multicolumn{1}{c}{ 2.615} & 1.007\\ 
\multicolumn{1}{c|}{} & 48 & \multicolumn{1}{c} {\textcolor{red}{\textbf{1.520}}}  & {\textcolor{red}{\textbf{0.807} }} & \multicolumn{1}{c}{2.130} & 1.024 & \multicolumn{1}{c}{\textcolor{blue}{{1.553}}} & {\textcolor{blue}{0.815}}  & \multicolumn{1}{c}{2.238} & 0.940 & \multicolumn{1}{c}{2.622} & 1.078  & \multicolumn{1}{c}{2.010} & 0.900 & \multicolumn{1}{c}{2.359} & 0.972 \\ 
\multicolumn{1}{c|}{} & 60 & \multicolumn{1}{c}{\textcolor{blue}{{1.553}}} &  {\textcolor{blue}{0.852}}  &\multicolumn{1}{c}{2.368} & 1.096  & \multicolumn{1}{c}{\textcolor{red}{\textbf{1.520}}}    & {\textcolor{red}{ \textbf{0.788}}}  & \multicolumn{1}{c}{2.027} & 0.928 & \multicolumn{1}{c}{2.857} & 1.157  & \multicolumn{1}{c}{2.178} & 0.963 & \multicolumn{1}{c}{2.487} & 1.016 \\ 
\multicolumn{1}{c|}{} & avg & \multicolumn{1}{c} {\textcolor{blue}{{1.439}}}  & {\textcolor{blue}{0.805}}    & \multicolumn{1}{c}{2.169} &  1.041 & \multicolumn{1}{c}{\textcolor{red}{\textbf{1.443}}}   & {\textcolor{red}{\textbf{0.797}}}  & \multicolumn{1}{c}{2.139} & 0.931 & \multicolumn{1}{c}{2.847} &  1.144 & \multicolumn{1}{c}{2.077} & 0.914 & \multicolumn{1}{c}{2.497} & 1.004  \\ 
\hline

\end{tabular}%

%}
\caption{Multivariate long-term time-series forecasting: lower values indicate better performance. Forecasting horizons are $h \in \{24, 36, 48, 60\}$ for ILI and $h \in \{96, 192, 336, 720\}$ for the rest. %Optimal results are emphasized in \textbf{bold}, while the top results from Transformers are \underline{underlined} for distinction."
}
\label{tab:mlt_cmpr}
\end{table*}

\begin{table*}[h]
\centering

\centering

\setlength{\tabcolsep}{3pt}
\renewcommand{\arraystretch}{1.1}
\fontsize{7pt}{7pt}\selectfont
\begin{tabular}{cc|cc|cc|cc|cc|cc|cc|cc}
\hline
\multicolumn{2}{c|}{Methods} & \multicolumn{2}{c|}{\begin{tabular}[c]{@{}c@{}}RT\\ (Ours)\end{tabular}} & \multicolumn{2}{c|}{\begin{tabular}[c]{@{}c@{}}TIME-LLM\\ (2024)\end{tabular} } & \multicolumn{2}{c|}{\begin{tabular}[c]{@{}c@{}}GPT2TS\\ (2024)\end{tabular}} & \multicolumn{2}{c|}{Methods} & \multicolumn{2}{c|}{\begin{tabular}[c]{@{}c@{}}RT\\ (Ours)\end{tabular}} & \multicolumn{2}{c|}{\begin{tabular}[c]{@{}c@{}}TIME-LLM\\ (2024)\end{tabular} } & \multicolumn{2}{c}{\begin{tabular}[c]{@{}c@{}}GPT2TS\\ (2024)\end{tabular}} \\ \hline
\multicolumn{2}{c|}{Metric} & \multicolumn{1}{c}{MSE} & MAE & \multicolumn{1}{c}{MSE} & MAE & \multicolumn{1}{c}{MSE} & MAE & \multicolumn{2}{c|}{Metric} & \multicolumn{1}{c}{MSE} & MAE & \multicolumn{1}{c|}{MSE} & MAE & \multicolumn{1}{c}{MSE} & MAE  \\ \hline
\multicolumn{1}{c|}{\multirow{5}{*}{ETTh1}} & 96 & \multicolumn{1}{c}{\textcolor{red}{\textbf{0.361}}} & \textcolor{red}{\textbf{0.386}} & \multicolumn{1}{c}{\textcolor{blue}{0.362}} &  \textcolor{blue}{0.392} & \multicolumn{1}{c}{0.376} & 0.397  & \multicolumn{1}{c|}{\multirow{5}{*}{Weather}} & 96 & \multicolumn{1}{c}  {\textcolor{red}{\textbf{0.146}}} & {\textcolor{red}{\textbf{0.196}}}   & \multicolumn{1}{c}  {\textcolor{blue}{0.147}} & {\textcolor{blue}{0.201}} & \multicolumn{1}{c}{0.162} & 0.212  \\

\multicolumn{1}{c|}{} & 192 & \multicolumn{1}{c} {\textcolor{red}{\textbf{0.396}}}  & \textcolor{red}{\textbf{0.412}} & \multicolumn{1}{c} {\textcolor{blue}{0.398}} & {\textcolor{blue}{0.418}}  & \multicolumn{1}{c}{0.416} & {\textcolor{blue}{0.418}}  & \multicolumn{1}{c|}{} & 192 & \multicolumn{1}{c} {\textcolor{blue}{0.193}} & {\textcolor{blue}{0.236}} &  \multicolumn{1}{c}{\textcolor{red}{\textbf{0.189}}}  & {\textcolor{red}{\textbf{0.234}}}& \multicolumn{1}{c}{0.204} & 0.248  \\

\multicolumn{1}{c|}{} & 336 & \multicolumn{1}{c}{\textcolor{red}{\textbf{0.427}}} & \textcolor{red}{\textbf{0.422}} &  \multicolumn{1}{c}{\textcolor{red}{\textbf{0.4427}}} & {\textcolor{blue}{0.433}}  & \multicolumn{1}{c} {\textcolor{blue}{0.439}} &{\textcolor{blue}{0.433}}  & \multicolumn{1}{c|}{} & 336 & \multicolumn{1}{c} {\textcolor{red}{\textbf{0.239}}}  & {\textcolor{red}{\textbf{0.240}}}  & \multicolumn{1}{c}{0.262} & {\textcolor{blue}{0.279}} & \multicolumn{1}{c} {\textcolor{blue}{ 0.254}}  &  0.286  \\ 

\multicolumn{1}{c|}{} & 720 & \multicolumn{1}{c}{\textcolor{red}{\textbf{0.441}}} & \textcolor{red}{\textbf{0.455}} & \multicolumn{1}{c}{\textcolor{blue}{0.442}} & 0.457 & \multicolumn{1}{c}{0.477} & \textcolor{blue}{0.456} & \multicolumn{1}{c|}{} & 720 & \multicolumn{1}{c} {\textcolor{red}{\textbf{0.301}}}  & {\textcolor{red}{ \textbf{0.316}}}  &  \multicolumn{1}{c} {\textcolor{blue}{ 0.300} } & {\textcolor{red}{\textbf{0.316}}}  & \multicolumn{1}{c}{0.326} & {\textcolor{blue}{0.337}}   \\

\multicolumn{1}{c|}{} & avg & \multicolumn{1}{c} {\textcolor{red}{\textbf{0.408}}} & \textcolor{red}{\textbf{0.423}} & \multicolumn{1}{c} {\textcolor{red}{\textbf{0.408}}}  & {\textcolor{red}{0.423}}  & \multicolumn{1}{c} {\textcolor{blue}{0.465}}  & {\textcolor{red}{\textbf{0.455}}}   & \multicolumn{1}{c|}{} & avg & \multicolumn{1}{c} {\textcolor{red}{\textbf{0.219}}}  & {\textcolor{red}{\textbf{0.247}}}  & \multicolumn{1}{c} {\textcolor{blue}{0.225}} & {\textcolor{blue}{0.257}}  & \multicolumn{1}{c}{0.237} & 0.270   \\ \hline

\multicolumn{1}{c|}{\multirow{5}{*}{ETTh2}} & 96 & \multicolumn{1}{c} {\textcolor{red}{\textbf{0.262}}} &  \textcolor{red}{\textbf{0.320}} &  \multicolumn{1}{c} {\textcolor{blue}{0.268}} &  \textcolor{blue}{0.328} & \multicolumn{1}{c}{0.285} & 0.342  & \multicolumn{1}{c|}{\multirow{5}{*}{Electricity}} & 96 & \multicolumn{1}{c} {\textcolor{red}{\textbf{0.130}}}  & {\textcolor{red}{\textbf{0.222}}}  & \multicolumn{1}{c} {\textcolor{blue}{0.131} }  & {\textcolor{blue}{0.224} }  & \multicolumn{1}{c}{0.139} & 0.238  \\

\multicolumn{1}{c|}{} & 192 & \multicolumn{1}{c} {\textcolor{blue}{0.331}} & {\textcolor{red}{\textbf{0.373}}} & \multicolumn{1}{c} {\textcolor{red}{\textbf{0.329}}} & {\textcolor{blue} {0.375}} & \multicolumn{1}{c}{0.354} & 0.389 & \multicolumn{1}{c|}{} & 192 & \multicolumn{1}{c} {\textcolor{blue}{0.154} }  & {\textcolor{blue}{0.243} }  & \multicolumn{1}{c} {\textcolor{red}{\textbf{0.152}}}  & {\textcolor{red}{\textbf{0.241}}} & \multicolumn{1}{c}{0.153} & 0.251  \\

\multicolumn{1}{c|}{} & 336 & \multicolumn{1}{c} {\textcolor{red}{\textbf{0.358}}} & {\textcolor{red}{\textbf{0.403}}} & \multicolumn{1}{c} {\textcolor{blue}{0.368}} & 0.409 & \multicolumn{1}{c}{0.373} & {\textcolor{blue}{0.407}} & \multicolumn{1}{c|}{} & 336 & \multicolumn{1}{c} {\textcolor{red}{\textbf{0.158}}}  &{\textcolor{red}{\textbf{0.245}}}  & \multicolumn{1}{c}{\textcolor{blue}{0.160} }  & {\textcolor{blue}{0.248} }  & \multicolumn{1}{c}{0.169} & 0.266 \\

\multicolumn{1}{c|}{} & 720 & \multicolumn{1}{c}{\textcolor{red}{\textbf{0.369}}} & {\textcolor{red}{\textbf{0.411}}} & \multicolumn{1}{c} {\textcolor{blue}{0.372}} & {\textcolor{blue}{0.420}} & \multicolumn{1}{c}{0.406} & 0.441  & \multicolumn{1}{c|}{} & 720 & \multicolumn{1}{c} {\textcolor{red}{\textbf{0.190}}}  &  {\textcolor{blue}{0.294}}  &  \multicolumn{1}{c}{\textcolor{red}{\textbf{0.192}}}  & {\textcolor{blue}{0.298}}  & \multicolumn{1}{c}{0.206} & 0.297 \\ 

\multicolumn{1}{c|}{} & avg & \multicolumn{1}{c}{\textcolor{red}{\textbf{0.330}}} & {\textcolor{red}{\textbf{0.376}}}& \multicolumn{1}{c} {\textcolor{blue}{0.334}} & {\textcolor{blue}{ 0.383 }}& \multicolumn{1}{c}{0.381} & 0.412 & \multicolumn{1}{c|}{} & avg & \multicolumn{1}{c} {\textcolor{red}{\textbf{0.158}}}  & {\textcolor{red}{\textbf{0.251}}}  &\multicolumn{1}{c} {\textcolor{red}{\textbf{0.158}}}  &{\textcolor{blue}{0.252}}  & \multicolumn{1}{c}{0.167} & {\textcolor{blue}{0.263}} \\ \hline

\multicolumn{1}{c|}{\multirow{5}{*}{ETTm1}} & 96 & \multicolumn{1}{c}{0.272} & {\textcolor{blue}{0.336}} &  \multicolumn{1}{c}{0.272} & {\textcolor{red}{\textbf{0.334}}} & \multicolumn{1}{c}{0.292} & 0.346 & \multicolumn{1}{c|}{\multirow{5}{*}{Traffic}} & 96 & \multicolumn{1}{c}{\textcolor{red}{\textbf{0.361}}}  & {\textcolor{red}{\textbf{0.248}}}  &  \multicolumn{1}{c} {\textcolor{blue}{0.362}}  & {\textcolor{red}{\textbf{0.248}}}  & \multicolumn{1}{c}{0.388} & {\textcolor{blue}{0.282}}   \\  

\multicolumn{1}{c|}{} & 192 & \multicolumn{1}{c}{0.310} & {\textcolor{red}{\textbf{0.357}}}  & \multicolumn{1}{c}{0.310} & {\textcolor{blue}{0.358}} & \multicolumn{1}{c}{0.332} & 0.372 & \multicolumn{1}{c|}{} & 192 & \multicolumn{1}{c} {\textcolor{blue}{0.376}}  & {\textcolor{blue}{0.250}}  & \multicolumn{1}{c}{\textcolor{red}{\textbf{0.374}}}  & {\textcolor{red}{\textbf{0.247}}}  & \multicolumn{1}{c}{0.407} & 0.290  \\ 

\multicolumn{1}{c|}{} & 336 & \multicolumn{1}{c} {\textcolor{red}{\textbf{0.351}}} & {\textcolor{red}{ \textbf{0.382} }} &  \multicolumn{1}{c} {\textcolor{blue}{0.352}} & {\textcolor{blue}{0.384}}  & \multicolumn{1}{c}{0.366} & 0.394 & \multicolumn{1}{c|}{} & 336 & \multicolumn{1}{c} {\textcolor{red}{\textbf{0.382}}}  & {\textcolor{red}{\textbf{0.268}}}  &  \multicolumn{1}{c}{\textcolor{blue}{0.385}}  & {\textcolor{blue}{0.271}} & \multicolumn{1}{c}{0.412} & 0.294  \\ 

\multicolumn{1}{c|}{} & 720 & \multicolumn{1}{c}{\textcolor{red}{\textbf{0.380}}} &  {\textcolor{red}{\textbf{0.409}}}  & \multicolumn{1}{c} {\textcolor{blue}{0.383}}& {\textcolor{blue}{0.411}} & \multicolumn{1}{c}{0.417} & 0.421  & \multicolumn{1}{c|}{} & 720 & \multicolumn{1}{c} {\textcolor{red}{\textbf{0.426}}}  & {\textcolor{red}{\textbf{0.281}}}  & \multicolumn{1}{c}{\textcolor{blue}{0.430}}  & {\textcolor{blue}{0.288}} & \multicolumn{1}{c}{0.450} & 0.312  \\

\multicolumn{1}{c|}{} & avg & \multicolumn{1}{c} {\textcolor{red}{\textbf{0.328}}} & {\textcolor{red}{\textbf{0.371}}}  & \multicolumn{1}{c} {\textcolor{blue}{0.329}} & {\textcolor{blue}{0.372}} & \multicolumn{1}{c}{0.388} & 0.403  & \multicolumn{1}{c|}{} & avg & \multicolumn{1}{c} {\textcolor{red}{\textbf{0.386}}}  & \textcolor{red}{\textbf{0.261}} & \multicolumn{1}{c}{\textcolor{blue}{0.388}}  & {\textcolor{blue}{0.264}}  & \multicolumn{1}{c}{0.414} & 0.294  \\ 
\hline

\multicolumn{1}{c|}{\multirow{5}{*}{ETTm2}} & 96 & \multicolumn{1}{c} {\textcolor{blue}{0.163}} & {\textcolor{blue}{0.254 }} &  \multicolumn{1}{c} {\textcolor{red}{\textbf{0.161}}}  & {\textcolor{red}{\textbf{0.253}}}  & \multicolumn{1}{c}{0.173} & 0.262  & \multicolumn{1}{c|}{\multirow{5}{*}{ILI}} & 24 & \multicolumn{1}{c}{\textcolor{blue}{1.301}}  & {\textcolor{blue}{0.739}}  & \multicolumn{1}{c} {\textcolor{red}{\textbf{1.285}}} & {\textcolor{red}{\textbf{0.727}}}  & \multicolumn{1}{c}{2.063} &  0.881   \\

\multicolumn{1}{c|}{} & 192 & \multicolumn{1}{c}{\textcolor{red}{\textbf{0.218}}}  & {\textcolor{red}{\textbf{0.290}}}  & \multicolumn{1}{c} {\textcolor{blue}{0.219 }}  & {\textcolor{blue}{0.293 }} & \multicolumn{1}{c}{0.229} & 0.301  & \multicolumn{1}{c|}{} & 36 & \multicolumn{1}{c} {\textcolor{blue}{1.409}}  & {\textcolor{blue}{0.824}}   & \multicolumn{1}{c} {\textcolor{red}{\textbf{1.404}}}  & {\textcolor{red}{\textbf{0.814}}}  & \multicolumn{1}{c}{1.868} &  0.892   \\

\multicolumn{1}{c|}{} & 336 & \multicolumn{1}{c}{0.271} & {\textcolor{red}{\textbf{0.328}}} & \multicolumn{1}{c}{0.271} & {\textcolor{blue}{0.329}}  & \multicolumn{1}{c}{0.286} & 0.341   & \multicolumn{1}{c|}{} & 48 & \multicolumn{1}{c} {\textcolor{red}{\textbf{1.520}}}  & {\textcolor{red}{\textbf{0.807}}}   & \multicolumn{1}{c}{\textcolor{blue}{1.523}}  & {\textcolor{red}{\textbf{0.807}}}  & \multicolumn{1}{c}{1.790} & {\textcolor{blue}{ 0.884}}  \\ 

\multicolumn{1}{c|}{} & 720 & \multicolumn{1}{c}{\textcolor{blue}{0.356}}  & {\textcolor{blue}{0.382 }}  & \multicolumn{1}{c} {\textcolor{red}{\textbf{0.352}}} & {\textcolor{red}{\textbf{0.379}}} & \multicolumn{1}{c}{0.378} &  0.401  & \multicolumn{1}{c|}{} & 60 & \multicolumn{1}{c} {\textcolor{red}{\textbf{1.529}}}  & {\textcolor{red}{\textbf{0.852}}}  &\multicolumn{1}{c}{\textcolor{blue}{1.531}}  & {\textcolor{blue}{0.854}}  & \multicolumn{1}{c}{1.979} & 0.957 \\

\multicolumn{1}{c|}{} & avg & \multicolumn{1}{c} {\textcolor{blue}{0.252 }} & 0.316 & \multicolumn{1}{c} {\textcolor{red}{\textbf{0.251}}}  & {\textcolor{red}{\textbf{0.313}}} & \multicolumn{1}{c}{0.284} & {\textcolor{blue}{0.339 }}    & \multicolumn{1}{c|}{} & avg & \multicolumn{1}{c} {\textcolor{blue}{1.439}}  & {\textcolor{blue}{ 0.805}} & \multicolumn{1}{c} {\textcolor{red}{\textbf{1.435}}}  & {\textcolor{red}{ \textbf{0.801}}} & \multicolumn{1}{c}{1.925} & 0.903  \\\hline

\multicolumn{2}{c|}{Avg. Training Time} & \multicolumn{2}{c|}  {\textcolor{red}{\textbf{3.150 Minutes}}} & \multicolumn{2}{c|} {55.41 Minutes}  & \multicolumn{2}{c|}{\textcolor{blue}{14.76  Minutes}} & \multicolumn{2}{c|}{Avg. Training Time} & \multicolumn{2}{c|}  {\textcolor{red}{\textbf{36.52 Minutes}}} & \multicolumn{2}{c|} {510.2 Minutes}   & \multicolumn{2}{c}{\textcolor{blue}{146.3 Minutes}} \\

\multicolumn{2}{c|}{Avg. Occupied GPU} & \multicolumn{2}{c|}  {\textcolor{red}{\textbf{2.033 GB}}} & \multicolumn{2}{c|} {74.42 GB}  & \multicolumn{2}{c|}{\textcolor{blue}{16.31 GB}} & \multicolumn{2}{c|}{Avg. Occupied GPU} & \multicolumn{2}{c|}  {\textcolor{red}{\textbf{2.110 GB}}} & \multicolumn{2}{c|} {77.271 GB}  & \multicolumn{2}{c}{\textcolor{blue}{17.680  GB}} \\ \hline

\end{tabular}%

\caption{ Multivariate time-series forecasting (low chaos)  comparison with pre-trained LLM model. `Avg. Training Time' is the average training time for the model. `Avg. Occupied GPU' is the average GPU memory usage for the model.}
\label{tab:side-by-side-tables}

\end{table*}

\begin{table}[!h]
\centering
% Changing the font size
%\scalebox{0.73}{
%\resizebox{\textwidth}{!}{%
\setlength{\tabcolsep}{3pt}
\renewcommand{\arraystretch}{1.1}
\fontsize{7.6pt}{7.6pt}\selectfont
%\scriptsize 
\begin{tabular}{l|cc|cc|cc|cc|cc||cc|cc} 
\hline 
 Method & \multicolumn{2}{c|}{Air Quality} & \multicolumn{2}{c|}{DWV} & \multicolumn{2}{c|}{DGP} & \multicolumn{2}{c|}{DDF} & \multicolumn{2}{c||}{BTC} & \multicolumn{2}{c|}{AW}  & \multicolumn{2}{c}{TR} \\
\hline
Metric & MSE  & MAE & MSE & MAE & MSE & MAE & MSE & MAE & MSE & MAE &  Acc. & F-Score & Acc. & F-Score \\\hline
Transformer & 89.53 & 11.34  & 42.95  & 5.181& 168.6K & 353.3 & 1014 & 22.27   & 79.40K & 232.1 & 0.928 & 0.940 & 0.649 & 0.934 \\
RT (Ours)  &  \textcolor{red}{\textbf{70.58 }}& \textcolor{red}{\textbf{9.040}}  & \textcolor{red}{\textbf{6.181}} & \textcolor{red}{\textbf{1.893}} & \textcolor{red}{\textbf{28.79K}} & \textcolor{red}{\textbf{119.2}} & \textcolor{red}{\textbf{10.17}} &  \textcolor{red}{\textbf{410.6}} & \textcolor{red}{\textbf{211.6}} & \textcolor{red}{\textbf{211.6}}  & \textcolor{red}{\textbf{0.899}}& \textcolor{red}{\textbf{0.901}}  & \textcolor{red}{\textbf{0.802}} & \textcolor{red}{\textbf{0.975}}\\
\hline \\
\end{tabular}
%}
\caption{\small  Time series comparison (medium chaos) in MSE, MAE, Accuracy, and F-score with RT and baseline Transformer. `Acc.' is the Accuracy.}
\vspace{-2.5em}
\label{table:result-mchao}
\end{table}

\textbf{Signature Feature Extraction:}
Table~\ref{tab:entropy} shows the entropy of extracted features by baseline ($\mathbf{h}_{t}$), reservoir ($\mathbf{z}_t$), and RT ($\mathbf{f}_t$). Our RT model extracts more informative features than the baseline Transformer model. Using the Shannon entropy to measure the information amount, our reservoir output features $\mathbf{z}_t$ have higher entropy (more informative) than the embedding output features  $\mathbf{h}_{t}$ in the baseline Transformer. Combining both $\mathbf{f}_t$ gives the highest entropy among them, demonstrating more information extraction with our RT. Details of the experiments are enclosed in the Appendix~\ref{sec:more_sing_feature}.

\begin{table*}[!ht]
    \centering
    \begin{minipage}[t]{.6\textwidth}
        \centering
        \scalebox{.75}{
        \begin{tabular}{l|lll|lll|lll} 
\hline
          Method   & \multicolumn{3}{c|}{RT (Ours)} & \multicolumn{3}{c|}{DLinear}& \multicolumn{3}{c}{PatchTST}\\

\hline 
 Horizons & MGS & ECG & LA & MGS & ECG & LA  & MGS & ECG & LA \\
\hline
50 & {\textcolor{red}{\textbf{0.173}}} & {\textcolor{blue}{0.301}}   & {\textcolor{blue}{0.313}} 
  & 0.204 & 0.321 & 0.356  &{\textcolor{blue}{0.179}}&{\textcolor{red}{\textbf{0.297}}}& {\textcolor{red}{\textbf{0.310}}} \\
100 & {\textcolor{red}{\textbf{0.185}}}  & {\textcolor{red}{\textbf{0.314}}}  & {\textcolor{blue}{0.331}} & 0.225 & 0.343 & 0.370 &{\textcolor{blue}{0.186}}& {\textcolor{blue}{ 0.315}}& {\textcolor{red}{\textbf{0.313}}}  \\
500 &{\textcolor{red}{\textbf{0.206}}}  & {\textcolor{red}{0.341}}  & {\textcolor{red}{\textbf{0.359}}}  & 0.240 &0.371& 0.389 &{\textcolor{blue}{0.209}}& {\textcolor{blue}{0.347}} & {\textcolor{blue}{0.360}} \\
1000 & {\textcolor{red}{\textbf{0.225}}}  & {\textcolor{red}{\textbf{0.374}}} & {\textcolor{red}{\textbf{0.447}}}  & 0.250 & 0.390& 0.529  &{\textcolor{blue}{0.228}}& {\textcolor{blue}{0.378}} & {\textcolor{blue}{0.449}}  \\
\hline 
\end{tabular}}
        \caption{MSE on three typical chaos datasets (high chaos). }
        \label{figc:sub4}
    \end{minipage}%
    \hfill
    \begin{minipage}[t]{.32450\textwidth}
        \centering
       \scalebox{.75}{
        \begin{tabular}{l|lll}
            \hline
             \multicolumn{4}{c}{Entropy}\\\hline
            Datasets & $\mathbf{h}_t$ & $\mathbf{z}_t$ & $\mathbf{f}_t$ \\
            \hline
            ETTh1 & 1.011 & \textcolor{blue}{1.966} &\textcolor{red}{\textbf{2.331}} \\ 
            ETTh2 & 0.993 & \textcolor{blue}{1.892} & \textcolor{red}{\textbf{2.015}}\\
            ETTm1 & 0.816 & \textcolor{blue}{2.056} & \textcolor{red}{\textbf{2.831}}\\
            ETTm2 &0.599 & \textcolor{blue}{1.248} & \textcolor{red}{\textbf{2.131}}\\
            \hline
        \end{tabular}}
        \caption{Feature entropy.} \label{tab:entropy}
    \end{minipage}
\vspace{-1em}    
\end{table*}

\subsection{Ablation Analysis}

\begin{figure*}[!ht]
    \centering
    \begin{minipage}[t]{.32450\textwidth}
        \centering
        \includegraphics[width=0.9950\linewidth]{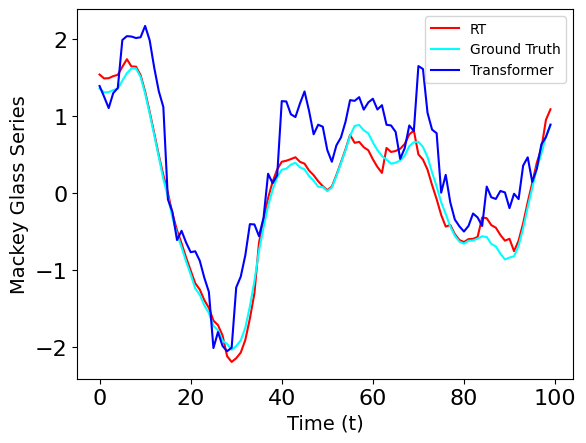}
        \caption*{ Mackey Glass Series.}
        \label{figc:sub4}
    \end{minipage}%
    \hfill
    \begin{minipage}[t]{.33450\textwidth}
        \centering
        \includegraphics[width=0.9950\linewidth]{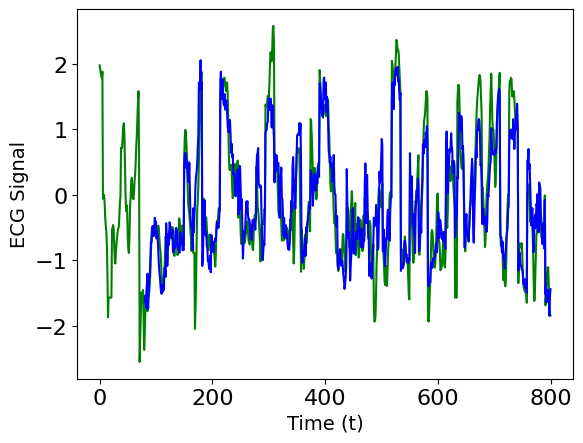}
        \caption*{ ECG Signal.}
        \label{figc:sub5}
    \end{minipage}
    \hfill
    \begin{minipage}[t]{.33450\textwidth}
        \centering
        \includegraphics[width=0.9950\linewidth]{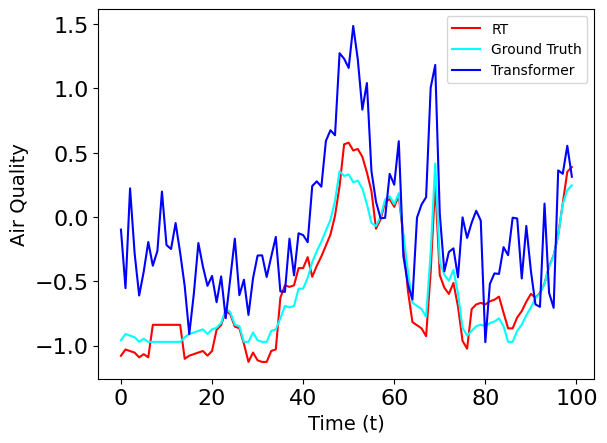}
        \caption*{Air Quality.}
        \label{figc:sub6}
    \end{minipage}
    \caption{The figures represent chaotic dataset examples. The cyan line  is ground truth; the red line is the prediction using RT; and the blue line is the prediction using  Transformer.}
    \label{fig:chaos_examples}
\end{figure*}

\textbf{Output Examples: }
Figure~\ref{fig:attentionreservoir} (Right) and Figure~\ref{fig:chaos_examples} plots the examples from three datasets. We show that our RT method output is more close to the ground truth than the Transformer baseline.

\begin{figure}[!h]
\centering
\begin{minipage}{.45\textwidth}
\hspace{-1cm}
  \centering
  \includegraphics[width=1.1472\linewidth]{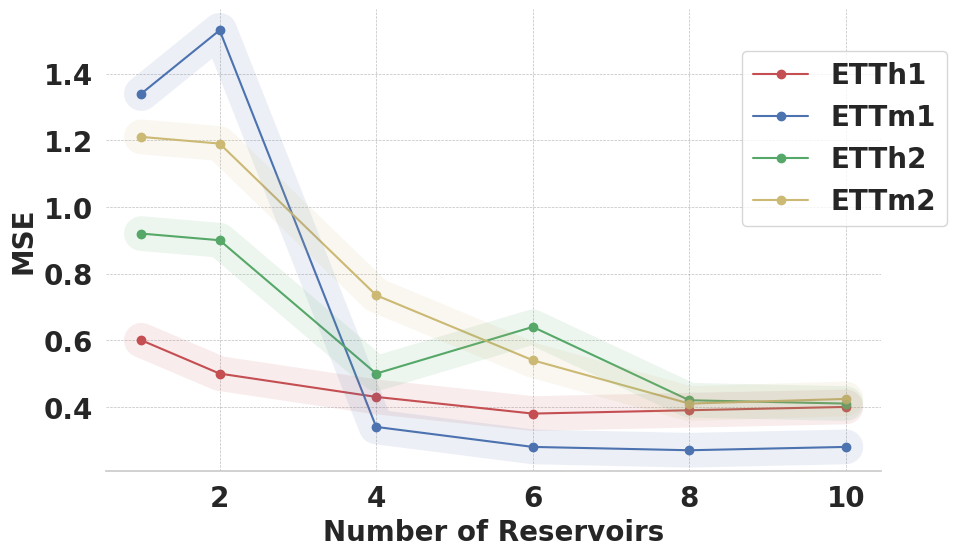}
  \captionof{figure}{MSE vs. reservoir numbers.}
  \label{fig:tradeoff}
\end{minipage}%
\begin{minipage}{.1\textwidth}
\end{minipage}
\begin{minipage}{.45\textwidth}
  \centering
  \includegraphics[width=1.0\linewidth]{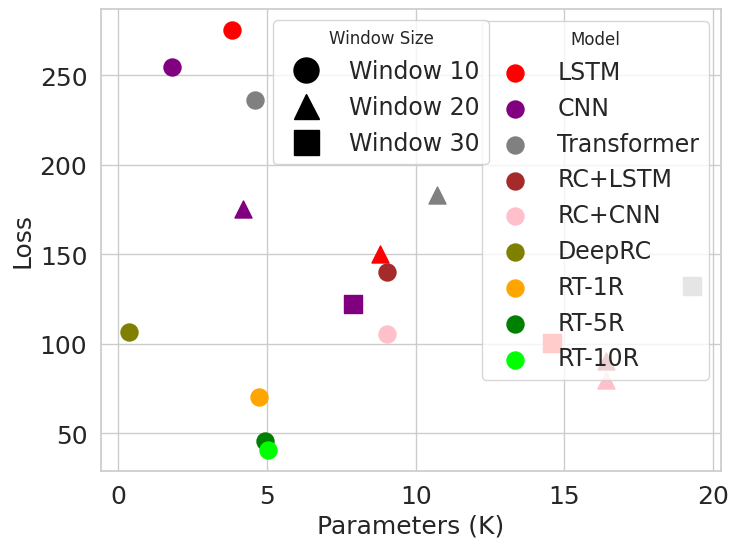}
  \captionof{figure}{RT outperforms baselines with small parameter size and loss in training.  }%In the presented graph, the X-axis represents the size of the model parameters in thousands, while the Y-axis illustrates the mean absolute error of the prediction outputs. The yellow boxes in the graph indicate the history lengths used in the computation of the DRC-1R, DRC-5R, and DRC-10R models, which were each constructed with one, five, and ten reservoirs, respectively.}
  \label{fig:parameter}
\end{minipage}
\vspace{-2em}
\end{figure}

\textbf{Impact of Reservoir Count on Performance:}
Figure~\ref{fig:tradeoff} illustrates the relationship between the number of ensembling reservoirs and system performance. Increasing the number of reservoirs improves the performance, leading to a reduction in loss until the loss converges for $10$ reservoirs.

\textbf{Model Parameter Size versus Loss:}
Conventional models require a large number of training parameters for long historical sequences.
In contrast, RT only train a vector representation (i.e. readout) for the next time step without explicit conditioning on the entire history.  Figure~\ref{fig:parameter} illustrates that RT provides a better trade-off between the number of training parameters and loss.

\iffalse
\begin{table}[ht]
\centering

\resizebox{\textwidth}{!}{%
\begin{tabular}{@{}l|cc|cc|cc|cc|cc|cc@{}}
\toprule
\multirow{2}{*}{Initialization}
 & \multicolumn{2}{c|}{ETTh1} & \multicolumn{2}{c|}{ETTh2} & \multicolumn{2}{c|}{ETTm1} & \multicolumn{2}{c|}{ETTm2}& \multicolumn{2}{c|}{Weather}& \multicolumn{2}{c}{Traffic} \\ 
\cline{2-13} 
               & Non-Group & Group & Non-Group & Group  & Non-Group  & Group  & Non-Group  & Group& Non-Group  & Group & Non-Group  & Group  \\ 
\hline
Random         & 0.543   & 0.402 & 0.435   & 0.342 & 0.447   & 0.328 & 0.301   & 0.228  & 0.334 & 0.198 & 0.424 &0.380\\
Zero           & 0.429   & 0.396 & 0.368   & 0.331 & 0.348   & 0.310 & 0.236   & 0.218  & 0.215 & 0.193 & 0.380&0.376\\
Constant       & 0.673   & 0.405 & 0.420   & 0.333 & 0.516   & 0.319 & 0.265   & 0.225  & 0.204 & 0.196 & 0.424&0.378\\
Normal         & 0.529   & 0.401 & 0.445   & 0.335 & 0.435   & 0.332 & 0.274   & 0.236  & 0.242 & 0.198 & 0.452&0.379\\
Uniform        & 0.432   & 0.399 & 0.395   & 0.331 & 0.401   & 0.314 & 0.232   & 0.219  & 0.200 & 0.195 & 0.401&0.375\\ \hline
Variance       & $8 \times 10^{-2}$ & $9\times 10^{-6}$ & $7\times 10^{-3}$ &  $1\times 10^{-6}$  & $ 3\times 10^{-2} $  & $6\times 10^{-5}$ & $6\times 10^{-3}$   & $4\times 10^{-5}$ & $2\times 10^{-2}$   & $3\times 10^{-6}$& $3\times 10^{-3}$   & $3\times 10^{-6}$\\
\bottomrule
\end{tabular}
}
\caption{MSE on different initialization using single reservoir and  group reservoirs. The prediction results are not sensitive to the initializations without using reservoirs. }

\end{table}
\fi

\begin{table}[ht]
\centering

\setlength{\tabcolsep}{3pt}
\renewcommand{\arraystretch}{1.1}
\fontsize{6.8pt}{6.8pt}\selectfont
%\setlength{\tabcolsep}{1pt}
%\renewcommand{\arraystretch}{1.1}
%\fontsize{6pt}{6pt}\selectfont
\begin{tabular}{@{}l|cc|cc|cc|cc|cc|cc@{}}
\toprule
\multirow{2}{*}{Initialization}
 & \multicolumn{2}{c|}{ETTh1} & \multicolumn{2}{c|}{ETTh2} & \multicolumn{2}{c|}{ETTm1} & \multicolumn{2}{c|}{ETTm2}& \multicolumn{2}{c|}{Weather}& \multicolumn{2}{c}{Traffic} \\ 
\cline{2-13} 
               & NG & G & NG & G  & NG  & G  & NG  & G& NG  & G & NG  & G  \\ 
\hline
Random         & 0.543   & 0.402 & 0.435   & 0.342 & 0.447   & 0.328 & 0.301   & 0.228  & 0.334 & 0.198 & 0.424 &0.380\\
Zero           & 0.429   & 0.396 & 0.368   & 0.331 & 0.348   & 0.310 & 0.236   & 0.218  & 0.215 & 0.193 & 0.380&0.376\\
Constant       & 0.673   & 0.405 & 0.420   & 0.333 & 0.516   & 0.319 & 0.265   & 0.225  & 0.204 & 0.196 & 0.424&0.378\\
Normal         & 0.529   & 0.401 & 0.445   & 0.335 & 0.435   & 0.332 & 0.274   & 0.236  & 0.242 & 0.198 & 0.452&0.379\\
Uniform        & 0.432   & 0.399 & 0.395   & 0.331 & 0.401   & 0.314 & 0.232   & 0.219  & 0.200 & 0.195 & 0.401&0.375\\ \hline
%$t$-value / $p$-value       & \multicolumn{2}{c|}{2.52 / 0.03}  & \multicolumn{2}{c|}{3.54 / 0.01} &  \multicolumn{2}{c|}{2.85 / 0.02}  & \multicolumn{2}{c|}{2.58 / 0.03}   & \multicolumn{2}{c|}{1.68 / 0.13}   & \multicolumn{2}{c}{2.57 / 0.03}\\
%$F$-test \& $p$-value       & \multicolumn{2}{c|}{49.38 / 0.001}  & \multicolumn{2}{c|}{1.34 / 0.39} &  \multicolumn{2}{c|}{2.91 / 0.16}  & \multicolumn{2}{c|}{17.21 / 0.009}   & \multicolumn{2}{c|}{441.29 / 0.000015}   & \multicolumn{2}{c}{6.96 / 0.04}\\
$p$-value & 0.249 & $2.31 \blank 10^{-6} $ & 0.049 & $4.36 \blank 10^{-5}$ & 0.120 & $1.94\blank 10^{-4}$ & 0.226 & $5\blank 10^{-3}$ & 0.436 & $2.2 \blank 10^{-5}$ & 0.187 & $3.15 \blank 10^{-5}$\\
\bottomrule
\end{tabular}
\vspace{.4cm}
\caption{MSE on different initialization using single reservoir (NG) and  group reservoirs (G). Group reservoir eliminates the prediction error sensitivity on the initialization.}
\vspace{-2.5em}
\label{tab:init_exp}
\end{table}

\begin{table*}[h]

\centering
%\setlength{\tabcolsep}{3pt}
%\renewcommand{\arraystretch}{1.1}
%\fontsize{8pt}{8pt}\selectfont
\setlength{\tabcolsep}{3pt}
\renewcommand{\arraystretch}{1.1}
\fontsize{7.8pt}{7.8pt}\selectfont
\begin{tabular}{l|c|c||c|c|c}
\hline
Model            & Memory (Cache) Footprint            & Context Length  & Model & Time Complexity & Seq. Ops.   \\  \hline
Transformer-XL          & \((d_{key} + d_{value}) \cdot H \cdot k \cdot ly\) & \(k \cdot ly\)  & Transformer   & \(O(T^2 \cdot d_{\epsilon})\)            & \(O(1)\)           \\ 
Compressive Transformer    & \(d_{\epsilon} \cdot (c + T) \cdot ly\)           & \((c \cdot r + k) \cdot ly\) & PatchTST                     & \(O((T/P)^2 \cdot d_{\epsilon})\)             & \(O(1)\)        \\ 
Memorizing Transformers   & \((d_{key} + d_{value}) \cdot H \cdot k \cdot T\)  & T & Informer                   & \(O(T \cdot log(T) \cdot d_{\epsilon})\)            & \(O(1)\)           \\ 
RMT                      & \(d_{\epsilon} \cdot p \cdot ly \cdot 2\)        & T   & Autoformer                   & \(O(T \cdot log(T) \cdot d_{\epsilon})\)            & \(O(1)\)                  \\ 
AutoCompressors          & \(d_{\epsilon} \cdot p \cdot (v + 1) \cdot ly\)  & T      & Reformer                   & \(O(T \cdot log(T) \cdot d_{\epsilon})\)            & \(O(1)\)            \\ 
Infini-Transformers      & \(d_{key} \cdot (d_{value} + 1) \cdot H \cdot ly\) & T      & RNN                           & \(O(T \cdot d_{\epsilon}^2)\)            & \(O(T)\)                   \\  %\hline
Reservoir     &  \( N_r^2 \) & T &  Reservoir   &  \( O(T\cdot N_r^2) \) &    \(O(T)\)   \\ \hline
RT (Ours)     & \( N_r^2 \cdot L \) & T      & RT (Ours)    & \(O(T\cdot k^2 \cdot d_{\epsilon})\)  & \(O(T)\)       \\ \hline
\end{tabular}

\caption{Comparison of memory (cache) footprint and time complexity. }
\vspace{-1.5em}
\label{tab:mem_cmp}
\end{table*}

\textbf{Initialization Comparison:} Table \ref{tab:init_exp} shows how different initialization affects the performance for each of the datasets. We compute the the $p$-value to see whether the variation in performance due to different initialization is significant when comparing a single reservoir (Non-Group denoted with `NG') with an group reservoir (Group denoted with `G'). We use the combined mean for each dataset as the population mean to compute the $p$-values. We observe significant differences for all the datasets for only the Non-Group experiments. 

\textbf{Complexity Comparison:} Table \ref{tab:mem_cmp} shows the memory (cache) footprint and time complexity of different models. `Seq. Ops.' is the `Sequential Operations', and the notations are as follows. \(N_r\) is the reservoir state; \(L\) is the number of reservoirs in the group reservoir; \(k\) is the short-term context window length; \(T\) is the  long-term context total input length; \(ly\) is the number of layers; \(H\) is the number of attention heads; \(c\) is the Compressive Transformer memory size; \(r\) is the compression ratio; \(p\) is the number of soft-prompt summary vectors; \(v\) is the summary vector accumulation steps; $s$ is the kernel size; $P$ is the patch size; $d_{\epsilon}$ is the embedding model dimension; $d_{key}$ is the dimension of keys in attention; and $d_{value}$ is the dimension of values in attention.
RT maintains a constant memory complexity of $N_r^2\cdot L$, and the time complexity is linear regarding total input length. 
Additionally, RT updates its memory with each new input, resulting in retrieval and update occurring once per input. Other methods like Transformer-XL \citep{dai2019attentive}, Compressive Transformer \citep{rae2019compressive}, Memorizing Transformers \citep{wu2022memorizing}, RMT \citep{bulatov2022recurrent}, and AutoCompressors \citep{chevalier2023adapting} do not have constant memory usage; it depends on the total input segment length $N$ and still exhibits a large context-memory complexity. 
Furthermore, RMT \citep{bulatov2022recurrent} and AutoCompressors \citep{chevalier2023adapting} allow for a potentially infinite context length since they compress the input into summary vectors and then pass them as extra soft-prompt inputs for the subsequent segments. However, in practice, the success of those techniques highly depends on the size of soft-prompt vectors \citep{munkhdalai2024leave}.

\section{Related Work}
Time-series prediction using machine learning approaches has been investigated in the last decade using CNN, WNN, FNN, LSTM, etc.~\citep{ramadevi2022chaotic,sangiorgio2020robustness,karasu2022crude,sahoo2019long,timeseriespaperwithcodes} on datasets with chaotic properties. Recently, Transformer-based solutions have shown success for long-term time series forecasting (LTSF)~\citep{mohammadi2020transformer,huang2019dsanet}. 
Recently, researchers have also used various methods to improve long-term time series forecasting ~\citep{zhou2022fedformer,zhou2021informer,wu2021autoformer}. 
Nonetheless, the Transformer is expensive in terms of quadratic time and space complexity of the input length. Important works have been proposed in improving Transformer architectures and their complexity in long context in various ways, however, compressing the long input sequence into fixed length often misses the dependency of the temporal events.  

Researchers have also shown how using pre-trained LLMs for time-series data improves performance compared to training from scratch \citep{jin2023time,chang2024llm4ts,huang2024enhancing,munkhdalai2024leave}. \cite{de2023llm} used a combination of LLMs (LLaMA-2 and Zephyr-7b-$\alpha$) along with Multimodal LLM (LLaVA) for time-series forecasting.  However, there is still a restriction on the input length due to the complexity of the neural network models. 
Note that we will experimentally compare with \citep{munkhdalai2024leave} once their code releases. 

%While reservoirs have found extensive application in time-series tasks, their performance often lags behind the current state-of-the-art outcomes (Gauthier, 2021). 
Reservoir networks with their simplified training and superior performance for time-series forecasting compared to the traditional models, including time-series kernels, feed-forward networks, and RNN models, has been well documented. Researchers have also compared different types of reservoir networks for chaotic time-series data \citep{bo2020asynchronously}. However, the reservoir network fails to effectively handle long context time series data within fixed-size reservoir~\citep{lukovsevivcius2009reservoir,kim2020time,mansoor2021comparison,bianchi2020reservoir}. We have introduced reservoir computing to solve the input length problem. 
A concept of reservoir Transformer ~\citep{shen2020reservoir,gauthier2021next} was introduced by interspersing the reservoir layers into the regular Transformer layers and showing improvements in wall-clock compute time. 
These models improve the training efficiency by using parallelization of time-series methods ~\citep{bollt2021explaining}. 
However, such integration will reduce the feature extraction power of both reservoirs in the long sequences and Transformer in short sequences. We do not combine the reservoir and Transformer linearly but in a non-linear way and receive significant improvements over all previous methods in time-series prediction, including chaotic prediction. 
Additionally, there has been research on using ensemble of models for time-series prediction. Researchers have proposed ensemble methods \citep{feng2019chaotic} that combine multiple hybrid models using wavelet transform for chaotic time-series prediction and achieved improved performance on several chaotic systems.\citep{tang2020hybrid, oliveira2015ensembles, bertsimas2023ensemble} proposed an ensemble method that combines multiple deep-learning models for time-series prediction and showed improved performance on several datasets. In our work, we propose a novel method to ensemble multiple reservoir networks and combine them with Transformer architecture, contributing to improved performance in time-series prediction. %To sum up, the utilization of machine learning methods for predicting chaotic time series has gained significant traction recently. Diverse neural network architectures and ensemble techniques have emerged to enhance predictive capabilities. Our novel approach, involving the fusion of reservoir computing and Transformers, holds promise for substantial advancements in chaotic time-series prediction tasks. This approach presents a unique avenue that holds the promise of substantially enhancing performance in chaotic time-series prediction tasks.

\section{Conclusion}
Our RT has the unique advantage of handling any input length and producing robust prediction outputs. Thus, it is ideal for time-series forecasting, including event prediction in chaos. 
We introduce nonlinear readouts to our Reservoir Transformer that outperform  state-of-the-art models in terms of accuracy, keeping reasonable  complexity in space and time. Furthermore, we testify to long-horizon forecasting and show that RT makes the far-horizon prediction less sensitive to initialization. 
This research direction has the potential for succession works by integrating chaotic analysis, deep neural networks, and reservoir computing to understand and predict our environments better.

\bibliographystyle{unsrtnat}
\bibliography{output}

\begin{thebibliography}{70}
\providecommand{\natexlab}[1]{#1}
\providecommand{\url}[1]{\texttt{#1}}
\expandafter\ifx\csname urlstyle\endcsname\relax
  \providecommand{\doi}[1]{doi: #1}\else
  \providecommand{\doi}{doi: \begingroup \urlstyle{rm}\Url}\fi

\bibitem[Liu(2010)]{liunonlinear}
Zonghua Liu.
\newblock Nonlinear time series: Computations and applications, 2010.

\bibitem[Jordan and Smith(2007)]{jordan2007nonlinear}
Dominic Jordan and Peter Smith.
\newblock \emph{Nonlinear ordinary differential equations: an introduction for scientists and engineers}.
\newblock OUP Oxford, 2007.

\bibitem[Wu et~al.(2024)Wu, Yang, Song, and Li]{wu2024generalized}
Xiang Wu, Xujun Yang, Qiankun Song, and Chuandong Li.
\newblock Generalized lyapunov stability theory of continuous-time and discrete-time nonlinear distributed-order systems and its application to boundedness and attractiveness for networks models.
\newblock \emph{Communications in Nonlinear Science and Numerical Simulation}, 128:\penalty0 107664, 2024.

\bibitem[Zeng et~al.(2022)Zeng, Chen, Zhang, and Xu]{zeng2022transformers}
Ailing Zeng, Muxi Chen, Lei Zhang, and Qiang Xu.
\newblock Are transformers effective for time series forecasting?
\newblock \emph{arXiv preprint arXiv:2205.13504}, 2022.

\bibitem[Vaswani et~al.(2017)Vaswani, Shazeer, Parmar, Uszkoreit, Jones, Gomez, Kaiser, and Polosukhin]{vaswani2017attention}
Ashish Vaswani, Noam Shazeer, Niki Parmar, Jakob Uszkoreit, Llion Jones, Aidan~N Gomez, {\L}ukasz Kaiser, and Illia Polosukhin.
\newblock Attention is all you need.
\newblock \emph{Advances in neural information processing systems}, 30, 2017.

\bibitem[Bertsch et~al.(2023)Bertsch, Alon, Neubig, and Gormley]{bertsch2023unlimiformer}
Amanda Bertsch, Uri Alon, Graham Neubig, and Matthew~R Gormley.
\newblock Unlimiformer: Long-range transformers with unlimited length input.
\newblock \emph{arXiv preprint arXiv:2305.01625}, 2023.

\bibitem[Tay et~al.(2022)Tay, Dehghani, Bahri, and Metzler]{tay2022efficient}
Yi~Tay, Mostafa Dehghani, Dara Bahri, and Donald Metzler.
\newblock Efficient transformers: A survey.
\newblock \emph{ACM Computing Surveys}, 55\penalty0 (6):\penalty0 1--28, 2022.

\bibitem[Kitaev et~al.(2020)Kitaev, Kaiser, and Levskaya]{kitaev2020reformer}
Nikita Kitaev, {\L}ukasz Kaiser, and Anselm Levskaya.
\newblock Reformer: The efficient transformer.
\newblock \emph{arXiv preprint arXiv:2001.04451}, 2020.

\bibitem[Bollt(2021)]{bollt2021explaining}
Erik Bollt.
\newblock On explaining the surprising success of reservoir computing forecaster of chaos? the universal machine learning dynamical system with contrast to var and dmd.
\newblock \emph{Chaos: An Interdisciplinary Journal of Nonlinear Science}, 31\penalty0 (1), 2021.

\bibitem[Jin et~al.(2024)Jin, Wang, Ma, Chu, Zhang, Shi, Chen, Liang, Li, Pan, and Wen]{jin2024timellm}
Ming Jin, Shiyu Wang, Lintao Ma, Zhixuan Chu, James~Y. Zhang, Xiaoming Shi, Pin-Yu Chen, Yuxuan Liang, Yuan-Fang Li, Shirui Pan, and Qingsong Wen.
\newblock Time-{LLM}: Time series forecasting by reprogramming large language models.
\newblock In \emph{The Twelfth International Conference on Learning Representations}, 2024.
\newblock URL \url{https://openreview.net/forum?id=Unb5CVPtae}.

\bibitem[Nie et~al.(2022)Nie, Nguyen, Sinthong, and Kalagnanam]{nie2022time}
Yuqi Nie, Nam~H Nguyen, Phanwadee Sinthong, and Jayant Kalagnanam.
\newblock A time series is worth 64 words: Long-term forecasting with transformers.
\newblock \emph{arXiv preprint arXiv:2211.14730}, 2022.

\bibitem[Beltagy et~al.(2020)Beltagy, Peters, and Cohan]{beltagy2020longformer}
Iz~Beltagy, Matthew~E Peters, and Arman Cohan.
\newblock Longformer: The long-document transformer.
\newblock \emph{arXiv preprint arXiv:2004.05150}, 2020.

\bibitem[Kim et~al.(2021)Kim, Kim, Tae, Park, Choi, and Choo]{kim2021reversible}
Taesung Kim, Jinhee Kim, Yunwon Tae, Cheonbok Park, Jang-Ho Choi, and Jaegul Choo.
\newblock Reversible instance normalization for accurate time-series forecasting against distribution shift.
\newblock In \emph{International Conference on Learning Representations}, 2021.

\bibitem[Zhou et~al.(2021)Zhou, Zhang, Peng, Zhang, Li, Xiong, and Zhang]{zhou2021informer}
Haoyi Zhou, Shanghang Zhang, Jieqi Peng, Shuai Zhang, Jianxin Li, Hui Xiong, and Wancai Zhang.
\newblock Informer: Beyond efficient transformer for long sequence time-series forecasting.
\newblock In \emph{Proceedings of the AAAI conference on artificial intelligence}, pages 11106--11115, 2021.

\bibitem[Wu et~al.(2021)Wu, Xu, Wang, and Long]{wu2021autoformer}
Haixu Wu, Jiehui Xu, Jianmin Wang, and Mingsheng Long.
\newblock Autoformer: Decomposition transformers with auto-correlation for long-term series forecasting.
\newblock \emph{Advances in Neural Information Processing Systems}, 34:\penalty0 22419--22430, 2021.

\bibitem[Liu et~al.(2021)Liu, Yu, Liao, Li, Lin, Liu, and Dustdar]{liu2021pyraformer}
Shizhan Liu, Hang Yu, Cong Liao, Jianguo Li, Weiyao Lin, Alex~X Liu, and Schahram Dustdar.
\newblock Pyraformer: Low-complexity pyramidal attention for long-range time series modeling and forecasting.
\newblock In \emph{International Conference on Learning Representations}, 2021.

\bibitem[Zhang and Yan(2022)]{zhang2022crossformer}
Yunhao Zhang and Junchi Yan.
\newblock Crossformer: Transformer utilizing cross-dimension dependency for multivariate time series forecasting.
\newblock In \emph{The eleventh international conference on learning representations}, 2022.

\bibitem[Rombach et~al.(2022)Rombach, Blattmann, Lorenz, Esser, and Ommer]{rombach2022high}
Robin Rombach, Andreas Blattmann, Dominik Lorenz, Patrick Esser, and Bj{\"o}rn Ommer.
\newblock High-resolution image synthesis with latent diffusion models.
\newblock In \emph{Proceedings of the IEEE/CVF conference on computer vision and pattern recognition}, pages 10684--10695, 2022.

\bibitem[Wei et~al.(2020)Wei, Zhang, Li, Zhang, and Wu]{wei2020multi}
Xi~Wei, Tianzhu Zhang, Yan Li, Yongdong Zhang, and Feng Wu.
\newblock Multi-modality cross attention network for image and sentence matching.
\newblock In \emph{Proceedings of the IEEE/CVF conference on computer vision and pattern recognition}, pages 10941--10950, 2020.

\bibitem[Jaeger and Haas(2004)]{jaeger2004harnessing}
Herbert Jaeger and Harald Haas.
\newblock Harnessing nonlinearity: Predicting chaotic systems and saving energy in wireless communication.
\newblock \emph{science}, 304\penalty0 (5667):\penalty0 78--80, 2004.

\bibitem[Luko{\v{s}}evi{\v{c}}ius and Jaeger(2009)]{lukovsevivcius2009reservoir}
Mantas Luko{\v{s}}evi{\v{c}}ius and Herbert Jaeger.
\newblock Reservoir computing approaches to recurrent neural network training.
\newblock \emph{Computer science review}, 3\penalty0 (3):\penalty0 127--149, 2009.

\bibitem[Verstraeten et~al.(2007)Verstraeten, Schrauwen, d’Haene, and Stroobandt]{verstraeten2007experimental}
David Verstraeten, Benjamin Schrauwen, Michiel d’Haene, and Dirk Stroobandt.
\newblock An experimental unification of reservoir computing methods.
\newblock \emph{Neural networks}, 20\penalty0 (3):\penalty0 391--403, 2007.

\bibitem[Ti{\v{n}}o et~al.(2007)Ti{\v{n}}o, Hammer, and Bod{\'e}n]{tivno2007markovian}
Peter Ti{\v{n}}o, Barbara Hammer, and Mikael Bod{\'e}n.
\newblock Markovian bias of neural-based architectures with feedback connections.
\newblock In \emph{Perspectives of neural-symbolic integration}, pages 95--133. Springer, 2007.

\bibitem[Jaeger et~al.(2007)Jaeger, Luko{\v{s}}evi{\v{c}}ius, Popovici, and Siewert]{jaeger2007optimization}
Herbert Jaeger, Mantas Luko{\v{s}}evi{\v{c}}ius, Dan Popovici, and Udo Siewert.
\newblock Optimization and applications of echo state networks with leaky-integrator neurons.
\newblock \emph{Neural networks}, 20\penalty0 (3):\penalty0 335--352, 2007.

\bibitem[Triefenbach and Martens(2011)]{triefenbach2011can}
Fabian Triefenbach and Jean-Pierre Martens.
\newblock Can non-linear readout nodes enhance the performance of reservoir-based speech recognizers?
\newblock In \emph{2011 First International Conference on Informatics and Computational Intelligence}, pages 262--267. IEEE, 2011.

\bibitem[Gallicchio et~al.(2017)Gallicchio, Micheli, and Pedrelli]{gallicchio2017deep}
Claudio Gallicchio, Alessio Micheli, and Luca Pedrelli.
\newblock Deep reservoir computing: A critical experimental analysis.
\newblock \emph{Neurocomputing}, 268:\penalty0 87--99, 2017.

\bibitem[Shaw et~al.(2018)Shaw, Uszkoreit, and Vaswani]{shaw2018self}
Peter Shaw, Jakob Uszkoreit, and Ashish Vaswani.
\newblock Self-attention with relative position representations.
\newblock \emph{arXiv preprint arXiv:1803.02155}, 2018.

\bibitem[Gauthier et~al.(2021)Gauthier, Bollt, Griffith, and Barbosa]{gauthier2021next}
Daniel~J Gauthier, Erik Bollt, Aaron Griffith, and Wendson~AS Barbosa.
\newblock Next generation reservoir computing.
\newblock \emph{Nature communications}, 12\penalty0 (1):\penalty0 1--8, 2021.

\bibitem[Meyer(2021)]{meyer2021alternative}
Gregory~P Meyer.
\newblock An alternative probabilistic interpretation of the huber loss.
\newblock In \emph{Proceedings of the ieee/cvf conference on computer vision and pattern recognition}, pages 5261--5269, 2021.

\bibitem[Zeng et~al.(2023)Zeng, Chen, Zhang, and Xu]{zeng2023transformers}
Ailing Zeng, Muxi Chen, Lei Zhang, and Qiang Xu.
\newblock Are transformers effective for time series forecasting?
\newblock In \emph{Proceedings of the AAAI conference on artificial intelligence}, volume~37, pages 11121--11128, 2023.

\bibitem[Wu et~al.(2022{\natexlab{a}})Wu, Hu, Liu, Zhou, Wang, and Long]{wu2022timesnet}
Haixu Wu, Tengge Hu, Yong Liu, Hang Zhou, Jianmin Wang, and Mingsheng Long.
\newblock Timesnet: Temporal 2d-variation modeling for general time series analysis.
\newblock In \emph{The eleventh international conference on learning representations}, 2022{\natexlab{a}}.

\bibitem[Zhou et~al.(2022)Zhou, Ma, Wen, Wang, Sun, and Jin]{zhou2022fedformer}
Tian Zhou, Ziqing Ma, Qingsong Wen, Xue Wang, Liang Sun, and Rong Jin.
\newblock Fedformer: Frequency enhanced decomposed transformer for long-term series forecasting.
\newblock \emph{arXiv preprint arXiv:2201.12740}, 2022.

\bibitem[Liu et~al.(2022)Liu, Wu, Wang, and Long]{liu2022non}
Yong Liu, Haixu Wu, Jianmin Wang, and Mingsheng Long.
\newblock Non-stationary transformers: Exploring the stationarity in time series forecasting.
\newblock \emph{Advances in Neural Information Processing Systems}, 35:\penalty0 9881--9893, 2022.

\bibitem[Woo et~al.(2022)Woo, Liu, Sahoo, Kumar, and Hoi]{woo2022etsformer}
Gerald Woo, Chenghao Liu, Doyen Sahoo, Akshat Kumar, and Steven Hoi.
\newblock Etsformer: Exponential smoothing transformers for time-series forecasting.
\newblock \emph{arXiv preprint arXiv:2202.01381}, 2022.

\bibitem[Campos et~al.(2023)Campos, Zhang, Yang, Kieu, Guo, and Jensen]{campos2023lightts}
David Campos, Miao Zhang, Bin Yang, Tung Kieu, Chenjuan Guo, and Christian~S Jensen.
\newblock Lightts: Lightweight time series classification with adaptive ensemble distillation.
\newblock \emph{Proceedings of the ACM on Management of Data}, 1\penalty0 (2):\penalty0 1--27, 2023.

\bibitem[Chen et~al.(2021)Chen, Peng, Fu, and Ling]{chen2021autoformer}
Minghao Chen, Houwen Peng, Jianlong Fu, and Haibin Ling.
\newblock Autoformer: Searching transformers for visual recognition.
\newblock In \emph{Proceedings of the IEEE/CVF international conference on computer vision}, pages 12270--12280, 2021.

\bibitem[Zhou et~al.(2024)Zhou, Niu, Sun, Jin, et~al.]{zhou2024one}
Tian Zhou, Peisong Niu, Liang Sun, Rong Jin, et~al.
\newblock One fits all: Power general time series analysis by pretrained lm.
\newblock \emph{Advances in neural information processing systems}, 36, 2024.

\bibitem[P{\'a}nis et~al.(2020)P{\'a}nis, Kolo{\v{s}}, and Stuchl{\'\i}k]{panis2020detection}
Radim P{\'a}nis, Martin Kolo{\v{s}}, and Zden{\v{e}}k Stuchl{\'\i}k.
\newblock Detection of chaotic behavior in time series.
\newblock \emph{arXiv preprint arXiv:2012.06671}, 2020.

\bibitem[Dai et~al.(2019)Dai, Yang, Yang, Carbonell, Le, and Salakhutdinov]{dai2019attentive}
Z~Dai, Z~Yang, Y~Yang, J~Carbonell, Q~Le, and R~Transformer-XL Salakhutdinov.
\newblock Attentive language models beyond a fixed-length context. 2019.
\newblock \emph{arXiv preprint arXiv:1901.02860}, 2019.

\bibitem[Rae et~al.(2019)Rae, Potapenko, Jayakumar, and Lillicrap]{rae2019compressive}
Jack~W Rae, Anna Potapenko, Siddhant~M Jayakumar, and Timothy~P Lillicrap.
\newblock Compressive transformers for long-range sequence modelling.
\newblock \emph{arXiv preprint arXiv:1911.05507}, 2019.

\bibitem[Wu et~al.(2022{\natexlab{b}})Wu, Rabe, Hutchins, and Szegedy]{wu2022memorizing}
Yuhuai Wu, Markus~N Rabe, DeLesley Hutchins, and Christian Szegedy.
\newblock Memorizing transformers.
\newblock \emph{arXiv preprint arXiv:2203.08913}, 2022{\natexlab{b}}.

\bibitem[Bulatov et~al.(2022)Bulatov, Kuratov, and Burtsev]{bulatov2022recurrent}
Aydar Bulatov, Yury Kuratov, and Mikhail Burtsev.
\newblock Recurrent memory transformer.
\newblock \emph{Advances in Neural Information Processing Systems}, 35:\penalty0 11079--11091, 2022.

\bibitem[Chevalier et~al.(2023)Chevalier, Wettig, Ajith, and Chen]{chevalier2023adapting}
Alexis Chevalier, Alexander Wettig, Anirudh Ajith, and Danqi Chen.
\newblock Adapting language models to compress contexts.
\newblock \emph{arXiv preprint arXiv:2305.14788}, 2023.

\bibitem[Munkhdalai et~al.(2024)Munkhdalai, Faruqui, and Gopal]{munkhdalai2024leave}
Tsendsuren Munkhdalai, Manaal Faruqui, and Siddharth Gopal.
\newblock Leave no context behind: Efficient infinite context transformers with infini-attention.
\newblock \emph{arXiv preprint arXiv:2404.07143}, 2024.

\bibitem[Ramadevi and Bingi(2022)]{ramadevi2022chaotic}
Bhukya Ramadevi and Kishore Bingi.
\newblock Chaotic time series forecasting approaches using machine learning techniques: A review.
\newblock \emph{Symmetry}, 14\penalty0 (5):\penalty0 955, 2022.

\bibitem[Sangiorgio and Dercole(2020)]{sangiorgio2020robustness}
Matteo Sangiorgio and Fabio Dercole.
\newblock Robustness of lstm neural networks for multi-step forecasting of chaotic time series.
\newblock \emph{Chaos, Solitons \& Fractals}, 139:\penalty0 110045, 2020.

\bibitem[Karasu and Altan(2022)]{karasu2022crude}
Se{\c{c}}kin Karasu and Ayta{\c{c}} Altan.
\newblock Crude oil time series prediction model based on lstm network with chaotic henry gas solubility optimization.
\newblock \emph{Energy}, 242:\penalty0 122964, 2022.

\bibitem[Sahoo et~al.(2019)Sahoo, Jha, Singh, and Kumar]{sahoo2019long}
Bibhuti~Bhusan Sahoo, Ramakar Jha, Anshuman Singh, and Deepak Kumar.
\newblock Long short-term memory (lstm) recurrent neural network for low-flow hydrological time series forecasting.
\newblock \emph{Acta Geophysica}, 67\penalty0 (5):\penalty0 1471--1481, 2019.

\bibitem[PaperwithCodes(2022)]{timeseriespaperwithcodes}
PaperwithCodes.
\newblock Leaderboards of time series forecasting, 2022.

\bibitem[Mohammadi~Farsani and Pazouki(2020)]{mohammadi2020transformer}
R~Mohammadi~Farsani and Ehsan Pazouki.
\newblock A transformer self-attention model for time series forecasting.
\newblock \emph{Journal of Electrical and Computer Engineering Innovations (JECEI)}, 9\penalty0 (1):\penalty0 1--10, 2020.

\bibitem[Huang et~al.(2019)Huang, Wang, Wu, and Tang]{huang2019dsanet}
Siteng Huang, Donglin Wang, Xuehan Wu, and Ao~Tang.
\newblock Dsanet: Dual self-attention network for multivariate time series forecasting.
\newblock In \emph{Proceedings of the 28th ACM international conference on information and knowledge management}, pages 2129--2132, 2019.

\bibitem[Jin et~al.(2023)Jin, Wang, Ma, Chu, Zhang, Shi, Chen, Liang, Li, Pan, et~al.]{jin2023time}
Ming Jin, Shiyu Wang, Lintao Ma, Zhixuan Chu, James~Y Zhang, Xiaoming Shi, Pin-Yu Chen, Yuxuan Liang, Yuan-Fang Li, Shirui Pan, et~al.
\newblock Time-llm: Time series forecasting by reprogramming large language models.
\newblock \emph{arXiv preprint arXiv:2310.01728}, 2023.

\bibitem[Chang et~al.(2024)Chang, Wang, Peng, and Chen]{chang2024llm4ts}
Ching Chang, Wei-Yao Wang, Wen-Chih Peng, and Tien-Fu Chen.
\newblock Llm4ts: Aligning pre-trained llms as data-efficient time-series forecasters.
\newblock \emph{https://arxiv.org/abs/2308.08469}, 2024.

\bibitem[Huang(2024)]{huang2024enhancing}
Xiannan Huang.
\newblock Enhancing traffic prediction with textual data using large language models.
\newblock \emph{arXiv preprint arXiv:2405.06719}, 2024.

\bibitem[de~Zarz{\`a} et~al.(2023)de~Zarz{\`a}, de~Curt{\`o}, Roig, and Calafate]{de2023llm}
I~de~Zarz{\`a}, J~de~Curt{\`o}, Gemma Roig, and Carlos~T Calafate.
\newblock Llm multimodal traffic accident forecasting.
\newblock \emph{Sensors}, 23\penalty0 (22):\penalty0 9225, 2023.

\bibitem[Bo et~al.(2020)Bo, Wang, and Zhang]{bo2020asynchronously}
Ying-Chun Bo, Ping Wang, and Xin Zhang.
\newblock An asynchronously deep reservoir computing for predicting chaotic time series.
\newblock \emph{Applied Soft Computing}, 95:\penalty0 106530, 2020.

\bibitem[Kim and King(2020)]{kim2020time}
Taehwan Kim and Brian~R King.
\newblock Time series prediction using deep echo state networks.
\newblock \emph{Neural Computing and Applications}, 32\penalty0 (23):\penalty0 17769--17787, 2020.

\bibitem[Mansoor et~al.(2021)Mansoor, Grimaccia, Leva, and Mussetta]{mansoor2021comparison}
Muhammad Mansoor, Francesco Grimaccia, Sonia Leva, and Marco Mussetta.
\newblock Comparison of echo state network and feed-forward neural networks in electrical load forecasting for demand response programs.
\newblock \emph{Mathematics and Computers in Simulation}, 184:\penalty0 282--293, 2021.

\bibitem[Bianchi et~al.(2020)Bianchi, Scardapane, L{\o}kse, and Jenssen]{bianchi2020reservoir}
Filippo~Maria Bianchi, Simone Scardapane, Sigurd L{\o}kse, and Robert Jenssen.
\newblock Reservoir computing approaches for representation and classification of multivariate time series.
\newblock \emph{IEEE transactions on neural networks and learning systems}, 32\penalty0 (5):\penalty0 2169--2179, 2020.

\bibitem[Shen et~al.(2020)Shen, Baevski, Morcos, Keutzer, Auli, and Kiela]{shen2020reservoir}
Sheng Shen, Alexei Baevski, Ari~S Morcos, Kurt Keutzer, Michael Auli, and Douwe Kiela.
\newblock Reservoir transformers.
\newblock \emph{arXiv preprint arXiv:2012.15045}, 2020.

\bibitem[Feng et~al.(2019)Feng, Yang, and Han]{feng2019chaotic}
Tian Feng, Shuying Yang, and Feng Han.
\newblock Chaotic time series prediction using wavelet transform and multi-model hybrid method.
\newblock \emph{Journal of Vibroengineering}, 21\penalty0 (7):\penalty0 1983--1999, 2019.

\bibitem[Tang et~al.(2020)Tang, Bai, Yang, and Lu]{tang2020hybrid}
Li-Hong Tang, Yu-Long Bai, Jie Yang, and Ya-Ni Lu.
\newblock A hybrid prediction method based on empirical mode decomposition and multiple model fusion for chaotic time series.
\newblock \emph{Chaos, Solitons \& Fractals}, 141:\penalty0 110366, 2020.

\bibitem[Oliveira and Torgo(2015)]{oliveira2015ensembles}
Mariana Oliveira and Luis Torgo.
\newblock Ensembles for time series forecasting.
\newblock In \emph{Asian Conference on Machine Learning}, pages 360--370. PMLR, 2015.

\bibitem[Bertsimas and Boussioux(2023)]{bertsimas2023ensemble}
Dimitris Bertsimas and Leonard Boussioux.
\newblock Ensemble modeling for time series forecasting: an adaptive robust optimization approach.
\newblock \emph{arXiv preprint arXiv:2304.04308}, 2023.

\bibitem[Li et~al.(2019)Li, Jin, Xuan, Zhou, Chen, Wang, and Yan]{li2019enhancing}
Shiyang Li, Xiaoyong Jin, Yao Xuan, Xiyou Zhou, Wenhu Chen, Yu-Xiang Wang, and Xifeng Yan.
\newblock Enhancing the locality and breaking the memory bottleneck of transformer on time series forecasting.
\newblock \emph{Advances in neural information processing systems}, 32, 2019.

\bibitem[Cini et~al.(2021)Cini, Marisca, and Alippi]{cini2021filling}
Andrea Cini, Ivan Marisca, and Cesare Alippi.
\newblock Filling the gaps: Multivariate time series imputation by graph neural networks.
\newblock \emph{arXiv preprint arXiv:2108.00298}, 2021.

\bibitem[Cao et~al.(2018)Cao, Wang, Li, Zhou, Li, and Li]{cao2018brits}
Wei Cao, Dong Wang, Jian Li, Hao Zhou, Lei Li, and Yitan Li.
\newblock Brits: Bidirectional recurrent imputation for time series.
\newblock \emph{Advances in neural information processing systems}, 31, 2018.

\bibitem[Yi et~al.(2016)Yi, Zheng, Zhang, and Li]{yi2016st}
Xiuwen Yi, Yu~Zheng, Junbo Zhang, and Tianrui Li.
\newblock St-mvl: filling missing values in geo-sensory time series data.
\newblock In \emph{Proceedings of the 25th International Joint Conference on Artificial Intelligence}, 2016.

\bibitem[Yoon et~al.(2018)Yoon, Zame, and van~der Schaar]{yoon2018estimating}
Jinsung Yoon, William~R Zame, and Mihaela van~der Schaar.
\newblock Estimating missing data in temporal data streams using multi-directional recurrent neural networks.
\newblock \emph{IEEE Transactions on Biomedical Engineering}, 66\penalty0 (5):\penalty0 1477--1490, 2018.

\bibitem[Moritz and Bartz-Beielstein(2017)]{moritz2017imputets}
Steffen Moritz and Thomas Bartz-Beielstein.
\newblock imputets: time series missing value imputation in r.
\newblock \emph{R J.}, 9\penalty0 (1):\penalty0 207, 2017.

\end{thebibliography}

\pagebreak
\appendix

\section{Dataset and Parameter Configuration} \label{dataset_details}

In this subsection, we outline the approach taken for dataset partitioning and parameter selection within our research framework.

\textbf{Dataset Partitioning:}
For the datasets ETTh1, ETTh2, ETTm1, ETTm2, Traffic, Weather, and ILI, we adopted the data preparation settings outlined by \citep{zeng2023transformers}. For other datasets, we divided the data sequentially into three distinct sets: training (70\%), validation (10\%), and testing (20\%). This division strategy facilitates efficient training, validation, and testing of models.

\textbf{Tools and Hardware:} We have used Pytorch and Huggingface libraries to implement our framework. All the experiments were carried out on a single Nvidia RTX-4090 or Tesla H100 GPU.

\
\textbf{Model Parameters:}
The configuration of model parameters plays a crucial role in experiments. For the Transformer architecture, we employ four Transformer blocks, each comprising 12 multi-head attention mechanisms with a total of 12 attention layers. In the case of the Feedforward Neural Network (FNN), we utilize hidden layers containing 64 units each.

Within the network architecture, dropout is employed as a regularization technique. Specifically, a dropout rate of 0.4 is applied to the hidden layers, while a rate of 0.1 is used for the reservoir network. Additionally, dropout with a rate of 0.2 is integrated into the attention network.

Furthermore, the reservoir state size varies between 200 and 300 units across different ensembling reservoirs. To introduce diversity in the ensembling process, spectral radius values range from 0.5 to 0.9 for different reservoirs, accompanied by leaky units with values ranging from 0.2 to 0.6.

Finally, a learning rate of $1 \times 10^{-5}$ is adopted to facilitate the training process and achieve optimal convergence.

By meticulously configuring these parameters, we aim to attain a comprehensive understanding of the model's performance across various experimental setups.

\textbf{Evaluation metric:}
In order to analyze the performance and evaluation of the proposed work,  Mean Squared Error (MSE), and Mean Absolute
Error (MAE) has been used for regression. The MSE is a quantification of the mean squared difference between the actual and predicted values. The minimum difference, the better the model's performance. On the other hand, MAE is a quantification of the mean difference between the actual and predicted values. As like as MSE, the minimum score of MAE, the better the model's performance. For classification tasks, we have used Accuracy and F1 score metrics.

\textbf{Dataset Description:}
 \begin{itemize}
     \item \textbf{ETTh (Electricity Transformer Temperature hourly-level):}
        The ETTh dataset presents a comprehensive collection of transformer temperature and load data, recorded at an hourly granularity, covering the period from July 2016 to July 2018. The dataset has been meticulously organized to encompass hourly measurements of seven vital attributes related to oil and load characteristics for electricity transformers. These incorporated features provide a valuable resource for researchers to glean insights into the temporal behavior of transformers.
        
        For the ETTh1 subset of the dataset, our model architecture entails the utilization of reservoir units ranging in size from 20 to 50, coupled with an ensemble comprising five distinct reservoirs. The attention mechanism deployed in this context employs a configuration of 15 attention heads, each possessing a dimension of 10 for both query and key components.
        
        Likewise, for the ETTh2 subset, our approach involves employing a total of seven ensemble reservoirs, each incorporating reservoir units spanning from 20 to 50 in size. The remaining parameters for this subset remain consistent with the configuration established for ETTh1.

     \item \textbf{ETTm (Electricity Transformer Temperature 15-minute-level):}
        The ETTm dataset offers a heightened level of detail and granularity by capturing measurements at intervals as fine as 15 minutes. Similar to the ETTh dataset, ETTm spans the timeframe from July 2016 to July 2018. This dataset retains the same seven pivotal oil and load attributes for electricity transformers as in the ETTh dataset; however, the key distinction lies in its higher temporal resolution. This heightened level of temporal granularity holds the potential to unveil subtler patterns and intricacies in the behavior of transformers.
        
        In the context of model architecture, the configuration adopted for the ETTm dataset involves the use of ensemble reservoirs. Specifically, for the ETTm1 subset, an ensemble of seven reservoirs is employed, while for the ETTm2 subset, an ensemble of six reservoirs is utilized. This distinctive architectural setup is designed to harness the increased temporal resolution of the ETTm dataset, enabling the model to capture and interpret the finer nuances within the transformer data.

        \item \textbf{Exchange-Rate}
        This dataset includes the daily exchange rates for eight foreign countries: Australia, Britain, Canada, Switzerland, China, Japan, New Zealand, and Singapore. This data ranges from the year 1990 to 2016. To analyze this information, we are using 15 reservoirs, which are a specific type of model to help us understand and predict changes in these exchange rates.

      \item \textbf{AQI (Air Quality ):} The AQI dataset encompasses a collection of recordings pertaining to various air quality indices, sourced from 437 monitoring stations strategically positioned across 43 cities in China. Specifically, our analysis focuses solely on the PM2.5 pollutant for this study. Notably, previous research efforts in the realm of imputation have concentrated on a truncated rendition of this dataset, which comprises data from 36 sensors.

        For our investigation, we leverage 15 ensemble reservoirs as the architectural foundation for this dataset. This ensemble configuration is tailored to the distinctive characteristics and complexities inherent in the AQI data. By deploying this setup, we endeavor to enhance our capacity to effectively address the intricate imputation challenges associated with the AQI dataset.
        
    \item \textbf{Daily Website visitors:}\footnote{https://www.kaggle.com/datasets/bobnau/daily-website-visitors}The dataset pertinent to daily website visitors provides a comprehensive record of time series data spanning five years. This dataset encapsulates diverse metrics of traffic associated with statistical forecasting teaching notes. The variables encompass daily counts encompassing page loads, unique visitors, first-time visitors, and returning visitors to an academic teaching notes website. The dataset comprises 2167 rows, covering the timeline from September 14, 2014, to August 19, 2020.

    From this array of variables, our specific focus centers on the 'first-time visitors,' which we have identified as the target variable within the context of a regression problem. By delving into this specific variable, we aim to uncover insights regarding the behavior of new visitors to the website and its associated dynamics.
    
    In terms of the model architecture employed for this dataset, we have constructed a configuration centered around 12 ensemble reservoirs. This architectural choice reflects the dataset's intricacies and the requirement for an approach that can effectively capture the underlying patterns and variations within the daily website visitor data
    
    \item \textbf{Daily Gold Price:} \footnote{https://www.kaggle.com/datasets/nisargchodavadiya/daily-gold-price-20152021-time-series} The dataset pertaining to daily gold prices offers an extensive collection of time series data encompassing a span of five years. This dataset encapsulates a range of metrics pertaining to gold price dynamics. The variables included cover daily measurements related to various aspects of gold pricing. The dataset consists of 2167 rows, covering the time period from September 14, 2014, to August 19, 2020.

    The primary objective in this analysis is regression, wherein we seek to model relationships and predict outcomes based on the available variables. Specifically, we are interested in predicting the daily gold price based on the given set of features.
    
    In terms of the architectural setup for this dataset, we have established a model configuration that employs 10 ensemble reservoirs. This choice is rooted in the need to effectively capture the nuances and trends inherent in the daily gold price data. By employing this ensemble-based approach, we aim to enhance the model's capacity to discern patterns and fluctuations within the gold price time series.
        
    \item \textbf{Daily Demand Forecasting Orders: }\footnote{https://archive.ics.uci.edu/dataset/409/dailydemandforecastingorders} This is a regression dataset that was collected during 60 days, this is a real database of a Brazilian company of large logistics. Twelve predictive attributes and a target that is considered a non-urgent order. We used 17 ensemble reservoirs for this dataset. 

    \item \textbf{Bitcoin Historical Dataset (BTC): }\footnote{https://finance.yahoo.com/quote/BTC-USD/history/}
    The dataset used in this research contains the historical price data of Bitcoin from its inception in 2009 to the present. It includes key metrics such as daily opening and closing prices, trading volumes, and market capitalizations. This comprehensive dataset captures Bitcoin's significant fluctuations and trends, reflecting major milestones and market influences. It serves as an essential tool for analyzing patterns, understanding price behaviors, and predicting future movements in the cryptocurrency market. We used 10 ensemble reservoirs for this dataset.

%    \end{itemize}
%\paragraph{Multivariate time series for classification }
%    \begin{itemize}
    \item \textbf{Absenteeism at Work: }\footnote{https://archive.ics.uci.edu/ml/datasets/Absenteeism+at+work} The dataset focusing on absenteeism at work is sourced from the UCI repository, specifically the "Absenteeism at work" dataset. In our analysis, the dataset has been utilized for classification purposes, specifically targeting the classification of individuals as social drinkers. The dataset comprises 21 attributes and encompasses a total of 740 records. These records pertain to instances where individuals were absent from work, and the data collection spans the period from January 2008 to December 2016.

    Within the scope of this analysis, the primary objective is classification, wherein we endeavor to categorize individuals as either social drinkers or non-social drinkers based on the provided attributes.
    
    The architectural arrangement for this dataset involves the utilization of 15 ensemble reservoirs. This choice of model configuration is underpinned by the aim to effectively capture the intricacies inherent in the absenteeism data and its relationship to social drinking behavior. Employing ensemble reservoirs empowers the model to discern nuanced patterns and interactions among the attributes, thereby enhancing its classification accuracy.
    
    \item \textbf{Temperature Readings: IOT Devices: }\footnote{https://www.kaggle.com/datasets/atulanandjha/temperature-readings-iot-devices} The dataset pertaining to temperature readings from IoT devices centers on the recordings captured by these devices, both inside and outside an undisclosed room (referred to as the admin room). This dataset comprises temperature readings and is characterized by five distinct features. In total, the dataset encompasses 97605 samples.

    The primary objective in this analysis involves binary classification, where the aim is to classify temperature readings based on whether they originate from an IoT device installed inside the room or outside it. The target column holds the binary class labels indicating the origin of the temperature reading.
    
    In terms of dataset characteristics, the architecture of our model involves the utilization of 15 ensemble reservoirs. This architectural choice is rooted in the desire to effectively capture the underlying patterns and distinctions present within the temperature readings dataset. The use of ensemble reservoirs enhances the model's capacity to distinguish between the different temperature reading sources, thereby facilitating accurate classification.

\end{itemize}

\section{Algorithm for RT} \label{sec:code_rt}

\begin{algorithm}[!t]
\caption{Training algorithm of Deep Reservoir Computing}\label{alg:cap}
\begin{algorithmic}
\Require $\mathbf{u}_{1:T}$, a Transformer model  $\mathbb{M}(.)$, a groups reservoirs $\mathcal{R}(.)$, an embedding function $\epsilon(.)$
\Ensure Initialization  $\mathbb{M}(.)$ and $\mathcal{R}(.)$ with parameters

\While{$epoch < epochs$}
\For{$t \gets k$ to $T$}
    \State $\mathbf{h}_t = \epsilon(\mathbf{u}_{t-k:t})$  \Comment{Embedding the time input(\ref{sec:corss-att})}
    \State $\mathbf{z}_t = \mathcal{R}(\mathbf{h}_t)$ \Comment{Readout from Group reservoir (\ref{sec:drc}, \ref{sec:lr}, \ref{sec:nr}, \ref{sec:gr}, \ref{sec:gra} ) }

     \State{$ \mathbf{f}_t = (1 - \kappa) \mathbf{z}_t \uplus \kappa \mathbf{h}_{t}$} \Comment{Concatenation (\ref{sec:rt})}
     \State{$\mathbf{\hat{u}}_{t+1:t+\tau} \gets \mathbb{M}(\mathbf{f}_t)$}
  
\EndFor

    \State $loss \gets \mathcal{L}(\mathbf{u}_{t+1:t+\tau}, \mathbf{\hat{u}}_{t+1:t+\tau})$ \Comment{Calculating loss (\ref{sec:train})}
    \State $\text{update parameters}$ 

\EndWhile
\end{algorithmic}
\end{algorithm}

In the pseudo algorithm \ref{alg:cap}, we describe how to train RT models. The algorithm initiates by setting parameters for both the Transformer and the reservoir groups. During each training epoch, it processes sequences incrementally from a specified start time \( k \) up to the total length \( T \). Each sequence segment transforms an embedded vector, which is then input into the reservoirs to produce intermediate representations. These are subsequently merged with the original embedded input, controlled by a weighting parameter \( \kappa \), forming a composite feature vector. This vector is fed into the Transformer to predict future sequence values. The training cycle completes by calculating loss between the predicted and actual values and updating the model parameters accordingly, aiming to minimize prediction errors in following epochs.

\section{Variable Descriptions }
We have described the used variables and math notations in Table \ref{tab:variable_table}.
\begin{table}[!ht]
	\centering
	\begin{tabular}{p{0.15\linewidth} | p{0.8\linewidth}} 
		\toprule
		\textbf{Variable} & \textbf{Description}\\
		\midrule
    $\mathbb{M}(.)$ & Transformer model\\
    $\mathcal{R}(.)$ & Deep reservoir computing \\\hline
    $\mathbf{u}_t$ & Input at $t$ time steps\\
    $\mathbf{x}_t$ & Reservoir state at $t$ time steps\\
    $\mathbf{y}_t$ & Linear output of reservoir\\
    $\mathbf{h}_t$ & Embedding of time steps\\
    $\mathbf{\Lambda}_t$ & Non-linear readout \\
    $\mathbf{z}_t$ & Self attention of group reservoir\\
    $\mathbf{o}_t$ & Ensembling of group reservoir \\
    $\mathbf{f}_t$ & Input of Transformer\\ \hline
    $T$ & Total time input length \\
    $k$ & Look back window \\
    $H$ & Number of attention heads \\
    $N_r$& Reservoir size \\
    $N_u$& Reservoir input size \\
    $m$ & Reservoir output size \\
    $L$ & Number of reservoirs in Group \\
    $h$ & Forecasting horizons \\
    $d_\epsilon$ & Model dimension \\
    $d_{key}$ & Dimension of key in attention \\
    $d_{value}$ & Dimension of key in attention \\
    $l_y$ & Number of layers in Transformer \\
    $c$ & Compressive Transformer memory size \\
    $r$ & Compression ratio \\
    $p$ & Number of soft-prompt summary vectors \\
    $s$ & Kernel size \\
    $P$ & Patch size\\
    
    \hline
    $\mathbf{\Phi}$ & Transformer's learnable parameters \\
    $\mathbf{W}_{q}$ & Queries weights \\
    $\mathbf{W}_{v}$ & Values weights \\
    $\mathbf{W}_{k}$ & Keys weights \\
    $\mathbf{W}_{in}$ & Input-to-reservoir weight matrix \\
    $\mathbf{W}$ & Reservoir weight matrix \\
    $\boldsymbol{\theta}$ & Bias-to-reservoir weight \\
    $\mathbf{W}_{out}$ & Reservoir-to-readout weight matrix \\
    $\boldsymbol{\theta}_{out}$ & Bias-to-readout weight \\

		\bottomrule
	\end{tabular}
	\caption{Descriptions of all variables used in this paper}
	\label{tab:variable_table}
\end{table}

\section{Baselines}
We show our experimental results of Reservoir Transformer (RT) compared to baselines, including state-of-the-art methods in time-series prediction, such as FEDformer \citep{zhou2022fedformer}, Autoformer \citep{wu2021autoformer}, Informer \citep{zhou2021informer}, Pyraformer \citep{liu2021pyraformer}, log(T)rans \citep{li2019enhancing}, GRIN \citep{cini2021filling}, BRITS \citep{cao2018brits}, STMVL \citep{yi2016st}, 
M-RNN \citep{yoon2018estimating}, ImputeTS \citep{moritz2017imputets}, and Transformer \cite{vaswani2017attention}.  To conduct experiments on multivariate and uni-variate time series, we evaluate methods outlined by ~\citep{zeng2022transformers}, which includes data splitting, pre-processing, and parameter setting.

\section{Long Uni-Variate Forecasting}
In Table~\ref{tab:uni-time} we have compared the long uni-variate time-series forecasting with state of the art time series method and showed improvement of RT in uni-variate forecasting. 
\begin{table*}[!ht]

\resizebox{\textwidth}{!}{%
\begin{tabular}{cc|cc|cc|cc|cc|cc|cc|cc|cc|cc}
\hline
\multicolumn{2}{c|}{Methods} & \multicolumn{2}{c|}{RT} & \multicolumn{2}{c|}{Linear} & \multicolumn{2}{c|}{NLinear} & \multicolumn{2}{c|}{DLinear} & \multicolumn{2}{c|}{FEDFormer-f} & \multicolumn{2}{c|}{FEDFormer-w} & \multicolumn{2}{c|}{Autoformer} & \multicolumn{2}{c|}{Informer} & \multicolumn{2}{c|}{log(T)rans}  \\ \hline
\multicolumn{2}{c|}{Metric} & \multicolumn{1}{c}{MSE} & MAE & \multicolumn{1}{c}{MSE} & MAE & \multicolumn{1}{c}{MSE} & MAE & \multicolumn{1}{c}{MSE} & MAE & \multicolumn{1}{c}{MSE} & MAE & \multicolumn{1}{c}{MSE} & MAE & \multicolumn{1}{c}{MSE} & MAE  & \multicolumn{1}{c}{MSE} & MAE  & \multicolumn{1}{c}{MSE} & MAE\\ \hline
\multicolumn{1}{c|}{\multirow{5}{*}{ETTh1}} & 96 & \multicolumn{1}{c}{\textcolor{red}{0.050}}  & {\textcolor{red}{0.172}}  & 0.189 & 0.359 & {\textcolor{blue}{0.053}} & {\textcolor{blue}{0.177}}  & 0.056 & 0.180 & 0.079 & 0.215 & 0.080 & 0.214 & 0.071 & 0.206 & 0.193 & 0.377 & 0.283 & 0.468 \\ 
\multicolumn{1}{c|}{} & 192 & \multicolumn{1}{c} {\textcolor{red}{0.064}}   & {\textcolor{red}{0.201}}  & 0.078 &  0.212 & {\textcolor{blue}{0.069}}  & {\textcolor{blue}{0.204}}  & 0.071 & {\textcolor{blue}{0.204}}  & 0.104 & 0.245 & 0.105 &  0.256 &  0.114 & 0.262 & 0.217 &  0.395 &  0.234 & 0.409 \\ 
\multicolumn{1}{c|}{} & 336 & \multicolumn{1}{c} {\textcolor{red}{0.081}}  & {\textcolor{red}{ 0.225}} & {\textcolor{blue}{0.091}}  &  0.237 & {\textcolor{red}{0.081}}   & {\textcolor{blue}{ 0.226}} & 0.098 & 0.244 & 0.119 &  0.270 &  0.120 & 0.269 & 0.107 & 0.258 & 0.202 & 0.381 & 0.386 & 0.546 \\ 
\multicolumn{1}{c|}{} & 720 & \multicolumn{1}{c} {\textcolor{blue}{0.097}}  & {\textcolor{blue}{0.292}} & 0.172 &0.340 & {\textcolor{red}{0.080}}  & {\textcolor{red}{0.226}}  & 0.189 & 0.359 & 0.142 & 0.299 & 0.127 &  0.280 & 0.126 & 0.283 & 0.183 &  0.355 & 0.475 &  0.629 \\\hline

\multicolumn{1}{c|}{\multirow{5}{*}{ETTh2}} & 96 & \multicolumn{1}{c}{\textcolor{red}{0.127}}  & {\textcolor{blue}{0.275}}   & 0.133 & 0.283 &   0.129 &   0.278 &  0.131 & 0.279 & {\textcolor{blue}{0.128}}   &   {\textcolor{red}{0.271}}  &  0.156  &  0.306 & 
 0.153 &   0.306 &  0.213 &  0.373 &  0.217&   0.379 \\ 
\multicolumn{1}{c|}{} & 192 & \multicolumn{1}{c} {\textcolor{red}{0.168}}  & {\textcolor{red}{0.324}}  & 0.176 & 0.330 &{\textcolor{blue}{0.169}}  & {\textcolor{red}{0.324}}  & 0.176 & {\textcolor{blue}{0.329}}  &  0.185 & 0.330 &  0.238 & 0.380 & 0.204 & 0.351 &  0.227 & 0.387 & 0.281 & 0.429 \\ 
\multicolumn{1}{c|}{} & 336 & \multicolumn{1}{c} {\textcolor{red}{0.192}}  & {\textcolor{red}{0.351}}  & 0.213 &  0.371 & {\textcolor{blue}{ 0.194}} &  {\textcolor{blue}{0.355}}  &  0.209 &  0.367 & 0.231 &  0.378 & 0.271 & 0.412 &  0.246 &  0.389 & 0.242 & 0.401 &  0.293 & 0.437 \\ 
\multicolumn{1}{c|}{} & 720 & \multicolumn{1}{c}{0.369} & 0.411 & 0.292 & 0.440 & {\textcolor{blue}{0.225}}    & {\textcolor{red}{ 0.381 }}  & 0.276 &  0.426 & 0.278 &  0.420 & 0.288 & 0.438 &  0.268 & 0.409 & 0.291 &  0.439 &  {\textcolor{red}{0.218}}  & {\textcolor{blue}{0.387}} \\   \hline

\multicolumn{1}{c|}{\multirow{5}{*}{ETTm1}} & 96 & \multicolumn{1}{c}{\textcolor{red}{0.024}}  & {\textcolor{red}{0.119}} & 0.028 & 0.125 & {\textcolor{blue}{0.026}}  & {\textcolor{blue}{0.122}} &  0.028 & 0.123 & 0.033 & 0.140 & 0.036 & 0.149 & 0.056 & 0.183 & 0.109 & 0.277 & 0.049 & 0.171 \\ 

\multicolumn{1}{c|}{} & 192 & \multicolumn{1}{c}{\textcolor{red}{0.037}}  &  {\textcolor{red}{0.149}}   & 0.043 &{\textcolor{blue}{0.154}} &{\textcolor{blue}{0.039}} &  {\textcolor{red}{0.149}}  & 0.045 &  0.156 & 0.058 & 0.186 & 0.069 & 0.206 & 0.081 & 0.216 & 0.151 &  0.310 & 0.157 & 0.317 \\ 
\multicolumn{1}{c|}{} & 336 & \multicolumn{1}{c}{\textcolor{red}{0.050}}  & {\textcolor{red}{0.168}}  & 0.059 & 0.180 &{\textcolor{blue}{0.052}}  &  {\textcolor{blue}{0.172}} &  0.061 & 0.182 & 0.084 & 0.231 & 0.071 & 0.209 & 0.076 & 0.218 & 0.427 & 0.591 &  0.289 & 0.459 \\ 
\multicolumn{1}{c|}{} & 720 & \multicolumn{1}{c} {\textcolor{red}{0.072}}  & {\textcolor{red}{0.199 }}   & 0.080 & 0.211 & {\textcolor{blue}{0.073}}  &  {\textcolor{blue}{0.207}} & 0.080 & 0.210 & 0.102 & 0.250 & 0.105 & 0.248 & 0.110 & 0.267 & 0.438 & 0.586 & 0.430 & 0.579  \\ \hline

\multicolumn{1}{c|}{\multirow{5}{*}{ETTm2}} & 96 & \multicolumn{1}{c}{\textcolor{red}{0.062}}  & {\textcolor{red}{0.182}} & 0.066 & 0.189 &{\textcolor{blue}{0.063}}  & {\textcolor{red}{0.182}}  & {\textcolor{blue}{0.063}} & {\textcolor{blue}{0.183}}  &  0.067 & 0.198 & {\textcolor{blue}{0.063}} & 0.189 & 0.065 & 0.189 & 0.088 & 0.225 & 0.075 & 0.208 \\ 
\multicolumn{1}{c|}{} & 192 & \multicolumn{1}{c} {\textcolor{blue}{0.092}}  & {\textcolor{blue}{0.224 }} & 0.094 & 0.230 &{\textcolor{red}{0.090}} & {\textcolor{red}{0.223}} & {\textcolor{blue}{ 0.092}}  & 0.227 & 0.102 & 0.245 & 0.110 &  0.252 & 0.118 & 0.256 & 0.132 & 0.283 & 0.129 & 0.275 \\ 
\multicolumn{1}{c|}{} & 336 & \multicolumn{1}{c} {\textcolor{red}{0.115}}  & {\textcolor{red}{0.258}}  & 0.120 & 0.263 &  {\textcolor{blue}{0.117}} & {\textcolor{blue}{0.259}} & 0.119 & 0.261 & 0.130 & 0.279 & 0.147 & 0.301 & 0.154 & 0.305 & 0.180 & 0.336 &  0.154 & 0.302 \\ 
\multicolumn{1}{c|}{} & 720 & \multicolumn{1}{c}{0.356} & 0.382  & {\textcolor{blue}{ 0.175}}  & {\textcolor{blue}{0.320}}   & {\textcolor{red}{0.170}}   & {\textcolor{red}{ 0.318}}  & {\textcolor{blue}{0.175}}    & {\textcolor{blue}{0.115}}  & 0.178 & 0.325 & 0.219 & 0.368 & 0.182 & 0.335 & 0.300 & 0.435 &  0.160 & 0.321   \\ \hline

\end{tabular}%
}
\caption{ Univariate long-term forecasting results}
\label{tab:uni-time}
\end{table*}

\section{More results} \label{sec:more_results}
Table \ref{table: result-2} shows that RT model outperforms GRIN \citep{cini2021filling}, BRITS \citep{cao2018brits}, ST-MVL \citep{yi2016st}, M-RNN \citep{yoon2018estimating}, ImputeTS \citep{moritz2017imputets} on the air quality dataset. On regression and classification tasks, RT consistently outperforms the baseline Transformer with respect to all criteria, achieving significantly lower MSE and MAE values across various datasets like DWV, DGP, DDFO, and BTC. This suggests that the RT model is better equipped to capture complex temporal patterns. In the classification tasks, the RT model demonstrates better performance in terms of accuracy, F1 Score, Jaccard Score, Roc Auc, Precision, and Recall across different datasets like Absenteeism at work and Temperature Readings. Through this analysis, we provide evidence that the RT architecture excels not only in regression tasks but also in classification tasks, outperforming traditional Transformer architectures. Additionally, we have shown the comparison of multi-variate time series with other methods such as LightTS~\citep{campos2023lightts} and Autformer~\citep{wu2021autoformer} in Table~\ref{table: result-2}.

\begin{table*}[ht]
\renewcommand{\arraystretch}{1.3}

\centering
% Reducing the column separation
\setlength{\tabcolsep}{7pt} 
% Changing the font size
\scriptsize 
\scalebox{.9}{
\begin{tabular}{c c c c c c c c c c } % centered columns (2 columns)
\hline %inserts double horizontal lines
     \multicolumn{9}{c}{Air Quality} \\ %[1ex] % inserts table
%heading
\hline % inserts a single horizontal line               \\
     Metric/Model &GRIN & BRITS & ST-MVL & M-RNN & ImputeTS   & RT  &        \\
     MAE & 10.514 & 12.242 & 12.124 &  14.242  & 19.584  & \textbf{9.046}  &        \\

    \hline   
\multicolumn{1}{c}{\textbf{Method}}&
\multicolumn{3}{c}{\textbf{Training}}& \multicolumn{4}{c}{\textbf{Test}} &\\\hline
 \multicolumn{9}{l}{\textbf{REGRESSION TASKS}}\\ 
    Model/Metric     & MSE & MAE & R2 & MSE & MAE & R2    & Parameters  \\\hline
    \midrule

     \multicolumn{9}{c}{Daily website visitors (DWV)} \\ %[1ex] % inserts table
%heading

\hline % inserts a single horizontal line               \\
    Baseline & 98.45 & 7.583 & 0.996 &   42.95 & 5.181 & 0.997 &  $\sim$13.2k       \\
    RT    & \textbf{10.400 (-89.43\%)} & 2.114 &  0.999       & 6.181 & 1.893 &  0.999 &  $\sim$3.2k  \\
    \hline

     \multicolumn{9}{c}{Daily Gold Price (DGP)} \\
    \hline
    Baseline & 39459.790 & 112.251 & 0.996      & 168608.260 & 353.310 & 0.959 &  $\sim$13.9k    \\
    RT    & \textbf{22022.981} & 86.985 &  0.998      & 28790.317 & 119.227 &  0.994 & $\sim$3.4k \\   
    \hline
    \multicolumn{9}{c}{Daily Demand Forecasting Orders (DDFO)} \\ 
    \hline
    Baseline & 312.933 & 13.274 & 0.929       & 1013.962 & 22.265 & 0.099 &  $\sim$14.1k    \\
    RT   & \textbf{76.385} & 7.487 &  0.985       & 410.652 & 10.173 &  0.720 &  $\sim$3.7k \\   \hline
    \multicolumn{9}{c}{Bitcoin Historical Dataset (BTC)} \\
    \hline
    Baseline & 188614.931 & 427.801 &  0.968       & 79389.491 & 232.119 &  0.961 &  $\sim$10.5k \\ 
    RT   & \textbf{158651.081} & 310.918 &  0.998       & 74800.535 & 211.626 &  0.989 &  $\sim$2.8k \\   \hline\hline
    \multicolumn{9}{l}{\textbf{CLASSIFICATION TASKS:}}\\ 
        Model     & Accuracy & F1 Score & Jaccard Score & Roc Auc & Precision & Recall  & Parameters  \\\hline
   \multicolumn{9}{c}{Absenteeism at work} \\
    \hline
    Baseline & 0.928 & 0.940 & 0.887 & 0.929 & 0.953 & 0.927  & $\sim$14.2k & Training     \\
    RT   & \textbf{0.947} & 0.956 & 0.916 & 0.950 & 0.976 & 0.936 & $\sim$3.5k  & Training    \\  
    \hline
        Baseline & 0.844 & 0.839 & 0.722 & 0.844 & 0.833 & 0.845  & $\sim$14.2k & Test    \\
    RT    & \textbf{0.899} & 0.901 &0.819 & 0.901 & 0.944 & 0.860 & $\sim$3.5k  & Test    \\     \hline
    \multicolumn{9}{c}{Temperature Readings} \\
    \hline
    Baseline & 0.888 & 0.934 & 0.877 & 0.877 & 0.877 &  0.915  & $\sim$10.9k & Training    \\
    RT    & \textbf{0.959} & 0.975 &  0.952 & 0.934 &0.981 & 0.970 & $\sim$2.3k  & Training    \\   \hline
    Baseline & 0.649 & 0.688 & 0.524 & 0.646 & 0.632 & 0.755  & $\sim$10.9k  & Test   \\
    RT    & \textbf{0.802} & 0.843 &  0.728 & 0.794 & 0.867 & 0.820 & $\sim$2.3k   & Test   \\   \hline\hline
    
\end{tabular}} 
\caption{Assessing the performance of regression on both training and test datasets, we utilize a window size of $100$ to observe historical data in baseline Transformers and RT.} \label{table: result-2}

\end{table*}

\begin{table*}[h!]

\label{tab:my-table}
\resizebox{1.01\textwidth}{!}{%
\begin{tabular}{|cc|cc|cc|cc|cc|cc|cc|cc|cc|cc}
\hline
\multicolumn{2}{c|}{Methods} & \multicolumn{2}{c|}{\begin{tabular}[c]{@{}c@{}}RT\\ (Ours)\end{tabular}} &  \multicolumn{2}{c|}{\begin{tabular}[c]{@{}c@{}}DLinear\\ (2023)\end{tabular}} & \multicolumn{2}{c|}{\begin{tabular}[c]{@{}c@{}}PatchTST\\ (2023)\end{tabular}} & \multicolumn{2}{c|}{\begin{tabular}[c]{@{}c@{}}TimesNet\\ (2023)\end{tabular}} & \multicolumn{2}{c|}{\begin{tabular}[c]{@{}c@{}}FEDformer\\ (2022)\end{tabular}} &   \multicolumn{2}{c|}{\begin{tabular}[c]{@{}c@{}}Stationary\\ (2022)\end{tabular}} & \multicolumn{2}{c|}{\begin{tabular}[c]{@{}c@{}}ETSformer\\ (2022)\end{tabular}} & \multicolumn{2}{c|}{\begin{tabular}[c]{@{}c@{}}LightTS\\ (2022)\end{tabular}}  & \multicolumn{2}{c}{\begin{tabular}[c]{@{}c@{}}Autoformer\\ (2021)\end{tabular}} \\ \hline
\multicolumn{2}{c|}{Metric} & \multicolumn{1}{c}{MSE} & MAE & \multicolumn{1}{c}{MSE} & MAE & \multicolumn{1}{c}{MSE} & MAE & \multicolumn{1}{c}{MSE} & MAE & \multicolumn{1}{c}{MSE} & MAE & \multicolumn{1}{c}{MSE} & MAE & \multicolumn{1}{c}{MSE} & MAE & \multicolumn{1}{c}{MSE} & MAE & \multicolumn{1}{c}{MSE} & MAE\\ \hline
\multicolumn{1}{c|}{\multirow{5}{*}{ETTh1}} & 96 & \multicolumn{1}{c}{\textcolor{red} {0.362}} & {\textcolor{blue} {0.392}}  & \multicolumn{1}{c}{0.376} & 0.397 & \multicolumn{1}{c} {\textcolor{blue}{0.375}}  & 0.399  & \multicolumn{1}{c}{\textcolor{red}{0.370}} & 0.399  & \multicolumn{1}{c}{0.384} &  0.402 & \multicolumn{1}{c}{0.376} & 0.419 & \multicolumn{1}{c}{0.449} & 0.479 & \multicolumn{1}{c}{0.424} & 0.432  & \multicolumn{1}{c}{0.449} & 0.459\\ 
\multicolumn{1}{c|}{} & 192 & \multicolumn{1}{c} {\textcolor{red} {0.396}}   & {\textcolor{red} {0.412}}  & \multicolumn{1}{c}{\textcolor{blue}{0.405}}  & {\textcolor{blue}{0.416}}  & \multicolumn{1}{c}{0.413} & 0.421 & \multicolumn{1}{c}{0.436} &  0.429 & \multicolumn{1}{c}{0.420} &  0.448  & \multicolumn{1}{c}{0.534} & 0.504 & \multicolumn{1}{c}{0.538} & 0.504 & \multicolumn{1}{c}{ 0.475} & 0.462 & \multicolumn{1}{c}{0.500} & 0.482\\ 
\multicolumn{1}{c|}{} & 336 & \multicolumn{1}{c} {\textcolor{blue}{0.427}}  & {\textcolor{red} {0.422}}  & \multicolumn{1}{c}{0.439} & 0.443 & \multicolumn{1}{c} {\textcolor{red} {0.422}}  & {\textcolor{blue}{0.436}} & \multicolumn{1}{c}{0.491} & 0.469 & \multicolumn{1}{c}{0.459} & 0.465  & \multicolumn{1}{c}{0.588} & 0.535 & \multicolumn{1}{c}{0.574} &  0.521 & \multicolumn{1}{c}{0.518} & 0.488 & \multicolumn{1}{c}{0.521} & 0.496\\ 
\multicolumn{1}{c|}{} & 720 & \multicolumn{1}{c} {\textcolor{red} {0.441}}  & {\textcolor{red} { 0.455}}  & \multicolumn{1}{c}{0.472} & 0.490 & \multicolumn{1}{c} {\textcolor{blue}{0.447}}  & {\textcolor{blue}{0.466}} & \multicolumn{1}{c}{0.521} & 0.500 & \multicolumn{1}{c}{0.506} & 0.507  & \multicolumn{1}{c}{0.643} & 0.616 & \multicolumn{1}{c}{0.562} & 0.535 & \multicolumn{1}{c}{0.547} & 0.533 & \multicolumn{1}{c}{ 0.514} & 0.512\\ 
\multicolumn{1}{c|}{} & avg & \multicolumn{1}{c} {\textcolor{red} {0.406}} & {\textcolor{red} {0.418}}  & \multicolumn{1}{c}{0.422} &  0.437 & \multicolumn{1}{c} {\textcolor{blue}{0.413}}  & {\textcolor{blue}{0.430}}& \multicolumn{1}{c}{0.458} & 0.450 & \multicolumn{1}{c}{0.440} &  0.460  & \multicolumn{1}{c}{0.570} & 0.537 & \multicolumn{1}{c}{0.542} &  0.510 & \multicolumn{1}{c}{0.491} & 0.479 & \multicolumn{1}{c}{0.496} & 0.487\\ \hline
\multicolumn{1}{c|}{\multirow{5}{*}{ETTh2}} & 96 & \multicolumn{1}{c}{\textcolor{red} {0.262}}  &  {\textcolor{red} {0.320}}  &  \multicolumn{1}{c}{0.289} & 0.353 & \multicolumn{1}{c}{\textcolor{blue}{0.274}}  & {\textcolor{blue}{0.336}} & \multicolumn{1}{c}{0.340} & 0.374 & \multicolumn{1}{c}{0.358} & 0.397 &  \multicolumn{1}{c}{0.476} & 0.458 & \multicolumn{1}{c}{0.340} & 0.391 & \multicolumn{1}{c}{0.397} & 0.437  & \multicolumn{1}{c}{0.346} & 0.388\\ 
\multicolumn{1}{c|}{} & 192 & \multicolumn{1}{c} {\textcolor{red} {0.331}}  & {\textcolor{red} { 0.373}} & \multicolumn{1}{c}{0.383} & 0.418 & \multicolumn{1}{c} {\textcolor{blue}{0.339}}  & {\textcolor{blue}{0.379}}  & \multicolumn{1}{c}{0.402} & 0.414 & \multicolumn{1}{c}{0.429} & 0.439  & \multicolumn{1}{c}{0.512} & 0.493 & \multicolumn{1}{c}{0.430} & 0.439 & \multicolumn{1}{c}{ 0.520} & 0.504 & \multicolumn{1}{c}{0.456} & 0.452\\ 
\multicolumn{1}{c|}{} & 336 & \multicolumn{1}{c}{\textcolor{blue}{0.358}}  & {\textcolor{blue}{0.403}}  & \multicolumn{1}{c}{0.448} & 0.465 & \multicolumn{1}{c}{\textcolor{red} {0.329}}  & {\textcolor{red} {0.380}}  & \multicolumn{1}{c}{0.452} & 0.452 & \multicolumn{1}{c}{0.496} & 0.487  & \multicolumn{1}{c}{0.552} &  0.551 & \multicolumn{1}{c}{0.485} & 0.479 & \multicolumn{1}{c}{0.626} & 0.559 & \multicolumn{1}{c}{0.482} & 0.486\\ 
\multicolumn{1}{c|}{} & 720 & \multicolumn{1}{c} {\textcolor{red} {0.369}}  & {\textcolor{red}{0.411}}  & \multicolumn{1}{c}{0.605} & 0.551 & \multicolumn{1}{c} {\textcolor{blue}{0.379}}  & {\textcolor{blue}{0.422}}   & \multicolumn{1}{c}{0.462} & 0.468 & \multicolumn{1}{c}{0.463} & 0.474  & \multicolumn{1}{c}{0.562} & 0.560 & \multicolumn{1}{c}{0.500} & 0.497 & \multicolumn{1}{c}{0.863} & 0.672 & \multicolumn{1}{c}{0.515} & 0.511\\ 
\multicolumn{1}{c|}{} & avg & \multicolumn{1}{c} {\textcolor{red}{0.330}}  & {\textcolor{red}{0.376}}  & \multicolumn{1}{c}{0.431} & 0.446 & \multicolumn{1}{c}{\textcolor{red}{0.330}}  & {\textcolor{blue}{0.379}}  & \multicolumn{1}{c} {\textcolor{blue}{0.414}}  & 0.427 & \multicolumn{1}{c}{0.437} &  0.449   & \multicolumn{1}{c}{0.526} & 0.516 & \multicolumn{1}{c}{0.439} & 0.452 & \multicolumn{1}{c}{0.602} & 0.543 & \multicolumn{1}{c}{0.450} & 0.459\\ \hline
\multicolumn{1}{c|}{\multirow{5}{*}{ETTm1}} & 96 & \multicolumn{1}{c} {\textcolor{red}{0.272}} & {\textcolor{red}{0.336}}  &  \multicolumn{1}{c}{0.299} & 0.343 & \multicolumn{1}{c}  {\textcolor{blue}{0.290}}  & {\textcolor{blue}{0.342}}  & \multicolumn{1}{c}{0.338} & 0.375 & \multicolumn{1}{c}{0.379} & 0.419  & \multicolumn{1}{c}{0.386} & 0.398 & \multicolumn{1}{c}{0.375} & 0.398 & \multicolumn{1}{c}{0.374} & 0.400 & \multicolumn{1}{c}{0.505} & 0.475\\ 
\multicolumn{1}{c|}{} & 192 & \multicolumn{1}{c} {\textcolor{red}{0.310}}  &  {\textcolor{blue}{0.357}}    & \multicolumn{1}{c}{0.335} &  {\textcolor{blue}{0.365}}  & \multicolumn{1}{c} {\textcolor{red}{0.332}}  & 0.369 & \multicolumn{1}{c}{0.374} & 0.387 & \multicolumn{1}{c}{0.426} & 0.441 & \multicolumn{1}{c}{0.459} & 0.444 & \multicolumn{1}{c}{0.408} & 0.410 & \multicolumn{1}{c}{0.400} & 0.407 & \multicolumn{1}{c}{0.553} & 0.496 \\ 
\multicolumn{1}{c|}{} & 336 & \multicolumn{1}{c} {\textcolor{red}{0.351}}  &  {\textcolor{blue}{0.382}}   & \multicolumn{1}{c}{0.369} & {\textcolor{blue}{0.386}} & \multicolumn{1}{c} {\textcolor{red}{0.366}}  & 0.392 & \multicolumn{1}{c}{0.410} & 0.411 & \multicolumn{1}{c}{0.445} & 0.459  & \multicolumn{1}{c}{0.495} & 0.464 & \multicolumn{1}{c}{0.435} & 0.428 & \multicolumn{1}{c}{0.438} & 0.438 & \multicolumn{1}{c}{0.621} & 0.537\\ 
\multicolumn{1}{c|}{} & 720 & \multicolumn{1}{c} {\textcolor{red}{0.380}}  & {\textcolor{red}{0.409}}   & \multicolumn{1}{c}{0.425} & 0.421 & \multicolumn{1}{c} {\textcolor{blue}{0.416}}  & {\textcolor{blue}{0.420}}  & \multicolumn{1}{c}{0.478} & 0.450 & \multicolumn{1}{c}{0.543} & 0.490  & \multicolumn{1}{c}{0.585} & 0.516 & \multicolumn{1}{c}{0.499} & 0.462 & \multicolumn{1}{c}{0.527} & 0.502 & \multicolumn{1}{c}{0.671} & 0.561\\ 
\multicolumn{1}{c|}{} & avg & \multicolumn{1}{c} {\textcolor{red}{0.328}} & {\textcolor{red}{0.371}}   &  \multicolumn{1}{c}{0.357} & {\textcolor{blue}{0.378}} & \multicolumn{1}{c} {\textcolor{blue}{0.351}}  & 0.380 & \multicolumn{1}{c}{0.400} & 0.406 & \multicolumn{1}{c}{0.448} & 0.452 & \multicolumn{1}{c}{0.481} & 0.456 & \multicolumn{1}{c}{0.429} & 0.425 & \multicolumn{1}{c}{0.435} & 0.437 & \multicolumn{1}{c}{0.588} & 0.517 \\ \hline
\multicolumn{1}{c|}{\multirow{5}{*}{ETTm2}} & 96 & \multicolumn{1}{c}{\textcolor{red}{0.163}}  & {\textcolor{red}{0.254}} &  \multicolumn{1}{c}{0.167} & 0.269 & \multicolumn{1}{c} {\textcolor{blue}{0.165}}  & {\textcolor{blue}{0.255}}  & \multicolumn{1}{c}{0.187} & 0.267 & \multicolumn{1}{c}{0.203} & 0.287  & \multicolumn{1}{c}{0.192} & 0.274 & \multicolumn{1}{c}{0.189} & 0.280 & \multicolumn{1}{c}{0.209} & 0.308 & \multicolumn{1}{c}{0.255} & 0.339\\ 
\multicolumn{1}{c|}{} & 192 & \multicolumn{1}{c} {\textcolor{red}{0.218}}  & {\textcolor{red}{0.290}}  & \multicolumn{1}{c}{0.224} & 0.303 & \multicolumn{1}{c} {\textcolor{blue}{0.220}} & {\textcolor{blue}{0.292}}  & \multicolumn{1}{c}{0.249} & 0.309 & \multicolumn{1}{c}{0.269} & 0.328  & \multicolumn{1}{c}{0.280} & 0.339 & \multicolumn{1}{c}{0.253} & 0.319 & \multicolumn{1}{c}{0.311} & 0.382 & \multicolumn{1}{c}{0.281} & 0.340\\ 
\multicolumn{1}{c|}{} & 336 & \multicolumn{1}{c} {\textcolor{red}{0.271}}  &{\textcolor{red}{0.328}}  & \multicolumn{1}{c}{0.281} & 0.342 & \multicolumn{1}{c} {\textcolor{blue}{0.274}}  & {\textcolor{blue}{0.329}}  & \multicolumn{1}{c}{0.321} & 0.351 & \multicolumn{1}{c}{0.325} & 0.366 & \multicolumn{1}{c}{0.334} & 0.361 & \multicolumn{1}{c}{0.314} & 0.357 & \multicolumn{1}{c}{0.442} & 0.466 & \multicolumn{1}{c}{0.339} & 0.372  \\ 
\multicolumn{1}{c|}{} & 720 & \multicolumn{1}{c} {\textcolor{red}{0.356}}  & {\textcolor{red}{0.382 }}  & \multicolumn{1}{c}{0.397} & 0.421 & \multicolumn{1}{c} {\textcolor{blue}{0.362}}  & {\textcolor{blue}{0.385}}  & \multicolumn{1}{c}{0.408} & 0.403 & \multicolumn{1}{c}{0.421} &  0.415  & \multicolumn{1}{c}{0.417} &  0.413 & \multicolumn{1}{c}{0.414} & 0.413 & \multicolumn{1}{c}{0.675} & 0.587 & \multicolumn{1}{c}{0.433} & 0.432\\ 
\multicolumn{1}{c|}{} & avg & \multicolumn{1}{c} {\textcolor{red}{0.252}}  & 0.316 & \multicolumn{1}{c}{ 0.267} & {\textcolor{blue}{0.333}}  & \multicolumn{1}{c} {\textcolor{blue}{0.255}}  & {\textcolor{red}{0.315}}  & \multicolumn{1}{c}{0.291} & {\textcolor{blue}{0.333}} & \multicolumn{1}{c}{0.305} & 0.349  & 0.371 & \multicolumn{1}{c|}{0.306} & 0.347 & \multicolumn{1}{c|}{0.293} & 0.342 & \multicolumn{1}{c|}{0.409} & 0.436 & \multicolumn{1}{c}{0.327}\\ \hline
\multicolumn{1}{c|}{\multirow{5}{*}{Weather}} & 96 & \multicolumn{1}{c}{\textcolor{red}{0.146}}   & {\textcolor{red}{0.196}}   & \multicolumn{1}{c}{0.176} & 0.237 & \multicolumn{1}{c} {\textcolor{blue}{0.149}} & {\textcolor{blue}{0.198}} & \multicolumn{1}{c}{0.172} & 0.220 & \multicolumn{1}{c}{0.217} & 0.296 & \multicolumn{1}{c}{0.173} & 0.223 & \multicolumn{1}{c}{0.197} & 0.281 & \multicolumn{1}{c}{0.182} & 0.242 & \multicolumn{1}{c}{0.266} & 0.336 \\ 
\multicolumn{1}{c|}{} & 192 & \multicolumn{1}{c} {\textcolor{red}{0.193}}   & {\textcolor{red}{0.236}}   &  \multicolumn{1}{c}{0.220} & 0.282 & \multicolumn{1}{c} {\textcolor{blue}{0.194}} 
  &  {\textcolor{blue}{0.241}}   & \multicolumn{1}{c}{0.219} & 0.261 & \multicolumn{1}{c}{0.276} & 0.336  & \multicolumn{1}{c}{0.245} & 0.285 & \multicolumn{1}{c}{0.237} & 0.312 & \multicolumn{1}{c}{0.227} & 0.287 & \multicolumn{1}{c}{0.307} & 0.367\\ 
\multicolumn{1}{c|}{} & 336 & \multicolumn{1}{c} {\textcolor{red}{0.239}}  & {\textcolor{red}{0.240}}  & \multicolumn{1}{c}{0.265} &  0.319 & \multicolumn{1}{c}  {\textcolor{blue}{0.245}}  &  {\textcolor{blue}{0.282}}  & \multicolumn{1}{c}{0.280} & 0.306 & \multicolumn{1}{c}{0.339} & 0.380  & \multicolumn{1}{c}{0.321} & 0.338 & \multicolumn{1}{c}{0.298} & 0.353 & \multicolumn{1}{c} { 0.282} & 0.334 & \multicolumn{1}{c}{ 0.359} & 0.395\\ 
\multicolumn{1}{c|}{} & 720 & \multicolumn{1}{c} {\textcolor{red}{0.301}}   & {\textcolor{red}{0.316}}  &  \multicolumn{1}{c}{0.333} & 0.362 & \multicolumn{1}{c} {\textcolor{blue}{0.314}}  & {\textcolor{blue}{0.334}}  & \multicolumn{1}{c}{0.365} & 0.359 & \multicolumn{1}{c}{0.403} & 0.428  & \multicolumn{1}{c}{ 0.414} &  0.410 & \multicolumn{1}{c}{0.352} & 0.288 & \multicolumn{1}{c}{0.352} &  0.386 & \multicolumn{1}{c}{0.419} & 0.428\\ 
\multicolumn{1}{c|}{} & avg & \multicolumn{1}{c} {\textcolor{red}{0.219}}   & {\textcolor{red}{0.247}}   & \multicolumn{1}{c}{0.248} & 0.300 & \multicolumn{1}{c}{\textcolor{blue}{0.225}}  & {\textcolor{blue}{0.264}}  & \multicolumn{1}{c}{0.259} & 0.287 & \multicolumn{1}{c}{ 0.309} & 0.360  & \multicolumn{1}{c}{0.288} & 0.314 & \multicolumn{1}{c}{0.271} & 0.334 & \multicolumn{1}{c}{0.261} &  0.312 & \multicolumn{1}{c}{0.338} &0.382\\ \hline

\multicolumn{1}{c|}{\multirow{5}{*}{Electricity}} & 96 & \multicolumn{1}{c}{\textcolor{blue}{0.130}}  & {\textcolor{red}{0.219}}  & \multicolumn{1}{c}{0.140} & 0.237  & \multicolumn{1}{c}{\textcolor{red}{0.129}}  & {\textcolor{blue}{0.222}}  & \multicolumn{1}{c}{0.168} &  0.272 & \multicolumn{1}{c}{0.193} & 0.308  & \multicolumn{1}{c}{0.169} & 0.273 & \multicolumn{1}{c}{0.187} & 0.304 & \multicolumn{1}{c}{0.207} & 0.307 & \multicolumn{1}{c}{ 0.201} & 0.317\\ 
\multicolumn{1}{c|}{} & 192 & \multicolumn{1}{c} {\textcolor{blue}{0.154}}   & {\textcolor{blue}{0.243 }}  & \multicolumn{1}{c} {\textcolor{red}{0.153}}   & 0.249 & \multicolumn{1}{c}{0.157} & {\textcolor{red}{0.240}}   & \multicolumn{1}{c}{0.184} & 0.289 & \multicolumn{1}{c}{0.201} &  0.315  & \multicolumn{1}{c}{0.182} & 0.286 & \multicolumn{1}{c}{0.199} & 0.315 & \multicolumn{1}{c}{0.213} & 0.316 & \multicolumn{1}{c}{ 0.222} & 0.334\\ 
\multicolumn{1}{c|}{} & 336 & \multicolumn{1}{c} {\textcolor{red}{0.158}}  & {\textcolor{red}{0.245}}  & \multicolumn{1}{c}{0.169} & 0.267 & \multicolumn{1}{c} {\textcolor{blue}{{0.163}}}  & {\textcolor{blue}{0.259}} & \multicolumn{1}{c}{0.198} &  0.300 & \multicolumn{1}{c}{0.214} & 0.329 & \multicolumn{1}{c}{0.200} & 0.304 & \multicolumn{1}{c}{0.212} & 0.329 & \multicolumn{1}{c}{0.230} & 0.333  & \multicolumn{1}{c}{0.231} & 0.338\\ 
\multicolumn{1}{c|}{} & 720 & \multicolumn{1}{c} {\textcolor{red}{0.190}}   & {\textcolor{blue}{0.294}}  &  \multicolumn{1}{c}{0.203} & 0.301 & \multicolumn{1}{c} {\textcolor{blue}{0.197}}  & {\textcolor{red}{ 0.290}}  & \multicolumn{1}{c}{0.220} & 0.320 & \multicolumn{1}{c}{0.246} & 0.355  & \multicolumn{1}{c}{0.222} & 0.321 & \multicolumn{1}{c}{0.233} & 0.345 & \multicolumn{1}{c}{0.265} & 0.360 & \multicolumn{1}{c}{0.254} &  0.361\\ 
\multicolumn{1}{c|}{} & avg & \multicolumn{1}{c} {\textcolor{red}{0.158}}  & {\textcolor{red}{0.251}}    &\multicolumn{1}{c}{0.166} & 0.263 & \multicolumn{1}{c}{\textcolor{blue}{{0.161}}}  & {\textcolor{blue}{{ 0.252}}} & \multicolumn{1}{c}{0.192} &  0.295 & \multicolumn{1}{c}{0.214} & 0.327  & \multicolumn{1}{c}{ 0.193} & 0.296 & \multicolumn{1}{c}{ 0.208} & 0.323 & \multicolumn{1}{c}{0.229} &0.329 & \multicolumn{1}{c}{0.227} & 0.338\\ \hline

\multicolumn{1}{c|}{\multirow{5}{*}{Traffic}} & 96 & \multicolumn{1}{c}{\textcolor{blue}{{0.361}}}  & {\textcolor{red}{0.248}}  &  \multicolumn{1}{c}{0.410} & 0.282 & \multicolumn{1}{c}{\textcolor{red}{0.360}}   & {\textcolor{blue}{{0.249}}}  & \multicolumn{1}{c}{0.593} &  0.321 & \multicolumn{1}{c}{0.587} & 0.366 & \multicolumn{1}{c}{0.612} & 0.338 & \multicolumn{1}{c}{0.607} &  0.392 & \multicolumn{1}{c}{ 0.615} & 0.391 & \multicolumn{1}{c}{0.613} & 0.388 \\ 
\multicolumn{1}{c|}{} & 192 & \multicolumn{1}{c}{\textcolor{red}{0.376} }  & {\textcolor{red}{ 0.250}} & \multicolumn{1}{c}{0.423} & 0.287 & \multicolumn{1}{c}{\textcolor{blue}{{0.379}}}  & {\textcolor{blue}{ 0.256}}  & \multicolumn{1}{c}{0.617} &  0.336 & \multicolumn{1}{c}{ 0.604} &  0.373 & \multicolumn{1}{c}{0.613} & 0.340 & \multicolumn{1}{c}{0.621} &  0.399 & \multicolumn{1}{c}{ 0.601} & 0.382 & \multicolumn{1}{c}{0.616} & 0.382 \\ 
\multicolumn{1}{c|}{} & 336 & \multicolumn{1}{c} {\textcolor{red}{0.382} }  & {\textcolor{blue}{0.268}}  &  \multicolumn{1}{c}{0.436} &  0.296 & \multicolumn{1}{c}{\textcolor{blue}{0.392}}  & {\textcolor{red}{ 0.264}} & \multicolumn{1}{c}{0.629} & 0.336 & \multicolumn{1}{c}{0.621} & 0.383 & \multicolumn{1}{c}{0.618} &  0.328 & \multicolumn{1}{c}{0.622} & 0.396 & \multicolumn{1}{c}{0.613} & 0.386 & \multicolumn{1}{c}{0.622} & 0.337 \\ 
\multicolumn{1}{c|}{} & 720 & \multicolumn{1}{c} {\textcolor{red}{0.426}}  & {\textcolor{red}{0.382}}  & \multicolumn{1}{c}{0.466} & 0.315 & \multicolumn{1}{c}{\textcolor{blue}{{0.432}}}  &  {\textcolor{blue}{{0.286}}}  & \multicolumn{1}{c}{0.640} & 0.350  & \multicolumn{1}{c}{0.626} &  0.382  & \multicolumn{1}{c}{0.653} &  0.355 & \multicolumn{1}{c}{0.632} & 0.396 & \multicolumn{1}{c}{0.658} & 0.407 & \multicolumn{1}{c}{0.660} & 0.408\\ 
\multicolumn{1}{c|}{} & avg & \multicolumn{1}{c} {\textcolor{red}{0.386}}  & {\textcolor{red}{0.261}}  & \multicolumn{1}{c}{0.433} & 0.295 & \multicolumn{1}{c}{\textcolor{blue}{{0.390}}}   & {\textcolor{blue}{0.263}}   & \multicolumn{1}{c}{0.620} & 0.336 & \multicolumn{1}{c}{0.610} &  0.376 & \multicolumn{1}{c}{0.628} & 0.379 & \multicolumn{1}{c}{0.624} & 0.340 & \multicolumn{1}{c}{0.621} &  0.396 & \multicolumn{1}{c}{ 0.622} &  0.392 \\ 
\hline

\multicolumn{1}{c|}{\multirow{5}{*}{ILI}} & 24 & \multicolumn{1}{c} {\textcolor{red}{1.301}} & {\textcolor{red}{0.739}}  & \multicolumn{1}{c}{2.215} &  1.081 & \multicolumn{1}{c}{\textcolor{blue}{{1.319}}}   & {\textcolor{blue}{0.754}} 
 & \multicolumn{1}{c}{2.317} &  0.934 & \multicolumn{1}{c}{3.228} &  1.260  & \multicolumn{1}{c}{2.294} & 0.945 & \multicolumn{1}{c}{2.527} & 1.020 & \multicolumn{1}{c}{8.313} & 2.144 & \multicolumn{1}{c}{3.483} &  1.287\\ 
\multicolumn{1}{c|}{} & 36 & \multicolumn{1}{c}{\textcolor{red}{1.409}}  & {\textcolor{red}{0.824}}  & \multicolumn{1}{c}{1.963} & 0.963 & \multicolumn{1}{c}{\textcolor{blue}{{1.430}}}  & {\textcolor{blue}{0.834}}  & \multicolumn{1}{c}{1.972} &0.920 & \multicolumn{1}{c}{2.679} &  1.080  & \multicolumn{1}{c}{1.825} &  0.848 & \multicolumn{1}{c}{ 2.615} & 1.007 & \multicolumn{1}{c}{ 6.631} &  1.902 & \multicolumn{1}{c}{ 3.103} &  1.148\\ 
\multicolumn{1}{c|}{} & 48 & \multicolumn{1}{c} {\textcolor{red}{1.520}}  & {\textcolor{red}{0.807 }} & \multicolumn{1}{c}{2.130} & 1.024 & \multicolumn{1}{c}{\textcolor{blue}{{1.553}}} & {\textcolor{blue}{0.815}}  & \multicolumn{1}{c}{2.238} & 0.940 & \multicolumn{1}{c}{2.622} & 1.078  & \multicolumn{1}{c}{2.010} & 0.900 & \multicolumn{1}{c}{2.359} & 0.972 & \multicolumn{1}{c}{7.299} & 1.982 & \multicolumn{1}{c}{2.669} & 1.085\\ 
\multicolumn{1}{c|}{} & 60 & \multicolumn{1}{c}{\textcolor{blue}{{1.553}}} &  {\textcolor{blue}{0.852}}  &\multicolumn{1}{c}{2.368} & 1.096  & \multicolumn{1}{c}{\textcolor{red}{1.520}}    & {\textcolor{red}{ 0.788}}  & \multicolumn{1}{c}{2.027} & 0.928 & \multicolumn{1}{c}{2.857} & 1.157  & \multicolumn{1}{c}{2.178} & 0.963 & \multicolumn{1}{c}{2.487} & 1.016 & \multicolumn{1}{c}{7.283} & 1.985 & \multicolumn{1}{c}{2.770} & 1.125\\ 
\multicolumn{1}{c|}{} & avg & \multicolumn{1}{c} {\textcolor{blue}{{1.439}}}  & {\textcolor{blue}{0.805}}    & \multicolumn{1}{c}{2.169} &  1.041 & \multicolumn{1}{c}{\textcolor{red}{1.443}}   & {\textcolor{red}{0.797}}  & \multicolumn{1}{c}{2.139} & 0.931 & \multicolumn{1}{c}{2.847} &  1.144 & \multicolumn{1}{c}{2.077} & 0.914 & \multicolumn{1}{c}{2.497} & 1.004  & \multicolumn{1}{c}{7.382} & 2.003 & \multicolumn{1}{c}{3.006} &  1.161 \\ 
\hline

\end{tabular}%

}
\caption{Multivariate long-term time-series forecasting: lower values indicate better performance. Forecasting horizons are $h \in \{24, 36, 48, 60\}$ for ILI and $h \in \{96, 192, 336, 720\}$ for the rest. %Optimal results are emphasized in \textbf{bold}, while the top results from Transformers are \underline{underlined} for distinction."
}
\label{table: result-1}
\end{table*}

\section{More Ablation Analysis} \label{ablation}

\iffalse
We illustrate and measure how well-distributed the features input of the Transformer with or without reservoir are. More distribution means higher entropy and more information, and vice versa. To perform this experiment,  we have used the Local Interpretable Model-agnostic Explanations (LIME) algorithm \citep{ribeiro2016should}. 
The initial step generates perturbed inputs using $\mathbf{h}_{t-k:t}$,  $\mathbf{z}_t$ and  $\mathbf{f}_t$ by adding noise. A simpler interpretable model is then trained using these perturbed inputs along with corresponding output time series which is $\mathbf{u}_{t+1:t+\tau}$.
Ultimately, we utilize the coefficients of the interpretable model, along with LIME-calculated feature importances, to explain the $\mathbf{h}_{t-k:t}$, $\mathbf{z}_t$ and $\mathbf{f}_t$ individually. 
the importance  of $\mathbf{h}_{t-k:t}$, $\mathbf{z}_t$ and $\mathbf{f}_t$ for the ETTh1, ETTh2, ETTm1, and ETTm2 by describing the LIME calculation. We can see that the input from reservoir readout $\mathbf{z}_t$  has higher entropy than Transformer original input features $\mathbf{h}_{t-k:t}$ to predict $\mathbf{u}_{t+1:t+\tau}$. And the concatation using Equation \ref{eq:concat}, we achieve the highest entropy. (More ablation in the Appendix \ref{ablation}).
\fi
\subsection{Look-Back Window Size vs Loss}
\begin{figure*}[ht]
  \begin{center}
      \includegraphics[width=0.790\linewidth]{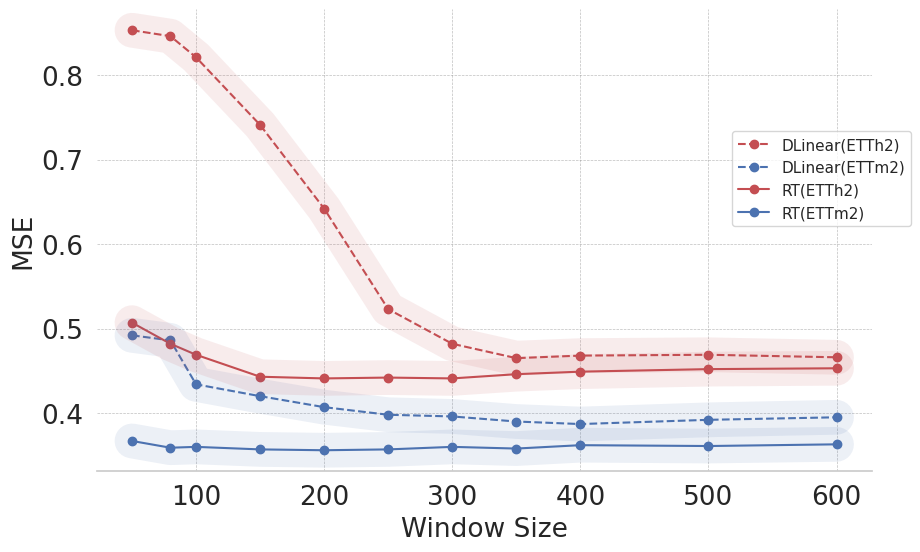}

    \caption{Look back window size vs loss}
   \label{fig:window_size}
  \end{center}
  \end{figure*}

In Figure~\ref{fig:window_size}, we have proved that RT works for the short look-back window better than the Dlinear method, since RT can capture the long-term dependencies using group reservoirs.

\subsection{Graphical Explanation of Features} \label{sec:more_sing_feature}

\begin{figure*}[!ht]
  \begin{center}
      \includegraphics[width=0.790\linewidth]{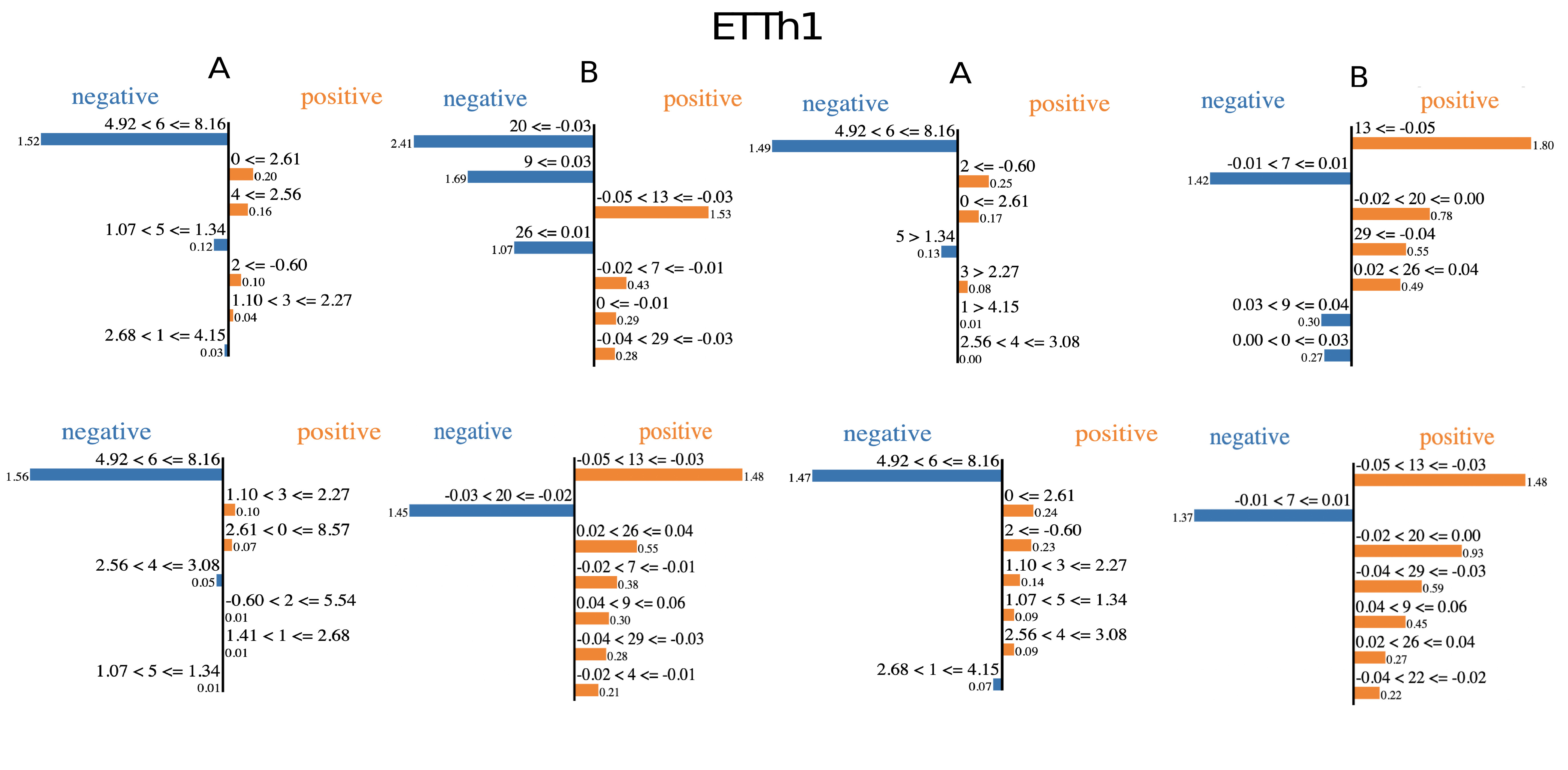}
      \includegraphics[width=0.790\linewidth]{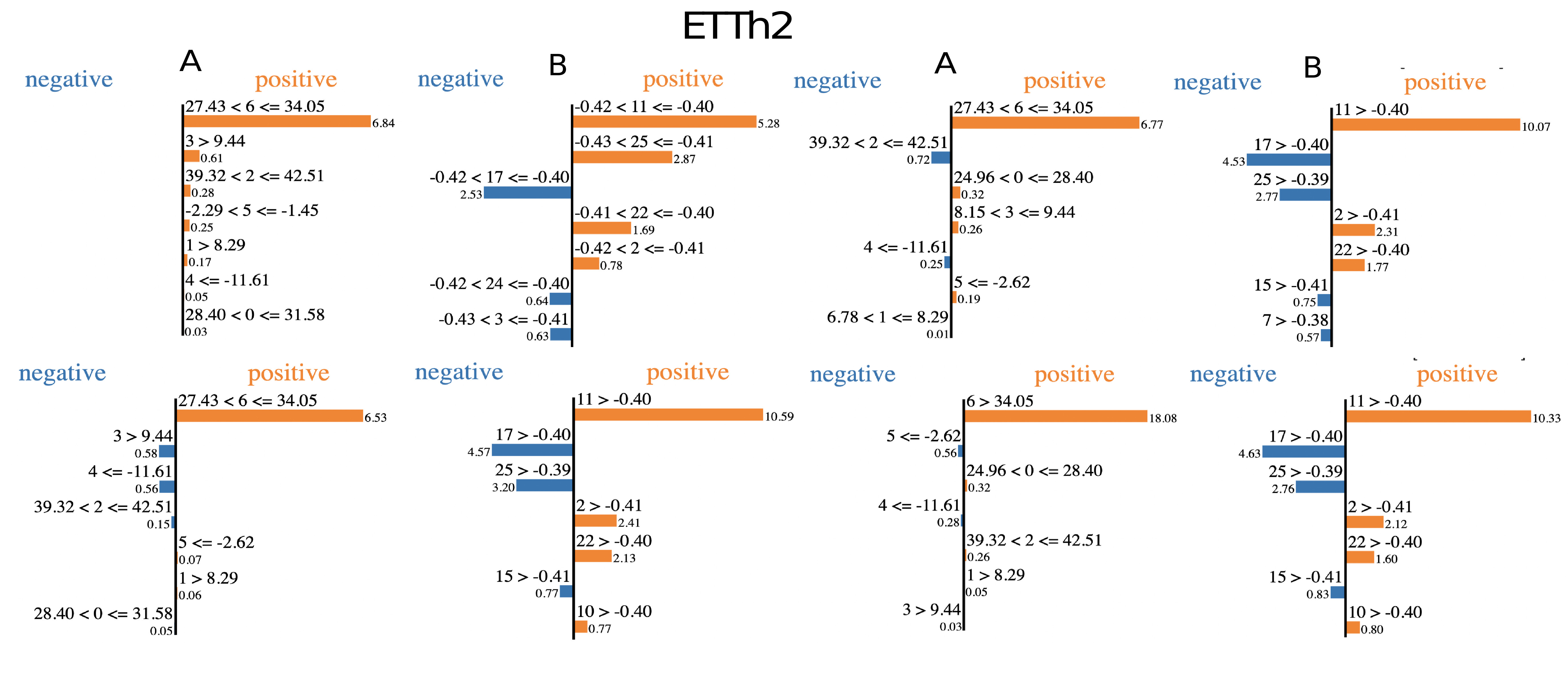}
      \includegraphics[width=0.790\linewidth]{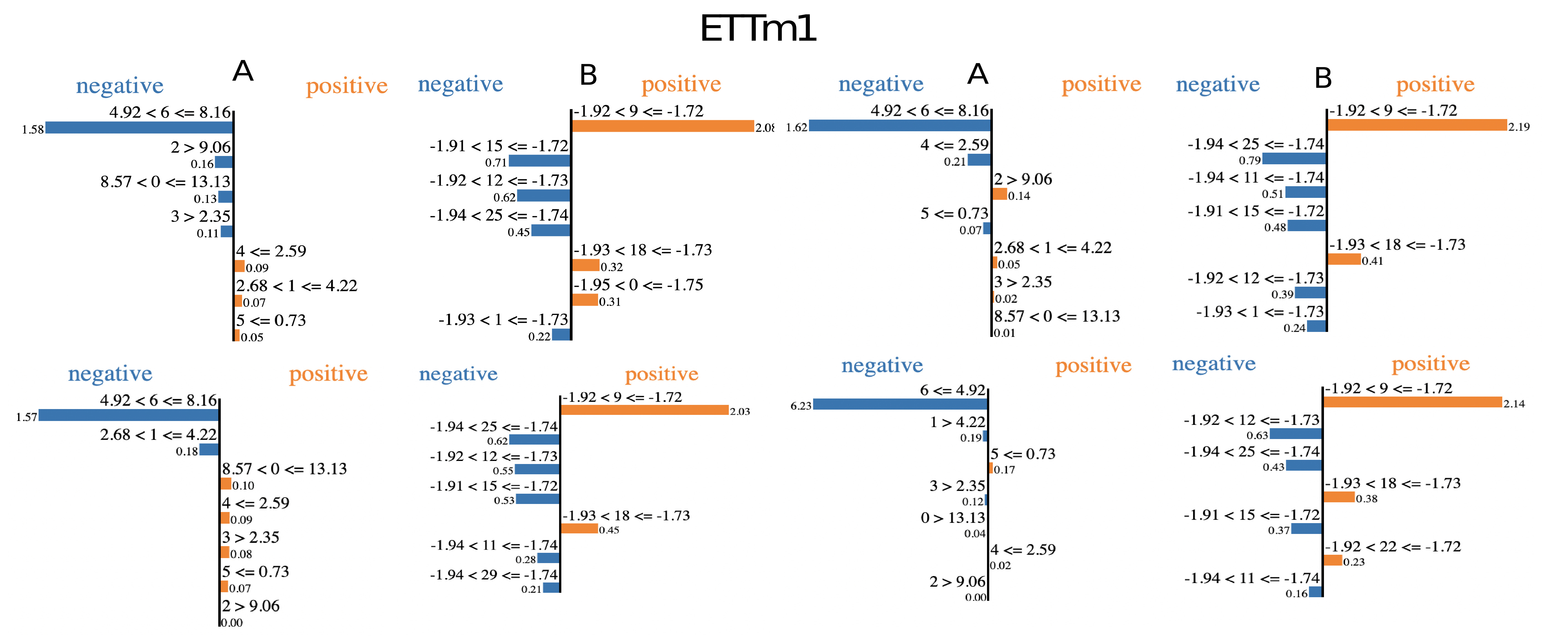}
      \includegraphics[width=0.790\linewidth]{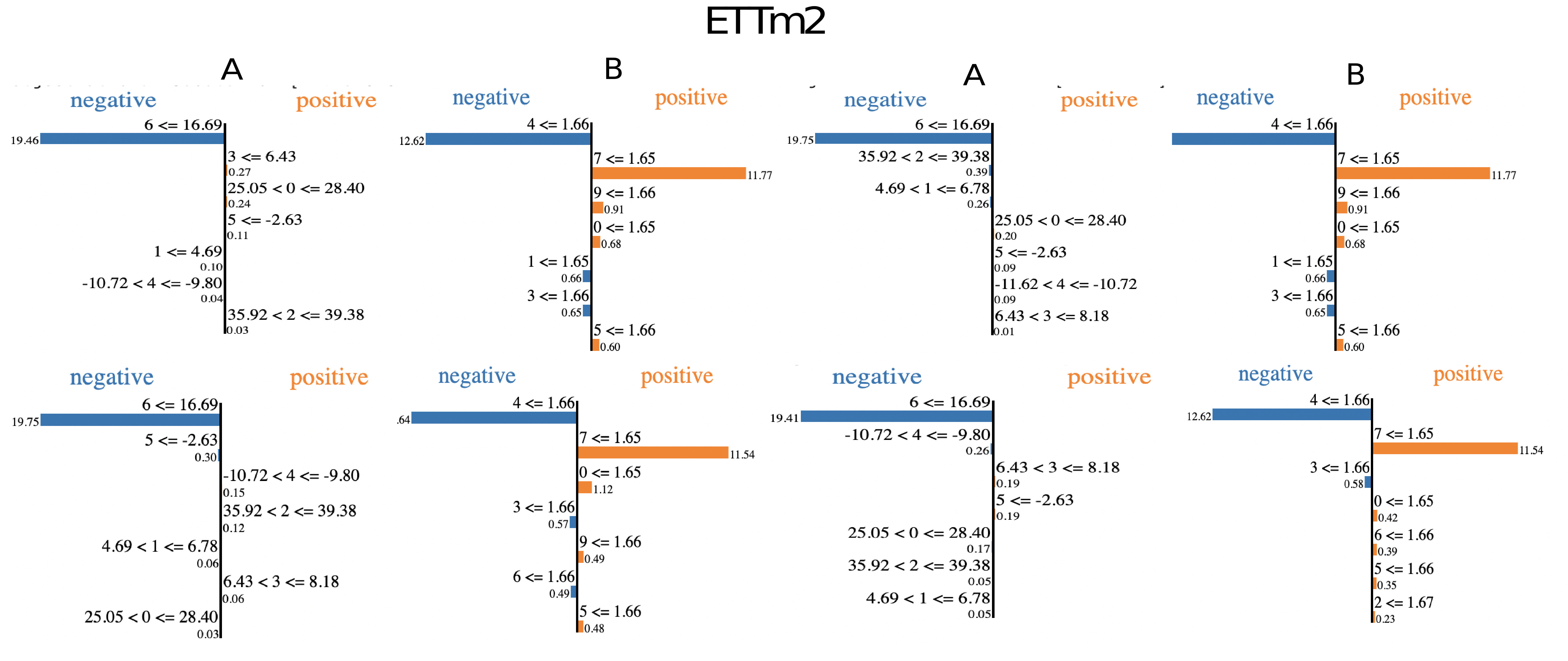}
    \caption{ The Lime Explanation effect of current input (A) and reservoir (B) on output prediction is shown. The test dataset is represented by ETTh1, ETTh2, ETTm1, and ETTm2. The results indicate that current input has a greater impact on feature 6, while the reservoir affects multiple features such as 20, 13, 29, and 7.}
   \label{fig:FXAI}
  \end{center}
  \end{figure*}

In Figure \ref{fig:FXAI}, we present a visual depiction that offers a comprehensive insight into the interpretability of features derived from both the reservoir output $\mathbf{z}_t$ and input feature $\mathbf{u}_t$.

\subsection{Non-Linear output}
In Table~\ref{tab.non_linear_readout}, we present a comparison of the non-linear readout of the reservoir using different activation functions.
\begin{table}[!ht]
\centering

\resizebox{\textwidth}{!}{%
%\setlength{\tabcolsep}{1pt}
%\renewcommand{\arraystretch}{1.1}
%\fontsize{6pt}{6pt}\selectfont
\begin{tabular}{@{}l|cccc|cccc|cccc@{}}
\toprule
\multirow{2}{*}{Activation}
 & \multicolumn{4}{c|}{ETTh1} & \multicolumn{4}{c|}{ETTh2} & \multicolumn{4}{c}{Traffic} \\ 
\cline{2-13} 
               & Linear & ReLu & Tanh & Self-Attention & Linear & ReLu & Tanh & Self-Attention & Linear & ReLu & Tanh & Self-Attention  \\ 
\hline
MSE         & 0.538   & 0.462 & 0.474   & 0.441 &   0.447   & 0.608 & 0.495   & 0.485  & 0.498 & 0.428 & 0.463 &0.426\\
MAE           & 0.562   & 0.470 & 0.475   & 0.455 &  0.348   & 0.552 & 0.468   & 0.478  & 0.327 & 0.293 & 0.314 &0.281\\

\bottomrule
\end{tabular}
}
\caption{Comparison of MSE for different readout activation.} \label{tab.non_linear_readout}

\end{table}

\section{Limitations}\label{sec:lim}
Our method assumes that the compressed memory represents the long context within the time-series data. However, in practice memory compression can result in potential information loss. Therefore, giving the whole context as input instead of compressing it to a fixed size may result in better performance. Additionally, we evaluated our method on standard time-series benchmarks and thus set the maximum horizon value as 720. There may exist datasets that have extremely long-term dependencies. To capture this dependency in such datasets, we would either have to increase the size of the reservoir's input.
%Extremely long context might require bigger reservoir

\clearpage

\end{document}